\documentclass[preprint,authoryear,10pt,a4paper]{elsarticle}
\usepackage[top=0.75in, bottom=0.75in, left=0.75in, right=0.75in]{geometry}
\usepackage{palatino}
\usepackage{amsmath,amsthm}
\usepackage{amssymb}
\usepackage{bbm}
\usepackage{amsfonts}
\usepackage{graphicx}
\usepackage{subfigure}
\usepackage{grffile}
\graphicspath{{graphics/}}
\usepackage[ruled,vlined,linesnumbered]{algorithm2e}
\usepackage{tabularx,booktabs}
\usepackage{inputenc} 
\usepackage[T1]{fontenc}   
\usepackage{url}            
\usepackage{booktabs}       %
\usepackage{amsmath,amsthm}
\usepackage{amssymb}
\usepackage{bbm}
\usepackage{amsfonts}       
\usepackage{nicefrac}       
\usepackage{microtype}      
\usepackage{tikz}
\usepackage{rotfloat}
\usepackage{subfigure}
\usepackage{graphicx}
\usepackage{caption}
\usepackage{appendix}
\usepackage{tablefootnote}
\usepackage{bm}
\usepackage{mathrsfs}
\usepackage{xcolor}
\usepackage{colortbl}

\usepackage{verbatim}
\usepackage{xcolor}
\definecolor{darkblue}{rgb}{0.0,0.5,0.5}
\definecolor{blue}{rgb}{0.0,0.59,0.84}
\definecolor{myblue}{RGB}{0,0,255}

\usepackage[colorlinks]{hyperref}
\hypersetup{colorlinks,breaklinks,linkcolor=blue,urlcolor=blue,anchorcolor=blue,citecolor=blue}
\usepackage{booktabs,caption}
\usepackage{multirow,bigdelim}
\usepackage{lscape}
\usepackage{lineno}
\usepackage{microtype}
\usepackage{soul}
\usepackage{lineno}
\usepackage{fancyhdr}

\usepackage{amsmath,amsfonts,bm}

\def\eqref#1{equation~\ref{#1}}

\def\1{\bm{1}}

\def\vx{{\bm{x}}}

\def\mK{{\bm{K}}}

\def\mQ{{\bm{Q}}}

\def\mV{{\bm{V}}}
\def\mW{{\bm{W}}}
\def\mX{{\bm{X}}}

\DeclareMathAlphabet{\mathsfit}{\encodingdefault}{\sfdefault}{m}{sl}
\SetMathAlphabet{\mathsfit}{bold}{\encodingdefault}{\sfdefault}{bx}{n}

\journal{Artificial Intelligence for Transportation}

\begin{document}

\begin{frontmatter}

\title{{\fontfamily{lmss}\selectfont \textbf{Exploring the Roles of Large Language Models in Reshaping Transportation Systems: A Survey, Framework, and Roadmap}}}

\author[label1]{Tong Nie}
\author[label2]{Jian Sun}
\author[label1]{Wei Ma\corref{cor1}}
\ead{wei.w.ma@polyu.edu.hk}

\address[label1]{Department of Civil and Environmental Engineering, The Hong Kong Polytechnic University, Hong Kong SAR, China}
\address[label2]{Department of Traffic Engineering, Tongji University, Shanghai, 201804, China \\
\vspace{3ex}
\textbf{Project page:} \url{https://github.com/tongnie/awesome-llm4tr}
}

\cortext[cor1]{Corresponding author.}

\begin{abstract}
Modern transportation systems face pressing challenges due to increasing demand, dynamic environments, and heterogeneous information integration. The rapid evolution of Large Language Models (LLMs) offers transformative potential to address these challenges. Extensive knowledge and high-level capabilities derived from pretraining evolve the default role of LLMs as text generators to become versatile, knowledge-driven task solvers for intelligent transportation systems. This survey first presents LLM4TR, a novel conceptual framework that systematically categorizes the roles of LLMs in transportation into four synergetic dimensions: information processors, knowledge encoders, component generators, and decision facilitators. Through a unified taxonomy, we systematically elucidate how LLMs bridge fragmented data pipelines, enhance predictive analytics, simulate human-like reasoning, and enable closed-loop interactions across sensing, learning, modeling, and managing tasks in transportation systems. For each role, our review spans diverse applications, from traffic prediction and autonomous driving to safety analytics and urban mobility optimization, highlighting how emergent capabilities of LLMs such as in-context learning and step-by-step reasoning can enhance the operation and management of transportation systems. We further curate practical guidance, including available resources and computational guidelines, to support real-world deployment. By identifying challenges in existing LLM-based solutions, this survey charts a roadmap for advancing LLM-driven transportation research, positioning LLMs as central actors in the next generation of cyber-physical-social mobility ecosystems.
\textbf{Online resources can be found in the project page}: \url{https://github.com/tongnie/awesome-llm4tr}.
\end{abstract}

\begin{keyword}
Large Language Models, Vision-Language Models, Intelligent Transportation Systems, Transportation Management, Foundation Models, Generative AI, Survey 
\end{keyword}

\end{frontmatter}

\tableofcontents

\section{Introduction}\label{Introduction}
\subsection{Motivation}


Modern transportation systems, characterized by their cyber-physical-social complexity, face unprecedented challenges such as congestion, resilience, sustainability, and adaptability to dynamic urban environments \citep{dimitrakopoulos2010intelligent}.
Traditional transportation management methods, however, often struggle with complex real-world data, increasing demand, human factors, and interaction with infrastructures.
With the advancement of deep learning and big data techniques, the integration of transportation systems with advanced artificial intelligence (AI) tools, known as intelligent transportation systems (ITS) \citep{zhang2011data}, has led to significant progress in both academia and industry.
Within the ITS framework, current transportation management strategies can often be organized around the \textit{sensing-learning-modeling-managing} paradigm. During the past decades, this data-centric paradigm has demonstrated promising results combined with machine learning and AI-driven solutions. 

However, the fast emergence of multimodal mobility ecosystems presents unprecedented technical and operational challenges. The combination of emerging mobility solutions such as autonomous vehicles, cloud computing, drone logistics, human-robot interactions, shared mobility platforms, and AI-powered intersection control has exposed critical limitations in current ITS architectures \citep{guerrero2015integration}. 
Traditional ITS rely heavily on static models and fragmented data pipelines. 
Thus key challenges may stem from the need to reconcile heterogeneous and large-scale data streams across collaborative cloud-edge-end interfaces, manage real-time decision conflicts between human and machine agents in mixed autonomy environments, ensure secure interoperability among proprietary platforms with competing optimization objectives, and develop novel modeling approaches that are applicable to overlapping mobility networks (ground, aerial, shared) beyond conventional traffic flow theories.

\begin{figure}[!htbp]
  \centering
  \captionsetup{skip=1pt}
  \includegraphics[width=1\columnwidth]{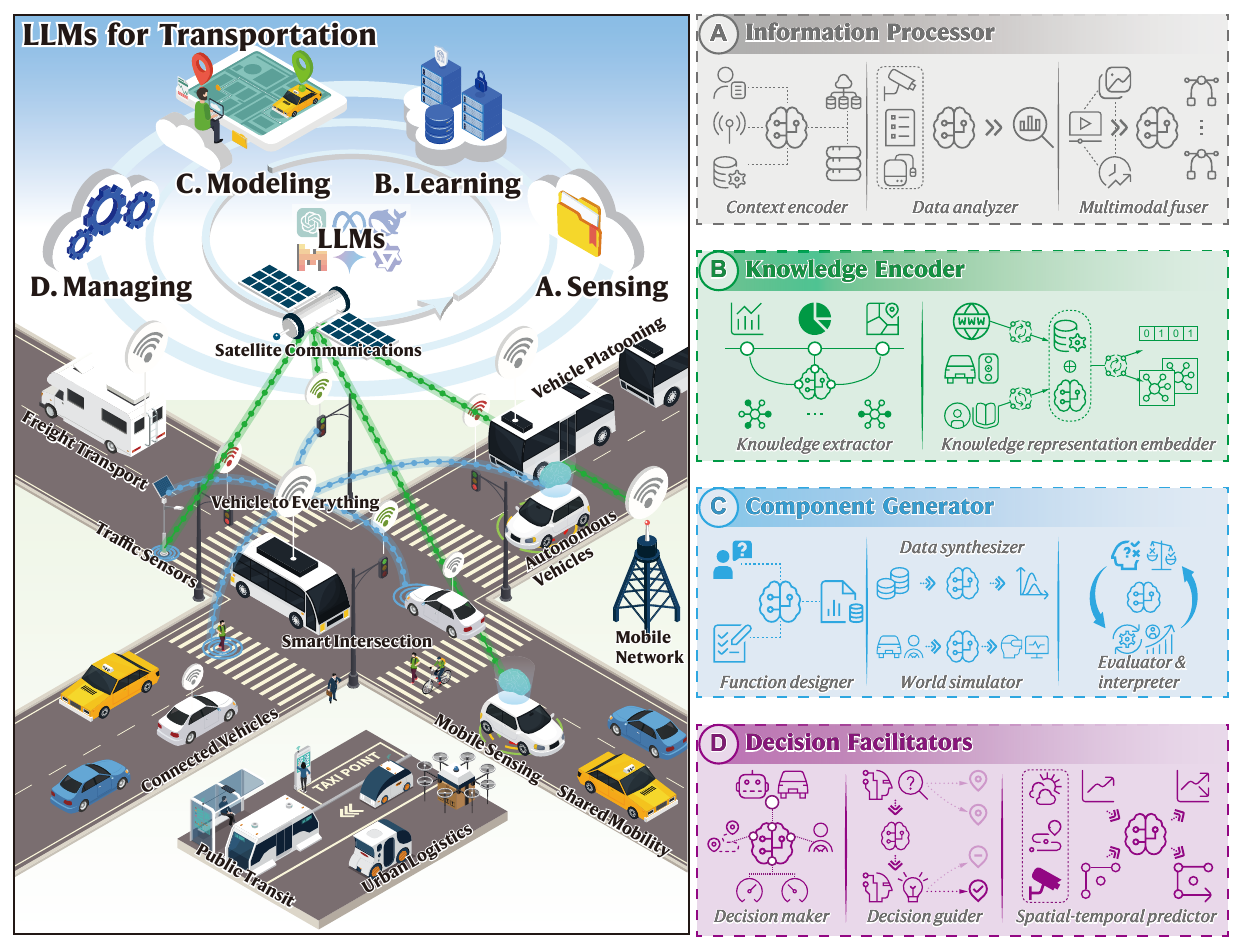}
  \caption{{The LLM4TR framework proposed in this survey. We categorize main tasks in transportation systems as sensing, learning, modeling, and managing. The functionality of LLMs for each of them is structured as information processor, knowledge encoder, component generator, and decision facilitator. Based on this taxonomy, we provide a comprehensive review of existing literature.}}
  \label{fig:intro}
\end{figure}

The rapid evolution of Large Language Models (LLMs) offers a paradigm shift to overcome these inherent barriers. With emergent capabilities such as language understanding, in-context learning, multimodal reasoning, and human-like decision making, LLMs have evolved beyond their initial functionality as text generators, becoming versatile and general problem solvers \citep{zhao2023survey,minaee2024largelanguagemodelssurvey,shanahan2024talking}.
They have not only revolutionized the field of machine learning, but also demonstrated remarkable performances in application-oriented domains \citep{kaddour2023challenges,naveed2023comprehensive}.
Their ability to process unstructured data, encode domain-agnostic knowledge, solve complex tasks, and generate context-aware solutions, aligns seamlessly with the demands of future ITS. In addition, natural language can serve as a universal interface for interpretable human-machine interaction in transportation systems.


Given these desirable properties of LLMs, the potential of LLMs to address the above challenges of modern ITS has recently gained great popularity. 
Far beyond conversational chatbots, LLMs can be a transformative force for transportation systems \citep{qu2023envisioning,lv2023large,mahmud2025integrating}.
For example, LLMs improve the accuracy of traffic prediction by integrating spatiotemporal patterns with semantic context \citep{ren2024tpllm}, automate multimodal traffic scenario understanding \citep{zhou2023vision}, optimize real-time traffic control through natural language interfaces \citep{lai2023large}, generate high-fidelity simulation environments \citep{zhao2024drivedreamer}, facilitate communication between vehicles and users \citep{cui2024drive}, and hence significantly improve the operational efficiency of ITS. 
These pioneering studies have suggested that LLMs can play a powerful role in shaping future transportation systems in various aspects and provide promising solutions to address the limitations of current ITS.

However, despite the early successes of pioneering studies in demonstrating the promise of integrating
LLMs into transportation systems, they have explored isolated applications without articulating how the functionalities of
LLMs interconnect throughout transportation systems. In addition, although these studies provide practical experiences for the utilization of LLMs, there is an absence of a principled framework to guide such integrations.
Therefore, there is a great need to build a unified and standard conceptual framework to guide future progress in this rapidly evolving area.

\begin{table}[!htbp]
{
\caption{Comparison with existing reviews about LLMs and generative AI for transportation (up to the date of this review).}
\label{tab:related_review}
\centering
 \resizebox{1\textwidth}{!}{
    \begin{tabular}{l|c|c|p{0.18\columnwidth}|p{0.25\columnwidth}|p{0.25\columnwidth}|p{0.3\columnwidth}}
    \toprule
    {\bf Survey} & {\bf Year} & {\bf Venue} & {\bf Taxonomy} & {\bf Scope} & {\bf Focus} & {\bf Perspective \& Contributions} \\
    \midrule
    \cite{zhou2023vision} & 2023 & IEEE TIV & Tasks & Perception, scene understanding, navigation, and decision making of autonomous driving & Applications of VLMs in autonomous driving & Focuses only on VLMs for AVs without covering broader LLMs' applications in transportation \\
    \midrule
    \cite{yan2023survey} & 2023 & ArXiv & Tasks & Traffic perception, prediction, simulation, and decision-making & Generative AI applications in ITS tasks & Task-based classification; emphasis on general generative models; lacks LLM-specific taxonomy \\
    \midrule
    \cite{zhou2024urban} & 2024 & TRANS. RES. E & Applications and architecture & Urban mobility planning and design & Hierarchical foundation model concept for urban systems & Focuses on urban mobility architecture and models; lacks LLM-specific taxonomy \\
    \midrule
    \cite{zhang2024advancing} & 2024 & IJCRS & Applications and tasks & Traffic management, safety, autonomous driving & LLMs’ benefits in ITS performance metrics & Application-level review; without system-level methodological framework \\
    \midrule
    \cite{wandelt2024large} & 2024 & Appl. Sci. & Applications & AV, traffic, tourism, etc. & Overview of literature and challenges & High-level summary of applications; without analytical framework or taxonomy \\
    \midrule
    \cite{gan2024large} & 2024 & Adv. Eng. Info. & Tasks and deployment & Perception, prediction, control, and simulation for ITS; AV navigation, planning, and decision making & Application and deployment of large models for ITS and AV & Task- and deployment-centric; lacks methodological taxonomy of LLMs across system levels \\
    \midrule
    \cite{zhang2024large} & 2024 & ArXiv & Data pipelines and models & Traffic and demand forecasting tasks & Tokenization, prompting, embedding and fine-tuning; zero/few-shot prediction & Focuses on training/prediction techniques only; lacks system integration perspective \\
    \midrule
    \cite{choi2024gentle} & 2024 & ArXiv & Models & Traffic data generation, estimation, prediction, and representation learning & Generative models in transportation, such as GAN, VAE, and diffusion models & Emphasis on architectures and applications of deep generative models; not LLM- or role-specific \\
    \midrule
    \cite{mahmud2025integrating} & 2025 & IEEE TITS & Applications & Prediction, control, signal optimization, V2X, and public transport & Use of LLMs in ITS optimization & Application-centric; lacks methodological or system-level taxonomy; without practical guidance \\
    \cmidrule[0.9pt]{1-7}
    \textbf{This survey} & 2025 & AIT & \textbf{Methodological roles of LLMs} & \textbf{Broader coverage across transportation domains: sensing, learning, modeling, and managing} & \textbf{Functional roles of LLMs in transportation system integration} & \textbf{First review to systematically define and categorize LLMs across the transportation lifecycle using a role-centric taxonomy; practical guidelines for role-based integration.} \\
    \bottomrule
    \end{tabular}}
    }
\end{table}

\subsection{Scope and focus}

Faced with both opportunities and challenges, a systematic survey and framework is urgently needed.
Therefore, this survey seeks to fill the above knowledge gaps by presenting a comprehensive review of the latest literature, introducing a unified conceptual framework called \textit{large language models for transportation (LLM4TR)}, and structuring a corresponding taxonomy to elucidate the roles of LLMs in transportation (see Fig. \ref{fig:intro}).
Specifically, this survey provides the first comprehensive methodological review of LLMs in transportation research, emphasizing their transformative roles rather than isolated applications. 
Our analysis spans diverse applications within the sensing-learning-modeling-managing paradigm, such as traffic prediction, autonomous driving, safety analytics, traffic control and operation, traffic simulation, and urban mobility optimization. 
To articulate the research scope and future directions, we propose LLM4TR, a conceptual framework that positions LLMs as a core that dynamically adapts its roles to synergize sensing, learning, modeling, and managing tasks.
A structured taxonomy is followed to classify how these studies integrate LLMs into transportation systems and what enhancements LLMs can offer.
Finally, we also provide an introductory review of key techniques about LLMs and an informative collection of available resources for real-world development of LLMs in transportation systems.

We are also aware of several related review articles on LLMs or generative AI techniques for transportation. {Table~\ref{tab:related_review} provides a comprehensive comparison of these surveys in terms of their publication details, taxonomy, scope, focus, and key differences compared to our work.
Most of the existing reviews adopt a taxonomy based on application domains or task categories. For example, \cite{yan2023survey}, \cite{gan2024large} and \cite{mahmud2025integrating} structure their surveys around application-specific use cases such as traffic prediction, autonomous driving, or traffic management. 
However, they do not systematically distinguish how LLMs function within transportation systems from a methodological perspective.
In contrast, our review differentiates itself from the prevailing tasks- or application-oriented ones with a fresh perspective, novel taxonomy, broader coverage, and practical guidance.
We systematically analyze how LLMs can serve distinct methodological functions across the transportation system lifecycle. This taxonomy enables us to unify technical features of LLMs with their practical integration in transportation systems, offering practitioners with actionable guidance on selecting and configuring LLMs for each stage of system design and operation.
}

Furthermore, to support future development and encourage community engagement, we curate a GitHub repository at link: \url{https://github.com/tongnie/awesome-llm4tr}, with a supporting collection of datasets, benchmarks, and tools for LLM-driven transportation research. We will continue to update this online project so as to provide a platform for tracking the {latest} advances in the field.
Finally, we hope that this review will provide the research community with a broader perspective that will advance the application of LLMs in transportation research and practices. 

\subsection{Contribution}
{Our role-centric framework not only unifies disparate application domains but also provides practical considerations that are absent from earlier task- or application-centric reviews.
To the best of our knowledge, this is the first review to propose a methodological taxonomy of LLMs in transportation, grounded in their functional roles, and supported by a broad and up-to-date coverage of the literature as well as practical guidelines.}
Specifically, this survey makes four key contributions to the community:
\begin{enumerate}
    \item \textit{LLM4TR framework}: We provide the first comprehensive review on LLMs in transportation research from the methodological perspective, focusing on the four main categories of problems in transportation systems such as sensing, learning, modeling, and managing, and covering a broad spectrum of topics.
    
    \item \textit{Unified taxonomy}: A structured taxonomy is further developed to elucidate the roles of LLMs in transportation, including \textit{information processor, knowledge encoder, component generator, and decision facilitator}. Based on this taxonomy, we clearly illustrate how to integrate LLMs to enhance current transportation systems. 
    
    \item \textit{Research trend and outlook}: We present a visualization of the current research trend in the community and identify the current focus and existing challenges. 
    Possible future directions are also discussed, focusing on both underexplored techniques and pathways for real-world deployment.
    
    \item \textit{Practical guidance}: We provide a collection of useful resources including datasets, literature, libraries, and hardware requirements, to support the grounding of LLMs in transportation domains. 
    We also create an online platform for researchers and practitioners in the community to track the latest advances of LLMs in transportation and hope to provide a potential roadmap for advancing this emerging field.
\end{enumerate}

\subsection{Text organization}
The remainder of this paper is structured as follows. Section \ref{sec:background} introduces foundational concepts and preliminaries, including tasks in transportation systems and background of LLMs.
Section \ref{sec:framework} introduces the LLM4TR framework and its taxonomy.
Sections \ref{sec:information_processor} to \ref{sec:decision_facilatator} detail specific roles of LLMs in transportation systems.
Section \ref{sec:practical_guidance} provides practical guidance for deploying LLMs, including datasets and resources.
Section \ref{sec:discussion} discusses future opportunities and challenges.
Section \ref{sec:conclusions} concludes this survey.

\section{Background and Overview}\label{sec:background}
In this section, we first provide an overview of fundamental tasks in transportation systems.
Then we briefly review the recent advances of LLMs by introducing the background, key findings, and mainstream techniques.

\subsection{Overview of tasks in transportation systems}\label{sec:transportation_overview}

Modern transportation systems, characterized by their cyber-physical-social complexity, require innovative methods to address increasing challenges in congestion, sustainability, and resilience \citep{vahidi2018energy,ganin2019resilience}. 
Conventional methods, based on static models and fragmented data, struggle with the exponential growth of multimodal networks, human-centric mobility behaviors, and dynamic urban environments \citep{wang2019enhancing}. 
To address these challenges, contemporary transportation research organizes methodological approaches around four fundamental tasks: \textit{sensing, learning, modeling, and managing.} 
These interconnected domains form the backbone of modern ITS, allowing data-driven operation and management of transportation networks. 
This section briefly explores advances in each task, emphasizing their methods, synergetic relationships, and potential to shape the next generation of smart transportation systems.



\subsubsection{Sensing}
\textbf{Sensing refers to the acquisition of traffic data and the environmental perception process.}
Sensing forms the foundation of modern ITS, and it focuses on data acquisition to capture real-time and historical traffic dynamics, environment conditions, and traveler behaviors \citep{gentili2012locating}. Traditional sensing methods rely on infrastructure-based sensors such as loop detectors, cameras, and radar systems, which provide aggregate traffic metrics such as volume, speed, and occupancy. However, emerging edge and mobile sensing technologies, including distributed IoT devices, GPS probes, smartphones, Wi-Fi sensors, and social media feeds, are improving data granularity by providing details of individual mobility patterns \citep{guerrero2018sensor, kanarachos2018smartphones,van2018autonomous}.
The integration of these advanced sensing methods is crucial not only for traffic monitoring but also for enabling downstream applications in learning, modeling, and real-time management of transportation networks.

These sources, while rich in detail, are often accompanied with challenges such as noise, sparsity, and heterogeneity \citep{zhang2024comprehensive,zheng2025estimating}. To address these issues, researchers employ advanced 
data enhancement techniques such as denoising, imputation, and super-resolution that can reconstruct high-resolution traffic states from sparse or low-quality inputs. 
In addition to industry-driven technology advances, academic studies have also focused on the challenges of heterogeneous data sources and privacy concerns \citep{fries2012meeting,zhu2018big}.
Key innovations such as edge computing architectures that preprocess sensor data at the source to reduce latency \citep{arthurs2021taxonomy}, and privacy-preserving sensing frameworks using distributed and federated learning to analyze mobility patterns without exposing individual trajectories \citep{zhou2015privacy}. 
Recent work also emphasizes integrating multi-source data (e.g., combining ride-hailing trajectories with loop detector counts) with data fusion strategies to enhance spatial-temporal coverage, thus enabling a more holistic view of urban mobility patterns \citep{el2011data}.

\subsubsection{Learning}
\textbf{Learning refers to pattern recognition and predictive analytics of traffic data.} Learning in transportation systems involves both machine learning and data-driven predictive learning approaches that extract actionable insights from large and heterogeneous datasets, bridging the gap between sensing and decision-making \citep{kumar2021applications,shaygan2022traffic}. 
Early work focused on statistical methods and shallow machine learning models to assist in pattern recognition and traffic data mining; however, the last decade has witnessed a surge in deep learning, reinforcement learning, and generative learning frameworks \citep{veres2019deep,haydari2020deep,lin2023generative}. Many complex predictive tasks can be solved by end-to-end learning methods, such as dynamic routing, traffic prediction, anomaly detection, and safety-critical applications \citep{veres2019deep}. This shift is driven by the need to handle massive amounts of traffic big data generated by increasing traffic participants and infrastructures with scalability.

Deep learning transforms raw sensor data into actionable insights through three primary paradigms: supervised learning, unsupervised learning, and reinforcement learning \citep{veres2019deep}. 
A representative application in transportation lies in graph-based deep learning, where heterogeneous data (e.g., traffic flow, social media activity, traffic participants) are structured into spatio-temporal graphs to uncover latent correlations and interactions \citep{xue2025data}. 
Graph neural networks (GNNs) have proven effective for tasks such as traffic prediction and interaction modeling by encoding road network topology and traffic dynamics \citep{rahmani2023graph}. 
In addition, unsupervised learning frameworks (e.g., deep generative models \citep{choi2024gentle}) are used to learn the distribution of mobility patterns and generate simulated system states in large-scale networks. 
More recent advances include physics-informed neural networks that embed traffic flow equations into learning architectures \citep{shi2021physics}, multi-agent RL frameworks coordinating connected vehicles or traffic controllers \citep{farazi2021deep}, transfer learning techniques enable knowledge sharing across cities with diverse traffic patterns \citep{tang2022domain}, and causal learning methods disentangle confounding factors in traffic analysis \citep{liu2024estimating}. 


\subsubsection{Modeling}
\textbf{Modeling refers to the formulation and simulation of transportation systems.}
Modeling in transportation research aims to replicate real-world traffic phenomena through the development of mathematical, simulation-based, and data-driven models that represent traffic dynamics, travel behaviors, and infrastructure interactions \citep{daganzo1997fundamentals}. 
These high-fidelity modeling techniques serve as a virtual testbed for scenario analysis. 
During the past decade, researchers have developed a spectrum of principled models from discrete choice and activity-based models for travel demand analysis to complex simulations using deep neural networks \citep{golob2003structural,di2016boundedly,chen2010review,raadsen2020aggregation}. 
These models represent system dynamics across multiple scales. Microscopic models such as car-following theory (e.g., Intelligent Driver Model) simulate individual vehicle interactions \citep{gipps1981behavioural}, while macroscopic models use fluid dynamics analogues through the Lighthill-Whitham-Richards equations \citep{papageorgiou1998some}. Mesoscopic approaches balance computational efficiency and behavioral realism using queueing networks \citep{burghout2005hybrid}.

Emerging approaches further couple analytical models with data-driven modeling techniques \citep{zhang2011data,chen2016promises}. These implementations combine traditional paradigms with real-time data streams for virtual-physical synchronization \citep{zhang2025operational}. For instance, dynamic traffic assignment (DTA) models are widely used to simulate network-wide vehicle and passenger trajectories, incorporating behavioral factors like route choice and departure time decisions \citep{janson1991dynamic,wang2018dynamic}. 
However, traditional DTA struggles with scalability in large, multimodal networks \citep{pi2019general}. 
To address this, data-driven optimization frameworks such as simulation-based optimization integrate multi-source data to estimate demand-supply parameters iteratively \citep{osorio2013simulation,osorio2015computationally}. This is achieved by minimizing discrepancies between simulated and observed traffic states by adjusting origin-destination matrices and network supply attributes \citep{zhou2010information,wu2018hierarchical,ma2018estimating,ma2020estimating}. 
Today, simulation frameworks have evolved from standalone tools (SUMO, VISSIM, MATSim, etc.) to holistic platforms enabling large-scale scenario testing \citep{li2023scenarionet}. Both microscopic and macroscopic behaviors and multimodal data can be integrated into foundational neural networks to simulate network-scale mobility patterns \citep{chen2024data}.

\subsubsection{Managing}
\textbf{Managing refers to the optimization and control strategies for the operation of transportation systems.}
Transportation management leverages insights from sensing, learning, and modeling tasks with optimization theory and control systems to improve traffic operations and network performance \citep{papageorgiou2003review}. Management tasks leverage real-time data to dynamically adjust operational strategies, such as adaptive traffic signal control and real-time route guidance \citep{fu2001adaptive,guo2019urban}. 
For example, model predictive control enables real-time signal timing adjustments \citep{ye2019survey}, mixed-integer programming optimizes fleet dispatching for shared mobility services \citep{mourad2019survey}, and network-level coordination uses Nash bargaining solutions to balance stakeholder interests for congestion pricing schemes \citep{de2011traffic}.

Cutting-edge approaches exploit end-to-end learning, where control policies directly map sensor inputs to actionable signals.
Deep reinforcement learning (DRL) has proven to be effective for adaptive control, where agents learn optimal policies by interacting with simulated or real-world environments and maximizing cumulative rewards \citep{farazi2021deep}. 
DRL has been widely used in various traffic control applications, such as ramp metering control \citep{han2023leveraging}, intersection signal control \citep{chu2019multi}, perimeter control \citep{chen2022data}, and vehicle platooning control \citep{li2021reinforcement}. 
In addition, the rise of connected vehicles and edge computing has prompted the development of cloud-based management systems that can process vast amounts of vehicle trajectory data to optimize signal timings without additional infrastructure \citep{wang2024traffic}. 



\subsection{Background of LLMs}
{LLMs} refer to Transformer-based \citep{vaswani2017attention} language models that contain hundreds of billions (or more) of parameters being trained on massive text data \citep{zhao2023survey,shanahan2024talking}.
The Internet-scale training data and extensive model parameters enable LLMs to have impressive capabilities, from a natural language modeler to a general problem solver.
To have a quick understanding of modern LLMs, this section briefly introduces the backgrounds and preliminaries of LLMs.

\subsubsection{Emergent capabilities of LLMs}
The emergent abilities of LLMs are formally defined as ``the abilities that are not present in small models but arise in large models \citep{wei2022emergent}'', which is one of the most significant properties that distinguish LLMs from previous pretrained language models. These emergent abilities include \textit{in-context learning, instruction following, and step-by-step reasoning}, which is the result of the \textit{scaling laws}.

\textbf{Scaling laws.} 
The emergent capabilities of LLMs are fundamentally tied to scaling laws, which describe predictable performance improvements as models scale in size, training data, and computational resources. Empirical studies have shown that as these models grow, they exhibit enhanced capabilities in understanding and generating human-like text. 
Pioneering work by \cite{kaplan2020scaling} established that model loss decreases predictably with increases in model parameters, dataset size, and training computation, enabling systematic optimization of LLM architectures. Subsequent research by \cite{hoffmann2022training} refined these principles, suggesting that optimal performance arises from balancing model size and training data size, as exemplified by the compute-optimal Chinchilla model. Crucially, scaling laws stand the emergence of novel abilities of LLMs. \cite{wei2022emergent} identified that  
performance rises abruptly once models surpass critical thresholds in scale. 

\textbf{In-context learning.} 
A prominent feature of modern LLMs is in-context learning (ICL), the ability to adapt to new tasks dynamically through input contextual examples or related knowledge without re-training or gradient updates. Formally introduced by \cite{brown2020language}, ICL enables few-shot or zero-shot generalization by inferring latent task structures from input prompts. 
Specifically, LLMs are provided with a natural language instruction and/or several task demonstrations, it can generate the expected output for the test task instances by completing the word sequence of input text.
Formally, given a set of $k$ paired natural language query-answer demonstrations $\mathcal{D}=\{(x_1,y_1),\dots,(x_k,y_k)\}$, the task description $T$, and the target query $x_{k+1}$, LLMs generate the prediction of the $\hat{y}_{k+1}$ by learning from the context:
\begin{equation}\label{eq:ICL}
    \text{LLMs}(T,\mathcal{D},(x_{k+1},\_\_))=\hat{y}_{k+1},
\end{equation}
where the ground truth answer is left as a blank to be predicted by the LLM. Eq. \ref{eq:ICL} requires no gradient steps.
Empirical studies show that ICL performance scales with model size, which transforms LLMs into versatile, prompt-programmable systems \citep{wei2022emergent}.

\textbf{Instruction following.}
Instruction following reflects the capacity of LLMs to execute open (unseen) tasks by adhering to natural language descriptions. This ability is cultivated through instruction tuning, a process where models are fine-tuned on datasets pairing instructions with desired outputs \citep{sanh2021multitask,wei2021finetuned}. \cite{ouyang2022training} suggested that reinforcement learning from human feedback (RLHF) \citep{christiano2017deep} aligns LLMs with user intent, enabling robust generalization to unseen instructions. \cite{chung2024scaling} further showed that multi-task instruction tuning can enhance cross-task transfer by teaching models to decode task semantics from prompts. Instruction following enables LLMs to follow the instructions to generalize new tasks, bridging the gap between human intent and model behavior.

\textbf{Step-by-step reasoning.}
Advanced LLMs exhibit step-by-step reasoning, solving complex problems via intermediate logical chains or hierarchical steps akin to human cognition. \cite{wei2022chain} formalized this as chain-of-thought (CoT) prompting, where models structure explicit reasoning traces before final answers, markedly improving performance on arithmetic, symbolic, and commonsense tasks. \cite{kojima2022large} found that even zero-shot CoT, triggered by phrases like ''Let’s think step by step," elicits coherent reasoning in sufficiently large models. This capacity transforms LLMs into interpretable problem solvers, enabling decision making in domains requiring structured logic, such as mathematics and program synthesis. Other advance prompting strategies include CoT with self-consistency (CoT-SC) \citep{wang2022self}, tree of thought (ToT) \citep{yao2023tree}, and graph of thought (GoT) \citep{besta2024graph}.

\subsubsection{Mainstream LLMs}

The evolution of LLMs has been driven by advances in Transformer architectures \citep{vaswani2017attention}, enabling unprecedented scalability and performance across natural language processing tasks. 
These models are pretrained on extensive text corpora, enabling them to understand and generate human-like text.
Early foundational models like BERT \citep{devlin2019bert} introduced bidirectional context learning through masked language modeling. While GPT-3 \citep{brown2020language} features autoregressive pretraining and few-shot learning via its 175 billion-parameter architecture. Subsequent innovations include T5 \citep{raffel2020exploring}, which unified NLP tasks under a text-to-text framework, and PaLM \citep{chowdhery2023palm}, which demonstrated emergent reasoning capabilities at scale. Recent models prioritize efficiency and human alignment. For example, LLaMA \citep{touvron2023llama,touvron2023llama2} optimized training for smaller and open-access models. GPT-4 \citep{achiam2023gpt} and Gemini \citep{team2024gemma} enhanced multimodal and instruction-following abilities.  
A recent addition to this landscape is DeepSeek-R1 \citep{guo2025deepseek}, an open-source LLM released in January 2025 by a Chinese startup. DeepSeek-R1 has garnered attention for its competitive performance in complex tasks such as mathematical reasoning and coding, achieved with significantly lower computational resources and cost compared to its counterparts. 
We summarize several representative LLMs from technical perspectives in Tab. \ref{tab:llm_summary_expanded}.

\begin{table}[t]
\centering
\caption{Overview of mainstream LLMs.}
\label{tab:llm_summary_expanded}
\resizebox{0.95\textwidth}{!}{
\begin{tabular}{lcccccl}
\toprule
\textbf{Model} & \textbf{Release} & \textbf{Organization} & \textbf{Size} & \textbf{Data} & \textbf{Hardware}  & \textbf{Public} \\
 & \textbf{Date} &  & \textbf{(B)} & \textbf{(TB)} & \textbf{Cost} & \textbf{Access} \\
\midrule
T5 \citep{raffel2020exploring} & 2019.10 & Google & 11 & 750 GB of text & 1024 TPU v3 & Yes \\
GPT-3 \citep{brown2020language} & 2020.5 & OpenAI & 175 & 300 B tokens & -  & No \\
PaLM \citep{chowdhery2023palm} & 2022.4 & Google & 540 & 780 B tokens &  6144 TPU v4  & No \\
LLaMA \citep{touvron2023llama} & 2023.2 & Meta & 65 & 1.4 T tokens & 2048
A100 GPU  & Partial\footnote{1} \\
GPT-4 \citep{achiam2023gpt} & 2023.3 & OpenAI & - & - & -  & No \\
LLaMA-2 \citep{touvron2023llama2} & 2023.7 & Meta & 70 & 2 T tokens & 2000
A100 GPU  & Yes \\
Mistral-7B \citep{jiang2023mistral7b} & 2023.9 & Mistral AI & 7 & - & -  & Yes \\
Qwen-72B \citep{bai2023qwen} & 2023.11 & Alibaba & 72 & 3 T tokens & -  & Yes \\
Grok-1 & 2024.3 & xAI & 314 & - & -  & Yes \\
Claude 3 & 2024.3 & Anthropic & - & - & -  & No \\
GLM-4-9B \citep{glm2024chatglm} & 2024.6 & Zhipu AI & 9 & 10 T tokens & -  & Yes \\
LLaMA-3.1 \citep{dubey2024llama} & 2024.7 & Meta & 405 & 15 T tokens & 16 thousand H100 GPU  & Yes\\

Gemma-2 \citep{team2024gemma} & 2024.6 & Google & 27 & 13 T tokens  & 6144 TPUv5p  & Yes \\


DeepSeek-V3 \citep{liu2024deepseek} & 2024.12 & DeepSeek & 671\footnote{2} & 14.8 T tokens & 2048 H800 GPU  & Yes \\
\bottomrule
\end{tabular}}

\vspace{2mm}
\footnotesize
\begin{enumerate}
\item[1] Non-commercial research license.
\item[2] MoE architecture, with 37B activated for each token.
\end{enumerate}
\end{table}

\subsection{Key techniques in LLMs}

LLMs have revolutionized natural language processing (NLP) by achieving state-of-the-art performance across diverse tasks. This section systematically introduces the foundational techniques underpinning modern LLMs: pretraining, architecture design, post-training optimization, and utilization strategies. Each component is critical to the development, refinement, and application of these models, as evidenced by their widespread adoption in academia and industry.

\subsubsection{Pretraining}

Pretraining is the foundational phase where LLMs learn language representations from large-scale corpora. This process enables models to capture syntactic and semantic patterns, facilitating their application to various downstream tasks. 
Since the capacities of LLMs largely rely on the pretraining corpus, high-quality and carefully processed datasets are important. 
Common public sources for pretraining include general-purpose text from the Internet such as webpages, conversations, and online books. In addition, specialized and structured datasets are also used to improve the capabilities of LLMs in a wide range of tasks, such as multilingual text, scientific text, and code bases \citep{min2023recent,naveed2023comprehensive,zhao2023survey}. The scale of pretraining data and computational resources significantly influences the model's performance and generalization capabilities .

The pretraining objectives guide the model to learn robust representations which typically involve unsupervised or self-supervised learning objectives, such as predicting masked tokens or the next word in a sequence. 
For instance, models like BERT employ masked language modeling, while GPT models utilize autoregressive training. 
There are two dominant paradigms:
\begin{itemize}
    \item \textbf{Autoregressive pretraining:} (e.g., GPT series \citep{brown2020language}) trains models to predict the next token in a sequence, promoting coherent text generation capabilities \citep{radford2019language}. 
    Given a token series $\vx=\{x_1,x_2,\dots,x_n\}$, the language modeling (LM) task aims to autoregressively predict the target tokens $\vx_i$ based on the preceding tokens $\vx_{<i}$ in this sequence:
    \begin{equation}\label{eq:lm}
     \max\ell_{\text{LM}}=\max\sum_{i=1}^n\log P(x_i|\vx_{<i}),
    \end{equation}
    \item \textbf{Autoencoding pretraining:} (e.g., BERT \citep{devlin2019bert} and BART \citep{lewis2019bart}) masks or corrupts random tokens with replaced spans or masks and trains models to reconstruct them, enhancing bidirectional context understanding. Formally, the objective of autoencoding pretraining is denoted as:
    \begin{equation}
    \max\ell_{\text{AE}}=\max\log P(\tilde{\vx}|\vx_{/\tilde{\vx}}).
    \end{equation}
\end{itemize}

More advanced strategies can be developed by combining the two prototypes.
Models like T5 \citep{raffel2020exploring} unify tasks into text-to-text frameworks, while UL2 (also known as Mixture-of-Denoisers) \citep{tay2022ul2} combines denoising autoencoding and autoregressive objectives.

After LLMs have been pre-trained, a decoding strategy is needed to generate desired textual output. Two prevailing methods include \textbf{search-based} and \textbf{sampling-based} strategies \citep{zhao2023survey}.
The greedy search predicts the most likely token at each step, conditioned on the previously generated context tokens. While the sampling-based method randomly selects the next token based on the probability distribution of contexts to enhance the diversity during generation. The mitigation of the selection of words with extremely low probabilities is crucial to improve the quality of the generation. To control the randomness of sampling, a practical method is to adjust the temperature coefficient of the softmax function to compute the probability of the $j$-th token over the vocabulary, called \textit{temperature sampling} \citep{renze2024effect}:

\begin{equation}
    P(x_k|\vx_{<i})=\frac{\exp(l_k/\tau)}{\sum_{k'}\exp(l_{k'}/\tau)},
\end{equation}
where $l_k$ is the logits of each word and $\tau$ is the temperature coefficient. Reducing $\tau$ increases the chance of selecting words with high probabilities while decreases the chances of selecting words with low probabilities. 

Training LLMs requires significant computational costs. Several optimization techniques are usually adopted to facilitate efficient training under a limited computational budget, such as distributed training (data, tensor and pipeline parallelism \citep{shoeybi2019megatron}) and mixed precision training \citep{micikevicius2017mixed}.  

\subsubsection{Architecture} 

The Transformer architecture, introduced by \cite{vaswani2017attention}, serves as the cornerstone of modern LLMs. 
It employs self-attention mechanism enables dynamic weighting of input tokens, capturing long-range sequential dependencies without recurrence or convolution.
The vanilla Transformer consists of an encoder-decoder structure, with each layer comprising multi-head self-attention and position-wise feed-forward networks. This design enables efficient computation and scalability, making it ideal for large-scale language modeling tasks.
Key components include:

\begin{itemize}
    \item \textbf{Multi-head self-attention.} 
    Central to the Transformer is the self-attention mechanism, which allows the model to weigh the importance of different parts of the input sequence when encoding each token. This mechanism computes attention scores in a pairwise way by comparing query, key, and value vectors derived from the input embeddings. The resulting weighted sum captures contextual relationships, enabling the model to understand the significance of each token in relation to others.
    Multi-head attention extends the self-attention mechanism by employing multiple attention heads, each learning different aspects of the input representation. The outputs of these heads are concatenated and linearly transformed, allowing the model to capture a diverse range of semantic features and relationships within the data. Formally, the input sequence \( \mX \in \mathbb{R}^{n \times d_{\text{model}}} \) is processed through \( h \) parallel attention heads, where each head \( i \) computes scaled dot-product attention as:
    \[
    \begin{aligned}
    &\mQ_i = \mX\mW_i^Q, \quad \mK_i = \mX\mW_i^K, \quad \mV_i = \mX\mW_i^V \quad (\mW_i^Q, \mW_i^K \in \mathbb{R}^{d_{\text{model}} \times d_k}, \ \mW_i^V \in \mathbb{R}^{d_{\text{model}} \times d_v}),  \\
    &\text{head}_i = \text{softmax}\left(\frac{\mQ_i\mK_i^\top}{\sqrt{d_k}}\right)\mV_i, \quad d_k = d_v = d_{\text{model}}/{h}, 
    \end{aligned}
    \]
    followed by concatenation and linear projection:
    \[
    \text{MultiHead}(\mX) = \text{Concat}[\text{head}_1, \ldots, \text{head}_h]\mW^O \quad (\mW^O \in \mathbb{R}^{hd_v \times d_{\text{model}}}).
    \]    
    An issue of standard self-attention is the quadratic complexity, which becomes a bottleneck when dealing with long sequences. Various efficient attention variants are proposed to reduce the computational complexity, such as sparse attention \citep{child2019generating} and FlashAttention \citep{dao2022flashattention}. 
    \item \textbf{Positional encoding.} 
    Since Transformers process input sequences in parallel without inherent sequential order, positional encoding is introduced to inject information about the position of tokens within the sequence. This is typically achieved by adding sinusoidal functions of different frequencies or learned vectors to the input, allowing the model to distinguish between tokens based on their positions. Recent studies also developed more advanced positional embedding techniques to enable Transformers to generalize to sequences longer than those sequences for training, i.e., extrapolation, such as relative position embedding \citep{raffel2020exploring}, rotary position embedding \citep{su2024roformer}, and ALiBi \citep{press2021train}.
    \item \textbf{Layer normalization.} 
    Training large-scale Transformers can be unstable due to factors such as gradient anomaly. Therefore, normalization is a widely adopted technique to stabilize the training process.
    Layer normalization (LN) \citep{ba2016layer} is applied in the vanilla Transformer, in which the mean and variance over all activations per layer are calculated to recenter and rescale the activations. Other techniques such as RMSNorm \citep{zhang2019root} and DeepNorm \citep{wang2024deepnet} are also widely employed in deep Transformers. In addition, recent studies have found that the position of normalization also has a notable impact on LLMs. There are generally three choices, i.e., post-LN, pre-LN, and sandwich-LN \citep{xiong2020layer,ding2021cogview}.
\end{itemize}

Built on the above basic elements, modern LLMs (e.g., GPT-3 \citep{brown2020language}, PaLM \citep{chowdhery2023palm}) scale Transformers by increasing depth (layers) and width (hidden dimensions). 
In existing LLMs, the main architectural variants include encoder-decoder, causal decoder, and non-causal decoder Transformers:
\begin{itemize}
    \item \textbf{Encoder-decoder.} 
    Models like T5 \citep{raffel2020exploring} and BART \citep{lewis2019bart} utilize both encoding and decoding mechanisms, enabling them to perform a wide range of tasks, including translation and summarization. The encoder applies stacked self-attention layers to encode the input sequence, and the decoder performs cross-attention on these representations and autoregressively generates the output.
    \item \textbf{Causal decoder.} 
    As a representative \textit{decoder-only} architecture, causal decoder models introduce the unidirectional attention mask to ensure that each input token can only attend to the past tokens and itself. This mechanism makes them suitable for text generation tasks. Prominent examples are the GPT-series models \citep{radford2018improving,radford2019language,brown2020language}.
    \item \textbf{Non-causal decoder.} Another kind of decoder-only architecture is the non-casual structure. This architecture performs bidirectional attention on prefix tokens and unidirectional attention only on generated tokens.
    One representative prefix decoder LLMs is GLM \citep{zeng2022glm}.
\end{itemize}

To scale the capacity of LLMs efficiently, the Mixture of Experts (MoE) technique can be exploited to combine the above architectures, such as in Swich Transformer \citep{fedus2022switch} and GLaM \citep{du2022glam}. MoE involves sparsely activating a subset of model parameters (the "experts") for each input, allowing the model to handle a vast number of parameters without incurring prohibitive computational costs. This is achieved by employing a trainable gating mechanism to route each input token to the most relevant subset of experts.

Apart from the mainstream Transformer architecture, there are also emerging architectures proposed to alleviate the inherent issues of Transformers (e.g., the quadratic complexity) such as State-Space Models (SSMs) \citep{gu2021efficiently}, Mamba \citep{gu2023mamba}, and RWKV \citep{peng2023rwkv}.

\subsubsection{Post-training}
After pretraining on massive corpus, LLMs obtain the ability to serve as a general problem solver. To adapt them for domain-specific tasks, several post-training techniques can be applied to further refine their capabilities beyond initial pre-training. Three pivotal methodologies in this phase are instruction tuning, alignment tuning, and model adaptation \citep{zhao2023survey,zhang2023instruction,wang2023aligning}. These techniques enhance task generalization, align outputs with human preferences, and optimize models for domain-specific or resource-constrained settings, respectively. In the following, we briefly introduce their objectives, methods, and impacts based on contemporary research.

\textbf{Instruction tuning.} 
Instruction tuning refines LLMs to follow task-specific natural language instructions, enabling zero-shot or few-shot generalization to unseen tasks \citep{wei2021finetuned,chung2024scaling}. Unlike conventional fine-tuning, which trains models on labeled examples for specific tasks, instruction tuning employs datasets comprising task descriptions, input-output pairs, and diverse prompts (e.g., "Summarize this article: [text]"). This approach conditions models to infer task requirements from instructions, better comprehend tasks, and satisfy human expectations across diverse tasks. Representative models that perform instruction tuning include InstructGPT \citep{ouyang2022training} and FLAN-T5 \citep{chung2024scaling}.

Instruction tuning is closed to supervised fine-tuning (SFT) \citep{ouyang2022training} and prompt tuning \citep{liu2021p}. 
SFT performs full-parameter fine-tuning based on pre-trained models using task-specific labeled data (input-output pairs). Instruction tuning is a special form of SFT that fine-tunes a model using instructional task descriptions, with the goal of allowing the model to understand and generalize to unseen instructions. Prompt tuning is a parameter-efficient fine-tuning method (which will be discussed in the latter) that guides the model output by adjusting the prompts in the inputs, usually without updating the pre-trained model parameters, and optimizing only a small number of prompt-related parameters.
The difference between the three concepts is relatively small.
To help differentiate them, we compare different aspects of these techniques in Tab. \ref{tab:sft_it_pt}. To summarize, SFT is the basic full-parameter fine-tuning method, instruction tuning is a variant of its instruction-oriented generalization, and prompt tuning is a parameter-efficient lightweight alternative.

\begin{table}[!htbp]
\centering
\small 
\caption{Comparison between instruction tuning, {SFT}, and prompt tuning.}
\label{tab:sft_it_pt}
\begin{tabularx}{0.95\textwidth}{l|X|X|X}
\toprule
\multicolumn{1}{c|}{\textbf{Aspect}} & 
\multicolumn{1}{c|}{\textbf{{SFT}}} & 
\multicolumn{1}{c|}{\textbf{Instruction Tuning}} & 
\multicolumn{1}{c}{\textbf{Prompt Tuning}} \\
\midrule
\textbf{Objective} & 
Adapt to a single task (e.g., classification, generation) & 
Enhance instruction understanding and cross-task generalization & 
Activate pretrained knowledge via prompts \\
\midrule
\textbf{Training Data} & 
Task-specific structured input-output pairs & 
Multi-task instruction-response pairs (with natural language instructions) & 
Minimal labeled/unlabeled data (reliant on prompt design) \\ 
\midrule
\textbf{Parameter Update} & 
Full parameter fine-tuning & 
Full parameter fine-tuning & 
Optimize only prompt-related parameters (fixed backbone) \\ 
\midrule
\textbf{Generalization} & 
Task-specific optimization (risk of overfitting) & 
Strong cross-task generalization (requires diverse instructions) & 
Depends on prompt design; effective for few/zero-shot learning \\ 
\midrule
\textbf{Computation Cost} & 
High (updates all parameters) & 
High (updates all parameters) & 
Very low (only prompts optimized) \\ 
\midrule
\textbf{Typical Use Cases} & 
Single-task models (e.g., text classifiers) & 
General-purpose models (e.g., ChatGPT prototypes) & 
Resource-constrained scenarios \\ 
\bottomrule
\end{tabularx}
\end{table}

To perform instruction tuning, the first step is to collect instruction-formatted instances in natural language. Task descriptions are obtained either by crowd-sourced human experts or synthetic instances.
Then, these formatted instances are employed to fine-tune LLMs in a supervised learning way. 
Recent studies \citep{wang2022self} have demonstrated that using instruction tuning on public instruction datasets such as Super-NaturalInstructions \citep{wang2022super} and PromptSource \citep{bach2022promptsource} can significantly improve performance on downstream tasks. 

\textbf{Alignment tuning.} 
While pretrained LLMs have impressive generation ability, they may output harmful, biased, or misleading content. 
Thus, alignment tuning focuses on the adjustment of LLMs to comply with human values and preferences, ethical guidelines, and safety standards \citep{wang2023aligning}. This is typically achieved by incorporating human feedback into training loops, often through reinforcement learning (RL) or contrastive learning techniques \citep{ouyang2022training,ziegler2019fine}.

Different from the goals of pretraining and instruction tuning, alignment tuning highlights different aspects of the model output, such as honesty with correctness. These human-centric criteria can be obscure for LLMs to comprehend. Thus, the first step to align LLMs is to collect human evaluations and feedback from experts. In existing LLMs, one of the dominant method for generating human feedback is human annotation \citep{ziegler2019fine,ouyang2022training,wang2023aligning}. Since high-quality human feedback data is crucial for aligning LLMs, this process can be resource-consuming and requires careful treatment.

After collecting and constructing feedback datasets from human experts, a prevalent method for alignment is {RLHF} \citep{christiano2017deep,ziegler2019fine}, where models are fine-tuned using datasets of human preferences to guide their behavior.
RLHF adapts LLMs to human feedback by learning a reward model, incorporating human in the training loop. It involves three stages: 
\begin{enumerate}
    \item \textbf{Collecting human rankings of SFT model output.} The base model is first fine-tuned on high-quality human-generated responses to specific prompts. 
    This step initially aligns the model’s outputs with desired formats and tasks. Then the SFT model generates multiple responses to sampled prompts. Invited human annotators rank these outputs by quality, appropriateness, or alignment with goals. Comparative rankings (e.g., pairwise preferences) are often collected to reduce subjectivity and inconsistency.
    \item \textbf{Training a reward model to predict preferences.} A separate reward model is trained to predict human preferences. It takes a prompt and response as input and outputs a scalar reward. It is trained using pairwise comparison data, optimizing for higher rewards for preferred outputs. The reward model can take on two forms: a fine-tuned language model or a model trained using human preference data, which has a parameter scale much smaller than that of the LLM to be aligned.
    \item \textbf{Optimizing the LLM policy using RL to maximize rewards.} The SFT model is optimized using RL (e.g., Proximal Policy Optimization, PPO; \citep{schulman2017proximal}) to maximize rewards from the reward model. 
    The pretrained LLM acts as the policy that generates an output text, the action is the choice of vocabulary, and the state is the current token sequence.
    A KL divergence penalty can be added to prevent the model from deviating too far from the SFT model, balancing reward optimization with output coherence.
\end{enumerate}

Pioneered by OpenAI's InstructGPT \citep{ouyang2022training}, RLHF reduced about 25\% fewer toxic outputs than GPT-3 compared to GPT-3 while improving response quality. 
RLHF has been pivotal in enhancing the safety and reliability of LLMs in real-world applications, ensuring that LLMs produce outputs that are both accurate and ethically sound.  
Alternative approaches, such as direct preference optimization (DPO; \citep{rafailov2023direct}), bypass explicit reward modeling by directly aligning model likelihoods with human preferences, offering a simpler and RL-free alternative.
Emerging study also explores constitutional AI \citep{bai2022constitutional}, where models self-critique outputs against predefined rules using LLM agents. 
Very recently, a powerful open-source LLM called DeepSeek-R1 \citep{guo2025deepseek} is trained using group relative policy optimization (GRPO) without SFT, simplifying the training by evaluating actions relative to a group of samples \citep{shao2024deepseekmath}.

\textbf{Parameter-efficient fine-tuning.} 
The discussed instruction tuning and alignment tuning methods typically require  full-parameter tuning.
Because LLMs are parameter-extensive, it is computationally prohibitive, memory intensive, and risks catastrophic forgetting. 
Parameter-efficient fine-tuning (PEFT) techniques address these limitations by updating only a small subset of parameters while preserving the model’s inherent capabilities \citep{liu2022few,ding2023parameter}. 
These methods such as adapter tuning, prefix tuning, prompt tuning, and low-rank adaptation strike a balance between task-specific performance and resource efficiency, enabling cost-effective deployment of LLMs across diverse applications. We briefly introduce these them as follows.

\begin{itemize}
    \item \textbf{Adapter tuning} introduces lightweight neural adapters within Transformer layers while keeping the base model frozen. First proposed by \citep{houlsby2019parameter}, adapters are inserted between feed-forward layers or attention blocks and trained on task-specific data. These modules typically consist of down-projection and up-projection layers with a bottleneck architecture, reducing parameter overhead (e.g., <1\% of total parameters). Subsequent work \citep{pfeiffer2020adapterhub} optimized adapter placement and design. During fine-tuning, the adapters are optimized based on task-specific goals, while the parameters of the original LLMs are frozen.
    
    \item \textbf{Prefix tuning \citep{li2021prefix}} prepends task-specific continuous vectors to each Transformer layer's key and value matrices.
    It avoids modifying the base model and enables context-aware adaptation of the attention computation. Since the number of parameters is determined only by the prefix length (typically 10-100 tokens) and the hidden layer dimension, prefix tuning is more scalable than adapter tuning.

    \item \textbf{Prompt tuning \citep{lester2021power}}. 
    Different from prefix tuning, prompt tuning simplifies this approach by prepending trainable tokens only to the input layer, achieving competitive performance with extremely low parameter counts (e.g., 0.01\% of base parameters). 
    During training, only the virtual prompt embeddings would be learned according to task-specific supervisions.

    \item \textbf{Low-Rank Adaptation (LoRA)} \citep{hu2022lora}. LoRA decomposes weight updates during fine-tuning into low-rank matrices, leveraging the hypothesis that task-specific adaptations reside in a low-dimensional subspace. By freezing pretrained weights and injecting trainable rank-decomposition matrices into dense layers, LoRA achieves parameter efficiency without inference overhead. LoRA has been widely adopted by open-source LLMs such as LLaMA \citep{touvron2023llama}.
\end{itemize}

PEFT techniques democratize access to LLMs by reducing computational barriers while retaining their internal knowledge. As LLMs grow in scale, these methods will remain critical for enabling scalable, sustainable, and versatile deployments across applications, especially in low-resource environments.
In summary, post-training techiniques such as instruction tuning, alignment tuning, and model adaptation collectively alleviate the limitations of raw pretrained LLMs, transforming them into controllable, safe, and adaptable systems.

\subsubsection{Practical utilization}
In addition to the above training and adaptation methods for developing powerful LLMs, there are emerging utilization techniques that unlocks the potential of LLMs in real-world applications. We briefly discuss them as these methods have been applied in transportation research by some pioneering work.

\begin{itemize}
    \item \textbf{Prompt engineering.} 
    Designing suitable prompts is crucial to guide LLMs to solve downstream tasks. Prompt engineering is the procedure to manually craft or automatically generate specific prompts that can elicit specific ability of LLMs to produce desired outputs \citep{white2023prompt}. 
    The standard template of a prompt contains four ingredients, they are task description, task input data, contextual information, and prompt style.
    There are several practical guidelines for users to design proper prompt contents \citep{white2023prompt,giray2023prompt,ekin2023prompt,chen2023unleashing}. Generally, expressing the task description understandably, decomposing the task into sub-tasks, using CoT prompting, providing few-shot demonstrations, and adopting role-playing strategies can improve model performances.
    \item \textbf{LLM-based agent.} 
     The context generation, strategic planning, and logical reasoning capabilities of LLMs enable them to solve complex tasks. 
     Therefore, LLMs are integrated into autonomous agents capable of performing tasks involving multistep reasoning and decision making \citep{wang2024survey}. The LLM-based agent systems include three components: \textit{memory, planning, and execution.}
    To solve a given task, the agent first gathers information from the environment and stores it in short-term memory. Then it processes these new data, potentially enhancing it with relevant details retrieved from long-term memory \citep{zhong2024memorybank}. Using the processed information, the planning component formulates the next plan \citep{song2023llm}. The execution component carries out this plan, possibly aided by external tools such as code \citep{gao2023pal}. By continuously repeating this cycle, the LLM-based agent automatically reflects and adjusts its behavior in response to environmental feedback \citep{shinn2023reflexion}, ultimately achieving its goal.

    \item \textbf{Retrieval-augmented generation (RAG).} 
    RAG enhances LLMs by integrating external knowledge retrieval mechanisms, addressing limitations in factual accuracy and domain specificity \citep{guu2020retrieval,gao2023retrieval}.
    Similar to the workflow of LLMs, RAG consists of three steps, including context retrieval, prompt construction, and response generation \citep{lewis2020retrieval}.
    The retriever uses a structured index representation such as dense vectors to search candidate documents. The selected retrieved documents are then integrated into the prompt along with instructions that guide LLMs to exploit the retrieved information to perform actions.
    Finally, LLMs synthesizes outputs conditioned on retrieved content.

    \item \textbf{Tool manipulation.} Fundamentally, LLMs are developed as text generators trained on extensive plain texts. This causes them to be less effective on tasks that aren't optimally represented in textual form, such as numerical computations. Additionally, their abilities are confined to the information available up to their last training update, limiting their access to the most recent data. To address these challenges, recent studies have proposed integrating external tools that can empower LLMs with capabilities that go beyond language modeling \citep{nakano2021webgpt,schick2023toolformer}. For example, LLMs can use the calculator for accurate computation \citep{nakano2021webgpt}, employ search engines to retrieve unknown information \citep{schick2023toolformer}, and adopt the compiler for programming \citep{gao2023pal}. Recetly, LLM-based Model Context Protocol (MCP) introduced by Anthropic provides a unified communication interface between LLMs and external data sources and tools. Through MCP, LLM applications can securely and efficiently access a variety of data resources such as files, databases, APIs, web pages, etc., and at the same time call external tools to perform specific tasks, thus breaking through the limitations of relying on pre-training data alone, and enhancing the LLM's context-awareness and real-world application capabilities.
    \item \textbf{Multimodal LLMs.} Recent advancements have introduced multimodal LLMs (MLLMs) capable of processing and generating not only text, but also images and other data types, thus broadening their applicability \citep{liang2024survey}.
     MLLMs adapt information from various modalities into the text modality to leverage the powerful capabilities of LLMs trained on textual data. 
     An MLLM typically comprises an image encoder for processing images and a language model for text generation, connected via a module that aligns visual and linguistic representations, such as CLIP \citep{radford2021learning}. 
    
\end{itemize}

\section{The LLM4TR Framework}\label{sec:framework}

This section introduces the cornerstone of this survey. By identifying existing challenges of traditional transportation management frameworks, we propose a novel conceptual framework based on LLMs. Then we structure a systematic taxonomy to elucidate the roles of LLMs in this framework. Finally, to give an intuitive overview of the status of current studies, we visualize the research trend matrix based on our taxonomy.

\subsection{Conceptual framework}

Modern transportation systems are facing increasing challenges rooted in the limitations of conventional "four-step" management frameworks introduced in section \ref{sec:transportation_overview}.
Traditional approaches to sensing, learning, modeling, and managing often operate in isolation and have three main dilemmas: (1) the rapid growth of multimodal and heterogeneous data versus the inefficiency of fragmented processing pipelines; (2) the demand for adaptive and human-centric decision making in dynamic environments versus the rigidity of static models trained on historical patterns; (3) the need for scalable and generalizable AI tools for ITS versus the computational and interpretability constraints of conventional machine learning.

These barriers are exacerbated by the exponential growth of cyber-physical-social interactions in transportation systems.
Traditional paradigms struggle to address these limitations. 
Sensing systems grapple with integrating multi-source, unstructured signals, and user queries into holistic representations. Learning frameworks often operate as black boxes, lacking mechanisms to embed domain knowledge or explain predictions. 
Modeling tools based on static or rigorous rules face a fidelity-efficiency trade-off, unable to adapt to changing environments and achieve self-refinement. Managing strategies are far from human-like reasoning about traffic semantics and contextual intentions.


The emergence of LLMs offers a new paradigm to overcome these barriers. 
Unlike conventional AI systems optimized for specific tasks, LLMs exhibit emergent properties such as few-shot learning, human-like reasoning, and cross-task generalization that align with the cyber-physical-social complexity of future transportation systems. 
Four intrinsic characteristics of LLMs position them as foundational enablers for next-generation "transportation intelligence":
\begin{itemize}
    \item \textbf{Context understanding and world knowledge}: LLMs contain vast common sense and domain-agnostic knowledge through pretraining, allowing them to interpret traffic semantics from a comprehensive picture through the lens of urban geography, human behavior, and physical laws.
    \item \textbf{Adeptness at sequence data processing}: Traffic dynamics is inherently sequential, from vehicle trajectories to demand fluctuations. 
    The backbone Transformer architecture of LLMs excel at modeling long-range spatiotemporal dependencies in sensor streams, travel behaviors, or driving interactions.
    \item \textbf{Excellent reasoning and planning abilities}: {CoT} prompting and recursive self-improvement enable LLMs to decompose complex tasks into multi-step reasoning processes with verifiable intermediate states. In addition, LLMs can perform complex tasks by planning to use tools, integrate external knowledge, and learn from demonstrations.
    \item \textbf{Multi-modal information integration}: By aligning textual, visual, and geometric data into unified representations, MLLMs can fuse heterogeneous inputs such as LiDAR point clouds, driver voice commands, infrastructure sensor records for robust traffic environmental perception, bridging the digital-physical divide in transportation networks.
\end{itemize}

However, despite the early successes of pioneering studies in demonstrating the promise of integrating LLMs into transportation systems, they have explored isolated applications without articulating how the roles of LLMs interconnect throughout transportation systems. 
There lacks a unified framework and standard concept to guide systematic future progress in this area.
Therefore, we address this knowledge gap by introducing \textit{LLM4TR}, a conceptual framework where LLMs serve as polymorphic agents that dynamically adapt their roles to synergize sensing, learning, modeling, and managing tasks.

\textbf{Definition:} 
\textit{LLM4TR} refers to the methodological paradigm that systematically harnesses emergent capabilities of LLMs to enhance transportation tasks through four synergistic roles: transforming raw data into understandable insights, distilling domain-specific knowledge into computable structures, synthesizing adaptive system components, and orchestrating optimal decisions.

As {Fig.}~\ref{fig:intro} illustrates, the LLM4TR framework shapes the four-step transportation management cycle through an LLM-centric lens:
\begin{itemize}
\item \textbf{Processing information}: 
Beyond traditional sensor-based information collection, LLMs can serve as multimodal and comprehensive information processors that use natural language as interfaces.

\item \textbf{Encoding knowledge}: 
As foundation models for general domains, LLMs shift data-centric solutions for specific traffic tasks to knowledge-driven encoders that can learn from the context.

\item \textbf{Generating components}: 
Possessed with impressive generative power, LLMs automate the modeling and system design process by generating modular components following human-interpretable instructions.

\item \textbf{Facilitating decision making}: 
LLMs facilitate transportation management by providing principled guidance or making human-like decisions after activating their task execution capabilities.

\end{itemize}

\begin{figure}[!htbp]
  \centering
  \captionsetup{skip=1pt}
  \includegraphics[width=1\columnwidth]{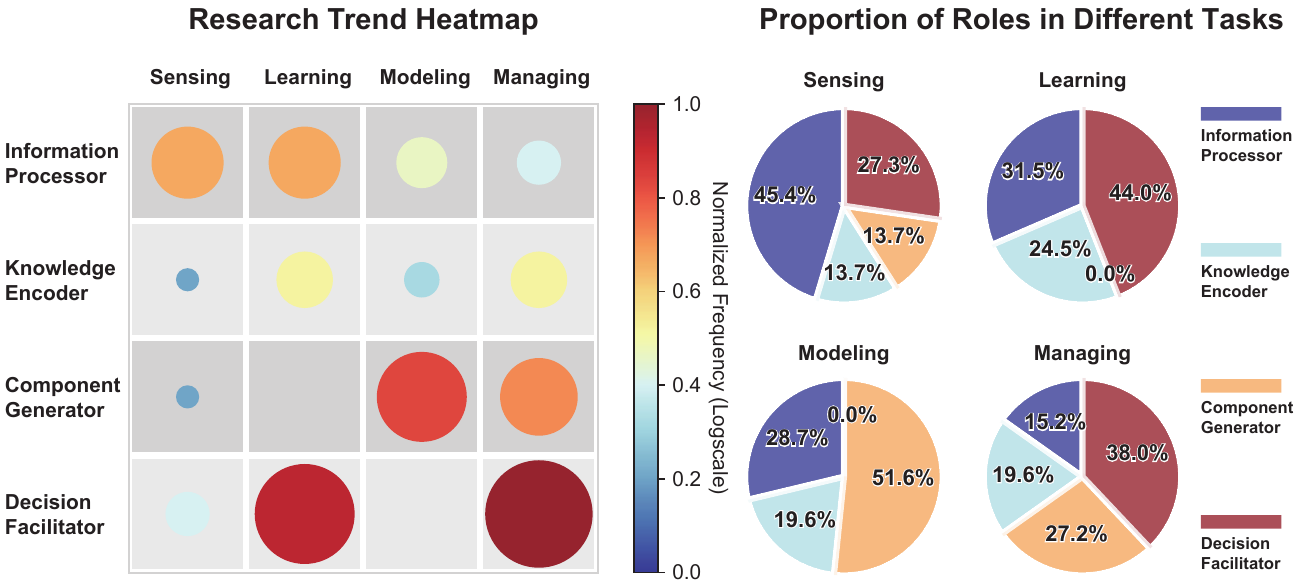}
  \caption{{Heatmap of the current research trend and pie chart of the proportion of the four roles of LLMs in different tasks.}}
  \label{fig:trend}
\end{figure}

These four roles intersect in the "four-step" traffic management strategy (will be discussed in Fig. \ref{fig:trend}).
Moreover, this framework transforms the conventional sequential pipeline by facilitating \textit{closed-loop synergy} among stages. Processed information is transmitted to knowledge encoders with contextualized data, which in turn informs the generation of adaptive models. These models enhance the managing process, and the outputs from managing decisions facilitate actions that refine subsequent sensing processes.
This creates a self-improving cycle through reflective feedback loops in which LLMs continuously align transportation operations with the evolving urban dynamics. 
Crucially, \textit{LLM4TR} shifts the paradigm from \textit{data-driven} to \textit{knowledge-and-data-driven} intelligence, where LLMs embed human-like reasoning into each management step. By unifying these roles, the framework establishes a roadmap for developing LLM-enhanced transportation solutions where language models evolve from auxiliary tools to central operators in system design and control.

\subsection{Taxonomy}
\begin{figure}[!htbp]
  \centering
  \includegraphics[width=1\columnwidth]{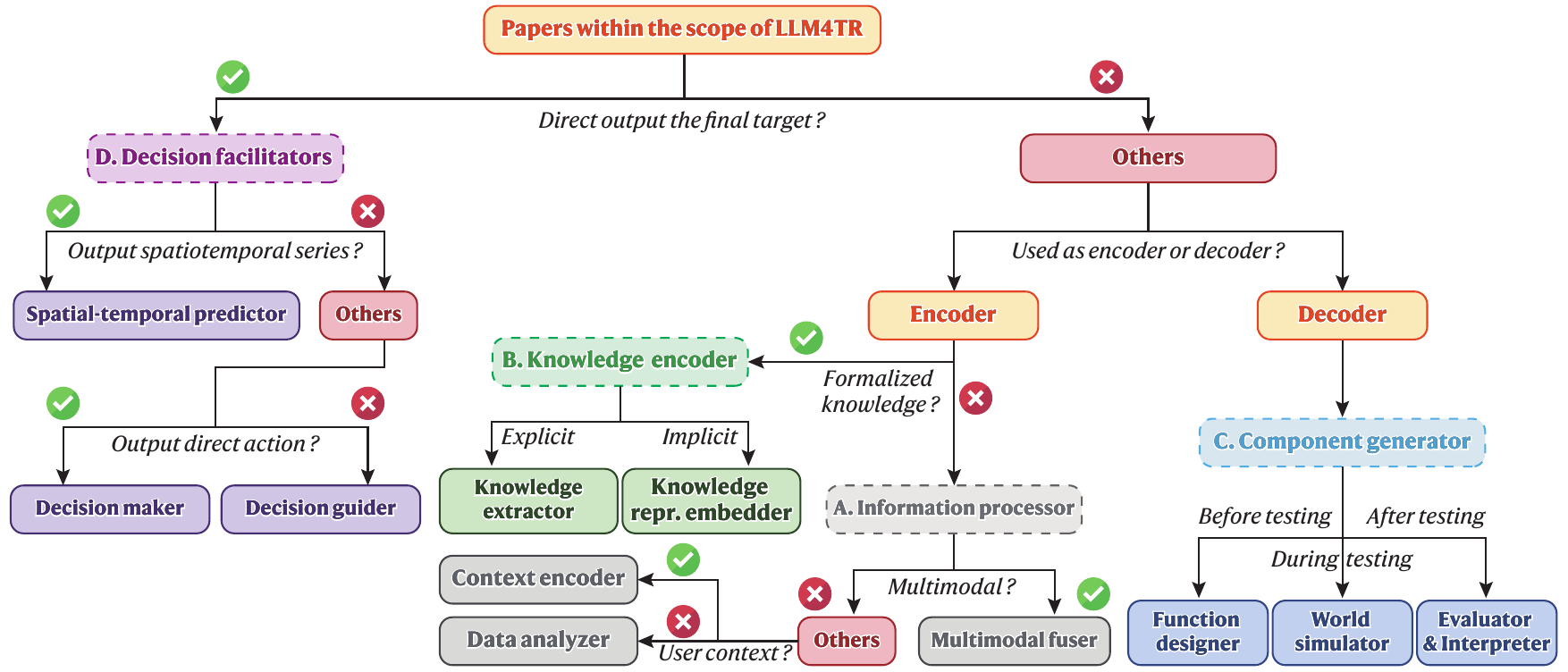}
  \caption{{\textbf{Taxonomy and classification principles in this survey.} Following this procedure, we classify the literature according to four primary roles: Information Processors, Knowledge Encoders, Component Generators, and Decision Facilitators. 
  Binary decision rules used to assign each method to exactly one primary role: 
  we begin by asking whether the model directly optimizes or outputs the final target (D. Decision Facilitator), which is further split into three subcategories according to the output type.
  If instead the focus is on whether LLMs are used as encoders or decoders. 
  If encoders, models that consume raw, unstructured inputs and process system information occupy A. Information Processor. Next, it is classified into three detailed roles according to the content of the input information.
  If not the case the focus is on formalizing that information into structured, domain-specific knowledge (explicit vs. implicit formalization), it is classified under B. Knowledge Encoder. 
  Finally, if LLMs as decoders synthesize new artifacts such as scenario simulations or network designs, it proceeds to C. Component Generator (differentiated by "before", "During" and “After” synthesis). 
  }}
  \label{fig:tree}
\end{figure}

Under the above framework, this section develops a structured and unified taxonomy to elaborate recent advances of LLMs in transportation. Inspired by the taxonomy in \citet{cao2024survey}, we survey the existing literature and summarize how LLMs are exploited to solve transportation problems from a methodological perspective, i.e., \textbf{the roles of LLMs in transportation systems}.
They generally include four aspects:

\begin{enumerate}
    \item \textbf{LLMs as information processors}
    \begin{itemize}
        \item \textit{Function:}  
        LLMs process and fuse heterogeneous transportation data from multiple sources (text, sensor data, task description, and user feedback) through contextual encoding, analytical reasoning, and multimodal integration. They enable unified processing of complex traffic patterns, parsing and integrating multi-source information to assist in the managing and semantic understanding of traffic data, reducing the complexity of downstream tasks.
        \item \textit{Example:} Using LLMs to analyze sensory traffic data \citep{zhang2024trafficgpt}, accident reports \citep{mumtarin2023large}, and convert user language queries to task-specific commands \citep{liao2024gpt}.
    \end{itemize}
    \item \textbf{LLMs as knowledge encoders}
    \begin{itemize}
        \item \textit{Function:} LLMs extract and formalize transportation domain knowledge from unstructured data through explicit rule extraction and latent semantic embedding. This role bridges the gap between the unstructured domain knowledge inherent in the data and computable (or comprehensible) representations for downstream applications.
        \item \textit{Example:} Building a knowledge base of traffic rules to assist traffic management \citep{wang2024transgpt}, formalizing traffic scenarios as knowledge graphs \citep{kuang2024harnessing}, and generating representation vectors that are applicable for subsequent computing \citep{he2024geolocation}.
    \end{itemize}
        \item \textbf{LLMs as component generators}
    \begin{itemize}
        \item \textit{Function:} 
        LLMs create functional algorithms, synthetic environments, and evaluation frameworks through instruction-followed content generation. This role utilizes generative capabilities of LLMs to automate the design, testing, and refinement of components in intelligent transportation systems. 
        \item \textit{Example:} Designing reward functions for traffic control agents in {RL} \citep{yu2024large}, assisting in synthesizing virtual driving environments \citep{zhao2024drivedreamer}, and providing feedback for the refinement of model component \citep{tian2024enhancing}.
    \end{itemize}
    \item \textbf{LLMs as decision facilitators}
    \begin{itemize}
        \item \textit{Function:} 
        LLMs predict traffic dynamics, optimize decisions, and simulate human-like reasoning, establishing new paradigms as generalized task solvers. 
        This role employs LLMs as predictive engines and decision facilitators for both micro-level agent behaviors and macro-level system states.
        \item \textit{Example:} Making control and planning decisions for autonomous driving \citep{sima2024drivelm}, guiding safety-critical actions \citep{wang2023accidentgpt}, and forecasting traffic states \citep{ren2024tpllm}.
    \end{itemize}    
\end{enumerate}

\textbf{Relationship and difference between the four roles of LLMs.}
This taxonomy reveals how LLMs go beyond traditional language processing to become versatile tools in transportation systems. For example, from raw data interpreters to central decision makers. 
Each role addresses different technical challenges while demonstrating synergetic effect when combined in integrated frameworks.
\textit{Information Processors} provide the fundamental data analysis that feeds \textit{Knowledge Encoders}, which in turn structure domain-specific insights. These structured insights then enable \textit{Component Generators} to produce context-aware simulations and algorithms, while \textit{Decision Facilitators} utilize both raw data and encoded knowledge for decision optimization. 
They collaboratively enhance the sensing, learning, modeling, and managing of transportation systems.

The key difference lies in their methodological focus: \textit{Information Processors} emphasize data transformation, \textit{Knowledge Encoders} focus on knowledge formalization, \textit{Component Generators} specialize in content synthesis, and \textit{Decision Facilitators} prioritize outcome and action prediction.
Generally, our taxonomy reflects the progress from data analytics (processing) to knowledge distillation (encoding), then to architecture design (generation), and finally to system implementation (decision).

Following this structured taxonomy, this survey classifies related literature using some principles shown in Fig.~\ref{fig:tree}.
{
Specifically, we classify each method according to the binary rules in Fig. \ref{fig:tree} and the following criteria:
\begin{enumerate}
    \item \textbf{Input Type}:
    \begin{itemize}
        \item \textit{Information Processors} consume raw or minimally preprocessed data (e.g. sensor streams, traffic logs, and user queries) and transform it into actionable formats for downstream processing.
        \item \textit{Knowledge Encoders} ingest processed representations (e.g., embeddings or metadata) to build domain‐specific knowledge.
    \end{itemize}
    \item \textbf{Output Form}:
    \begin{itemize}
        \item \textit{Component Generators} produce new artifacts such as textual description, synthetic data, code scripts, or model architectures that are tailored to downstream tasks.
        \item \textit{Decision Facilitators} generate actionable recommendations or predictions directly supporting operational decisions.
    \end{itemize}
    \item \textbf{Functional Objective}:
    \begin{itemize}
        \item Processing roles (\textit{Processors \& Encoders}) emphasize data understanding and knowledge structuring.
        \item Generation roles (\textit{Generators \& Facilitators}) emphasize content synthesis and action guidance.
    \end{itemize}
    \item \textbf{Pipeline Position}:
    \begin{itemize}
        \item \textit{Processors} appear at the front end of data pipelines.
        \item \textit{Encoders} capture and formalize insights for reuse after data processing.
        \item \textit{Generators} are placed within the model creation or solution design loops.
        \item \textit{Facilitators} locate at the final decision execution stage.
    \end{itemize}
\end{enumerate}

Please note that in some studies the role of LLMs are multifaceted. For example, they may serve simultaneously as both a context encoder and a decision maker. In our classification, we have identified and assigned each method to \textbf{the role that best reflects its primary functional contribution}, ensuring clarity and consistency.
}

\subsection{Research trend visualization}
As an intuitive overview of the current research trend and focus, we visualize the statistics of selected papers according to our taxonomy in Fig. \ref{fig:trend}. Several key findings in these figures can be summarized as follows: (1) For the four roles defined in this paper, the "decision facilitator" attracts the most interest in existing studies, where most of them focus on managing and learning tasks such as traffic prediction, autonomous driving, and traffic signal control. This is also one of the most crowded tracks in transportation research. (2) The "component generator" is also a popular direction in which the generative capabilities of LLMs can assist in designing modules or systems. (3) Some areas are still unexplored. For example, adopt LLMs as decision makers in traffic modeling and designers for learning tasks. The former can automate the traffic simulation procedure by acting as a decision engine, while the latter can inspire the neural architecture design for spatial-temporal predictive learning. (4) From another facet, different tasks highlight different functionalities of LLMs. For example, the sensing task populates the integration of {MLLMs} for information fusion; Among these four tasks, the application of the "knowledge encoder" accounts for the highest proportion in learning. These intuitive visual cues can guide future research choices and systematic overviews of this research field.

\section{LLMs as Information Processors}\label{sec:information_processor}
Central to ITS are data collection and analysis methods that integrate information from various sources, including sensors, cameras, and vehicle communication systems \citep{zhang2011data}. Machine learning algorithms, particularly deep learning models, are employed to process this data. However, this process may involve expert knowledge to develop problem-specific analytics methods.

\begin{figure}[!htbp]
  \centering
  \captionsetup{skip=1pt}
  \includegraphics[width=0.8\columnwidth]{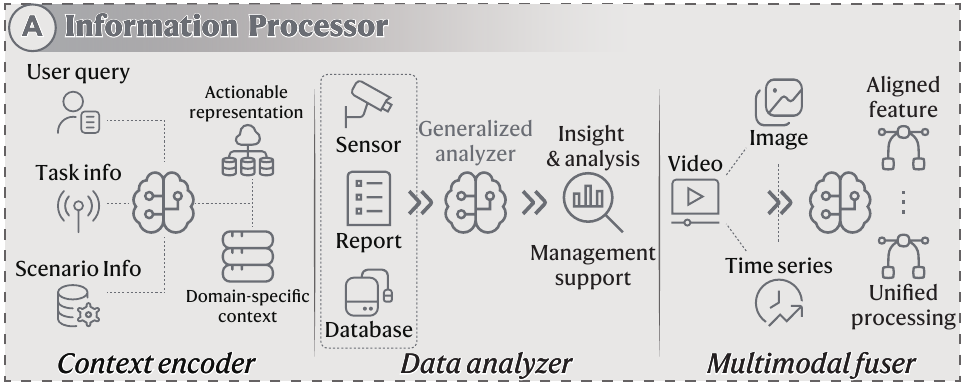}
  \caption{{LLMs as information processors.}}
  \label{fig:role1}
\end{figure}

Due to the {ICL} capability \citep{brown2020language} and the multimodal extension, LLMs have emerged as versatile tools for processing heterogeneous transportation data, enabling efficient encoding, analysis, and fusion of multimodal information generated by traffic participants. {As can be seen in Fig. \ref{fig:role1}}, this section explores three key methodological paradigms: (1) contextual encoding of traffic scenarios and task queries, (2) analytical reasoning over structured and unstructured traffic data, and (3) multimodal integration for holistic scenario understanding. 
These approaches collectively address challenges in handling heterogeneous information, multi-source data, and complex semantics in transportation systems, especially crucial for sensing and learning tasks.

\subsection{Context encoder}
A primary application of LLMs lies in encoding textual descriptions of dynamic traffic environments or task queries into machine-readable formats and task-specific patterns. 
This characteristic can facilitate the problem-solving process of ITS as LLMs can translate natural language instruction to domain-specific context that is compatible with other system components.
Studies such as \cite{keysan2023can}, \cite{zhao2024drivedreamer}, and \cite{liao2024gpt} demonstrate how LLMs transform free-form language descriptions of driving scenarios into encodings that can be directly consumed by downstream systems. Advanced implementations extend this capability to traffic simulation frameworks.
TP-GPT \citep{wang2024traffic} generates
accurate SQL queries for large-scale traffic databases and natural language interpretations by parsing user requests. It also employs a multi-agent collaboration strategy and few-shot learning to handle complex analytical tasks, such as traffic pattern recognition and privacy-preserving data interpretation. TP-GPT outperforms GPT-4 and PaLM 2 on a traffic analysis benchmark called TransQuery.
ChatSUMO \citep{li2024chatsumo} integrates LLMs with the SUMO traffic simulator to democratize traffic simulation for non-experts. The LLM (e.g., Llama3.1) processes natural language inputs and converts them into keywords, which trigger Python scripts to fetch OpenStreetMap data, generate road networks, and configure traffic conditions. 
By automating code generation and data interpretation, ChatSUMO reduces scenario creation time from 15 minutes to 30 seconds while achieving 96\% simulation accuracy. 

Such context encoders also enable customized scenario synthesis, as evidenced by ChatScene \citep{zhang2024chatscene} and \cite{ruan2024traffic}, which decompose high-level user instructions into parameterized CARLA simulator configurations.
\cite{zhang2024chatscene} introduce ChatScene, an agent that utilizes LLMs to generate safety-critical scenarios. By processing unstructured language instructions, ChatScene first creates textual descriptions of traffic scenarios. These descriptions are then decomposed into detailed sub-descriptions specifying vehicle behaviors and locations, enabling the generation of varied and safety-critical driving scenarios.
\cite{ruan2024traffic} adopt LLMs to generate diverse traffic scenes in the CARLA simulator from natural language contexts. LLMs are used to decompose user prompts into road conditions, agent types, and actions, retrieve candidate roads from a structured database, and plan agent behaviors. The LLM-based parser addresses the limitations of predefined scenarios by dynamically generating customizable traffic scenes.

The contextual encoding paradigm also extends to real-time situational awareness through hybrid architectures. \cite{xue2022leveraging} employs BERT as text encoder to generate feature embeddings for both contextual and numerical tokens. These embeddings are used to predict customer flows at Places-of-Interest (POIs). In \citep{abdelrahman2024video}, the authors combine computer vision techniques with LLMs to analyze pedestrian activities at intersections while preserving privacy. They convert raw video feeds into anonymized text descriptors using a vision encoder, then feed these descriptions into an LLM for contextual reasoning. This approach achieves real-time pedestrian behavior prediction without storing sensitive visual data.

\subsection{Data analyzer}
Beyond situational encoding, LLMs excel at extracting understandable patterns and insights from complex transportation datasets. Traditional methods of analyzing traffic data often involve domain-specific tools such as statistical methods and expert knowledge \citep{zhu2018big,washington2020statistical}, making the insights dependent on specific problems. The generic pattern understanding ability of LLMs enables them to become generalized traffic data analyzer.   
TrafficGPT proposed by \citep{zhang2024trafficgpt} demonstrates zero-shot analytical capabilities to analyze traffic data and provide insightful support for transportation management systems.
CrashLLM \citep{fan2024learning} employs LLMs to analyze traffic crash data. By treating crash event feature learning as a text reasoning problem, CrashLLM fine-tunes various LLMs to predict fine-grained accident outcomes, such as crash types, severity, and injury numbers. It also enables nuanced analysis of contributing factors and supports what-if situational awareness traffic safety analyzes.

The natural language understanding capabilities of LLMs prove particularly valuable for mining unstructured data. 
They have potential for traffic safety analysis using textual materials such as accident report \citep{zheng2023chatgpt}.
In \citep{mumtarin2023large}, the authors query LLMs to extract related information
and answer questions related to accidents from 100 crash narratives from Iowa and Kansas.
In \citep{arteaga2025large}, the authors use LLMs to identify underreported alcohol involvement in traffic crash narratives. Through iterative prompt engineering, the LLMs parse unstructured crash reports into binary classifications. The method achieves up to 1.0 recall and 0.93 precision, outperforming traditional text mining approaches.

Beyond text data, vision-language models
(VLMs) and multimodal language models (MLMs) are also applied to understanding traffic scenarios from heterogeneous data sources \citep{wen2023road,qasemi2023traffic,zhou2024gpt,lohner2024enhancing,you2024v2x,tong2024connectgpt,esteban2025scenario}.
These models can also be integrated with external knowledge for better understanding. For example, recent studies address data scarcity through retrieval-augmented architectures. 
RAG-Driver \citep{yuan2024rag} integrates {ICL} with a retrieval-augmentation mechanism that dynamically retrieves expert demonstrations from a database of past driving scenarios. These retrieved examples, encoded as video control signal pairs with natural language explanations, are prefixed to the input query to guide the MLM in generating human-readable driving action descriptions and justifications. 
Evaluations on benchmarks like BDD-X underscore the potential of combining parametric memory with external knowledge bases for robust traffic analysis.

\subsection{Multimodal fuser}\label{sec:Multimodal fuser}

MLLMs bridge the gap between heterogeneous data fusions in transportation systems. MLLMs can be incorporated into multimodal ITS by converting diverse types of data into aligned feature vectors or unified processing. This is often achieved by the cross-attention computation of the Transformer architecture \citep{vaswani2017attention}.
\cite{zheng2023chatgpt} discuss how LLMs can address key traffic safety issues, such as automating accident report generation, augmenting traffic data, and analyzing multisensory data. 
To mitigate these challenges, the paper proposes the concept of multi-modality representation learning, which integrates data from various sources to improve traffic safety analytics. 
While \cite{abu2024using} employs MLLMs such as Gemini-Pro-Vision and LLaVa to detect safety-critical events in driving videos. The framework fuses textual, visual, and audio inputs in a unified way through context-specific prompts to minimize hallucinations and improve reliability. A zero-shot learning approach is used to adapt the model to diverse scenarios without extensive retraining, addressing the limitations of traditional ML models that rely on annotated datasets. 
The SeeUnsafe framework \citep{zhang2025language} uses MLLMs to automate video-based traffic accident analysis. It replaces traditional "extract-then-explain" workflows with an interactive conversational approach, where the MLLM classifies accidents, grounds visual elements, and generates structured reports. Multimodal prompts are designed to align textual queries with visual data. Tested on the Toyota Woven Traffic Safety dataset, SeeUnsafe demonstrates improved processing throughput and adaptability.

\subsection{Summary and outlook}
As information processors, LLMs are revolutionizing transportation information systems through three interrelated capabilities: semantically rich context encoding, data-driven analytical reasoning, and robust multimodal fusion. From parsing scenario synthesis tasks in CARLA \citep{ruan2024traffic} to predicting crash severity through narrative analysis \citep{fan2024learning}, these approaches demonstrate remarkable adaptability across data modalities and application domains. 
In summary, current studies establish a foundation for LLM-powered transportation information processing tools that balance automation with interpretability.
{However, several critical limitations persist in current research that constrain the broader applicability of them.
First, most existing implementations rely heavily on benchmark datasets or constrained simulation settings, limiting their robustness in real-world, noisy, or out-of-distribution environments. For instance, context encoders often assume grammatically correct and unambiguous user input, which may not reflect field deployment conditions. Furthermore, multimodal fusion approaches typically focus on aligning no more than two data modalities and lack comprehensive evaluations on diverse and real-time traffic data streams.
Second, many proposed models exhibit a strong dependence on prompt engineering or fine-tuning strategies, leading to reproducibility and generalization concerns. The lack of standardized evaluation protocols for transportation-specific tasks exacerbates this issue. Despite the progress of frameworks like ChatSUMO and SeeUnsafe, their underlying LLMs are still vulnerable to hallucinations and fail to guarantee interpretability in high-stakes scenarios such as accident analysis or transport planning.

}

We list several potential directions for future explorations in the following:
\begin{enumerate}
    \item \textbf{Unified multi-source and cross-domain information fusion:} Although the use of MLLMs has shown promise in dealing with data or user queries from different data modalities, most of existing studies only consider the integration of information for simple use cases.
    The power of cross-domain data fusion from diverse sources (e.g., geographic, traffic, social media, and environmental data) and modalities (e.g., spatio-temporal, visual, and textual modalities) has not yet been fully exploited \citep{zou2025deep}.
    \item \textbf{Integrating external information for better response:} Since the capabilities of LLMs largely depend on pretraining data, they may not have specific knowledge of complex transportation problems themselves.
    {RAG \citep{guu2020retrieval}} provides a feasible approach for incorporating external knowledge or data base for more professional responses.
    \item {\textbf{Robust multimodal aligning in real-world scenarios:} Future work should investigate LLM-based architectures capable of operating under noisy, incomplete, or ambiguous real-world inputs. This may require robust data alignment, continual learning, and probabilistic inference mechanisms that can respond to dynamic and uncertain environments.}
\end{enumerate}

\section{LLMs as Knowledge Encoders}\label{sec:knowledge_encoder}
\begin{figure}[!htbp]
  \centering
  \captionsetup{skip=1pt}
  \includegraphics[width=0.8\columnwidth]{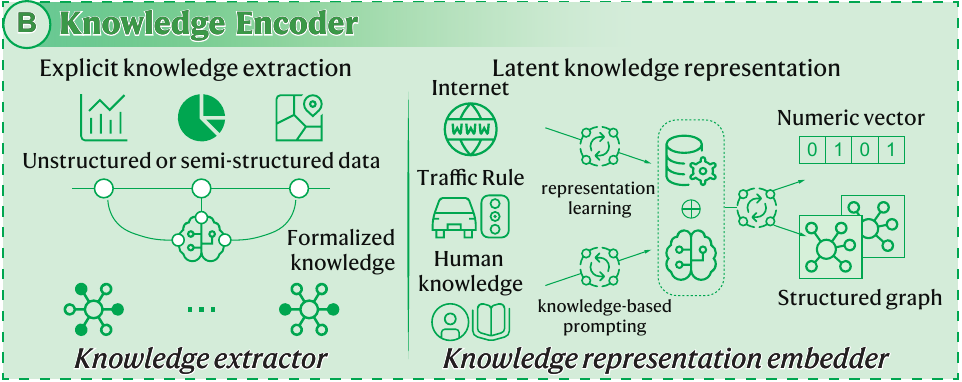}
  \caption{{LLMs as knowledge encoders.}}
  \label{fig:role2}
\end{figure}

LLMs can serve as repositories of implicit and explicit world knowledge \citep{jiang2020can,gurnee2023language}, enabling structured extraction, encoding, and application of domain-specific insights for transportation problems. This capability stems from their pretraining on vast corpora that contain traffic regulations, geographical semantics, and behavioral patterns. {Fig. \ref{fig:role2}} categorizes these studies of using LLMs to encode knowledge into: (1) explicit knowledge extraction for structured reasoning and (2) latent knowledge representation through embedding spaces.
These studies demonstrate how to extract useful knowledge from the large parameter space of LLMs to facilitate the solving of traffic problems.

\subsection{Knowledge extractor}
In addition to analyzing the patterns underlying the data (i.e., data analyzer), methods in this category further manipulate LLMs to explicitly distill unstructured or semi-structured data into formalized and structured knowledge representations such as text and knowledge graphs that allow systematic reasoning about transportation domains. 
A pioneering approach by 
\cite{kuang2024harnessing} introduced a framework that extracts common traffic knowledge from scene images using the Llava-7b vision-language model and generates visual traffic knowledge graphs by organizing the scene's information and reflecting relationships
between traffic elements. 
This structured output enables downstream applications like conflict prediction.

Specialized LLMs further enhance domain-specific knowledge retention. 
TransGPT \citep{wang2024transgpt} introduces a specialized LLM to serve as a transportation knowledge base, including two variants: TransGPT-SM for single-modal data and TransGPT-MM for multimodal data. TransGPT-SM is fine-tuned on textual transportation datasets to address tasks like traffic analysis and recommendation generation. TransGPT-MM extends this by incorporating both textual and visual data, handling tasks such as explaining traffic phenomena and answering traffic-related questions. 
Similarly, TrafficSafetyGPT \citep{zheng2023trafficsafetygpt} grounds LLMs in safety-critical contexts through {SFT} on the TrafficSafety-2K dataset, which is a curated corpus combining government-produced guidebooks and ChatGPT-generated instruction-output pairs.
This alignment enables precise identification of regulatory violations in accident reports compared to general-purpose LLMs.
These two methods explicitly convert LLMs to transportation knowledge bases.

Operational knowledge synthesis is exemplified by IncidentResponseGPT \citep{grigorev2024incidentresponsegpt}, 
which uses LLMs to create incident response plans based on traffic incident reports and regional response guidelines. It synthesizes these guidelines into a structured form using LLMs, and then combines them with real-time incident data to generate tailored, actionable traffic incident plans.
Emerging frameworks such as TARGET \citep{deng2023target} extend this paradigm by employing LLMs to automatically generate test scenarios by encoding traffic rules into a structured domain-specific language. In the TARGET framework, the LLM parses natural language traffic rules to extract key traffic components. 
By transforming unstructured rules into formal scenario representations, the LLM enables automated synthesis of executable test scripts in simulators like CARLA. This approach detected over 700 violations across four autonomous driving systems, demonstrating its effectiveness in translating regulatory knowledge into actionable testing scenarios while bypassing manual DSL coding.

\subsection{Knowledge representation embedder}
Beyond explicit knowledge extraction, LLMs encode transportation semantics into dense vector spaces that capture latent relationships between entities and scenarios. 
This is usually achieved by converting the contextual knowledge into computable format such as embedding vectors. In the embedding methodology, LLMs are strategically employed as encoder architectures that transform input queries into semantically rich vector representations. Unlike conventional approaches generating textual or numerical output, this paradigm specifically produces high-dimensional embedding vectors, which subsequently serve as input features for subsequent computational processes.

Traditionally, language models such as BERT are used to extract representations from unstructured text input.
For instance, \cite{das2023classifying} applies the BERT model to classify pedestrian maneuvers from unstructured police crash narratives. The authors fine-tuned BERT-base on Texas crash data for binary (intentionality) and multiclass (maneuver type) tasks. Text preprocessing included tokenization, lowercasing, and truncation to 512 tokens, with embeddings fed into a linear classification layer. 

With the advent of LLMs, an emerging approach is to directly use pretrained LLMs to extract embeddings from the text prompt \citep{he2024geolocation,nie2025joint}. Specifically, \cite{he2024geolocation} attempt to elicit the inherent geospatial knowledge from pretrained LLaMA-3 model. A structured prompt that describes basic geolocation information of the POI is first derived by retrieving OpenStreetMap. Then the prompt is fed to LLMs to obtain a continuous vector. Such vectors are demonstrated to be effective in improving spatial-temporal forecasting models, including traffic prediction.
\cite{nie2025joint} further extends this approach using a linear adapter and integrates it into an expert {GNN} predictor for city-wide traffic demand estimation. Evaluations also reveal the zero-shot transferability of such embeddings.

Similar principles of the LLM-based embedding approach have recently been explored in application domains.
ALT-Pilot \citep{omama2023alt} uses LLMs for autonomous navigation. 
Using multimodal data from LiDAR and cameras onboard, combined with {VLMs} (e.g., CLIP), the system enhances vehicle localization. LLMs are used to generate the embedding of language-based landmarks in the environment, enabling open-vocabulary navigation.
By encoding environmental features into a shared embedding space, the system reduces localization errors in unmapped urban areas.
Language-conditioned embedding techniques further enhance scenario generation and prediction tasks. LCTGen \citep{tan2023language} utilizes LLMs to convert textual descriptions of an expected scenario into structured scene vectors that condition a Transformer model to produce realistic traffic behaviors. 
This alignment between language embeddings and spatiotemporal features allows generation of critical scenarios. 
Similarly, \cite{zheng2024large} integrates GPT-4V-derived embeddings into motion forecasting through Transportation Context Maps. By encoding global scene semantics as weighted one-hot vectors fused via cross-attention, their Motion Transformer model achieves a 0.95\% mAP gain on the Waymo dataset, demonstrating that LLM-enhanced embeddings improve trajectory prediction by contextualizing local agent behaviors within holistic scene understanding.

\subsection{Summary and outlook}
As knowledge encoders, LLMs transform transportation data through dual mechanisms: structuring explicit domain knowledge (e.g., VTKGs in \cite{kuang2024harnessing}, protocol templates in \cite{grigorev2024incidentresponsegpt}) and encoding latent semantics into reusable embeddings (e.g., scenario vectors of LCTGen \citep{tan2023language}, LLM2Geovec in \cite{he2024geolocation}). Although specialized models such as TransGPT \citep{wang2024transgpt} demonstrate the value of domain adaptation, challenges persist to maintain the freshness of knowledge. For example, incident response plans can become outdated as regulations evolve, and navigation embeddings like in ALT-Pilot \citep{omama2023alt} require continuous map updates. Nevertheless, the fusion of parametric knowledge (learned during pretraining or fine-tuning) and nonparametric representations suggests a path toward applying LLMs that dynamically internalize transportation knowledge while remaining grounded in real-world traffic problems.
{However, key limitations remain unaddressed. First, static fine-tuning or prompt-based extraction methods cannot keep up with evolving traffic regulations, infrastructure changes, or emerging mobility patterns. For example, an incident plan generated by IncidentResponseGPT can quickly become outdated when local response guidelines are updated. There is no standard mechanism to incrementally update embeddings or graphs as new data arrives.
Second, there is no unified benchmark or tool to measure knowledge quality (e.g., correctness, coverage) or embedding utility across different transportation tasks. Finally, interpretability of learned representations is also underexplored, hindering trust and adoption by domain experts.
}

We list several potential directions for future explorations in the following:
\begin{enumerate}
    \item \textbf{Improving representation quality through alignment tuning:}
    One of the limitations of using frozen pretrained LLMs to encoding knowledge is that the representation embedding is fixed and cannot be adaptively refined to tasks. A possible solution is to use {SFT} \citep{wang2023aligning} to improve the quality and specificity of the representations and further align them with human preference.
    \item \textbf{Contrastive knowledge representation learning:} Most of the existing approaches obtain the knowledge embedding using pure text modality. As discussed in section \ref{sec:Multimodal fuser}, traffic data is usually multimodal and structured knowledge may exist beyond text.
    Contrastive language-image pretraining (CLIP) \citep{radford2021learning} provides a promising architecture for learning hybrid representations.
    \item \textbf{Up-to-date knowledge integration via online search:} 
    The inherent knowledge of LLMs is confined to the information available up to their last training update. This condition limits their access to the most recent data, e.g., real-time traffic condition.
    Integration of LLMs with web search tools such as WebGPT \citep{nakano2021webgpt} uses the search engine to find unknown knowledge, alleviating the limitation.
\end{enumerate}

\section{LLMs as Component Generators}\label{sec:component_generator}
\begin{figure}[!htbp]
  \centering
  \captionsetup{skip=1pt}
  \includegraphics[width=0.8\columnwidth]{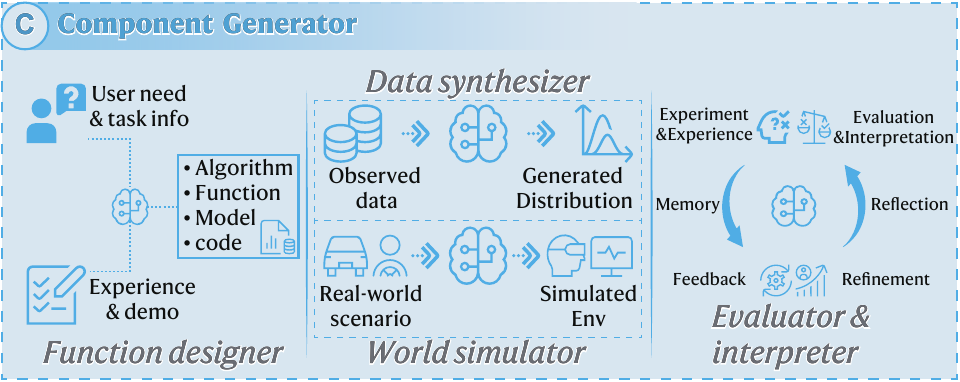}
  \caption{{LLMs as component generators.}}
  \label{fig:role3}
\end{figure}

One of the most impressive aspects of LLMs is their generative capability. They can generate high-quality content according to user instructions, which is skillful for solving transportation problems.
In this context, LLMs can emerge as powerful generative engines for components in ITS, enabling the automated creation of functional components, synthetic environments, and evaluative frameworks. 
As shown in Fig. \ref{fig:role3}, this section examines four critical generative paradigms: (1) algorithmic function design, (2) world simulation, (3) data synthesis, and (4) system evaluation. We systematically demonstrate how LLMs transcend traditional language modeling to become active creators of intelligent tools.

\subsection{Function designer}
Benefiting from pertained tasks such autoregressive language modeling in Eq. \ref{eq:lm} and the {ICL} ability from textual demonstration, LLMs specialize in the design or refinement of code- or rule-based functions for traffic management. Such functions like the reward function in {RL} are difficult to design manually and require expert guidance.
LLMs revolutionize algorithmic design by translating natural language specifications into executable functions or codes.

A straightforward approach is to prompt LLMs directly and provide several examples of desirable properties and behaviors.
InteractTraj \citep{xia2024language} exemplifies this by interpreting natural language descriptions into interactive traffic trajectories. It employs a language-to-code encoder with an interaction-aware encoding strategy to process language descriptions into formatted numerical codes. A code-to-trajectory decoder with interaction-aware feature aggregation then maps these codes to final interactive trajectories, considering vehicle interactions, environmental maps, and vehicle movements. \cite{zhong2023language} introduce CTG++, a scene-level conditional diffusion model that leverages LLMs to convert natural language instructions into differentiable loss functions for traffic simulation. The LLM translates user queries like "simulate a traffic jam" into code-based differentiable loss functions that guide the diffusion process during denoising.

This paradigm extends to RL systems through innovative reward and policy engineering. 
LLMs can generate executable codes of reward function explicitly and specify the details of the computing process \citep{ma2023eureka}.
Autoreward proposed by \cite{han2024autoreward}
designs reward functions for RL-based autonomous driving. Instead of ambiguous desired goals, it employs concrete undesired linguistic goals to compute rewards. The agent’s state and the undesired goal are embedded using pretrained models, and the cosine distance between these embeddings serves as the reward signal.
Similarly, \cite{yu2024large} leverage LLMs to automate reward function design for RL-based bus holding strategies. Four LLM modules (reward initializer, modifier, analyzer, refiner) interact to generate dense rewards from sparse objectives. The LLM converts domain knowledge (e.g., headway balancing, passenger demand) into code-based reward functions, iteratively refining them using training performance feedback. A refiner module filters ineffective rewards to ensure stability. \cite{villarreal2023can} instead investigate the use of ChatGPT to help design RL policies for mixed traffic control. ChatGPT translates user prompts into RL-aligned metrics and suggests creative reward functions. Participants without RL expertise used ChatGPT to formulate Markov Decision Process (MDP) components for traffic scenarios. ChatGPT-assisted users achieved 150\% and 136\% increases in successful policies for intersection and bottleneck environments compared to non-assisted novices, even outperforming experts in some cases.

Collaborative design frameworks combined with more general design targets push these boundaries further. 
In LearningFlow \citep{peng2025learningflow}, multiple LLM agents collaborate to automate curriculum learning and reward design. The framework employs a curriculum analysis agent to evaluate training progress and a generation agent to iteratively produce tailored training curricula and reward functions. The LLMs are prompted with contextual descriptors of driving tasks and historical training data stored in a memory module, enabling dynamic adaptation. 
OminiTester introduced by \cite{lu2024multimodal} uses MLLMs to generate realistic and diverse corner cases to test autonomous vehicles. The approach integrates tools from SUMO to simplify the complexity of code generation by LLMs. 
Additionally, RAG is used to enhance scenario realism by grounding LLM outputs in crash reports or historical data. 
A self-improvement mechanism iteratively refines scenarios based on simulation feedback. 
\cite{mei2025llm} introduces a closed-loop framework where LLMs identify adversarial vehicles and optimize their trajectories to test autonomous driving systems. Three LLM modules, including Initialization, Reflection, and Modification, collaborate to generate executable code for attacker identification. The LLM iteratively refines attack strategies using feedback from simulation results and employs techniques like CoT prompting and Best-of-N sampling to enhance code quality. The generated adversarial scenarios are used to train RL-based policies, forming a feedback loop to improve robustness.

In addition to the explicit design of the reward code, LLMs / VLMs are themselves the reward model \citep{huang2024vlm,gao2024vista}, which is studied as an \textit{implicit reward model} \citep{kwon2023reward,rafailov2023direct}.
They can provide an overall reward value based on the understanding of tasks and environments or score the alignment between feature embeddings of visual observations and language instructions. 
By integrating LLMs into the RL loop, the method reduces reliance on manual reward engineering and improves sample efficiency.

\subsection{World simulator}
LLMs are equipped with extensive common sense and world knowledge. They are assumed to have a basic understanding of the regularity of the world, such as space and time \citep{gurnee2023language}. This makes LLMs possible for assisting in simulating the environmental dynamics of real-world driving scenarios. Such generalized simulators are termed \textit{world models} \citep{ding2024understanding} that can learn to predict the future state of the environment with high fidelity, especially crucial for the evaluation of end-to-end autonomous driving systems \citep{feng2025survey}.

In autonomous driving, large-scale pretrained generative models are increasingly utilized as world models to generate realistic images and video sequences of driving scenarios, thus enhancing training and evaluation of autonomous systems \citep{guan2024world,feng2025survey}. 
These world models offer detailed representations of the driving environment by combining data from multiple sensors, semantic information, and temporal dynamics. 
Theys can learn world model dynamics for autonomous driving systems from action-free video demonstrations and additional conditions.
By integrating perception, prediction, and planning, world models allow autonomous systems to respond quickly and intelligently to complex and often unpredictable situations in a closed-loop manner \citep{gao2023magicdrive,hu2023gaia,wang2024drivedreamer,wang2024driving,gao2024vista,yang2024generalized,zheng2024doe,fu2024drivegenvlm}.

As synthetic environment generators, pioneering methods apply vision generative models to create photorealistic driving scenarios with precise controllability. MagicDrive \citep{gao2023magicdrive} pioneers this by generating high-fidelity streetview images and videos with precise 3D geometry control. It integrates various control signals such as camera poses, road maps, 3D geometry, and textual descriptions to generate diverse and realistic scenarios.
The consistency across different camera perspectives is achieved through a cross-view attention module.
Drive-WM \citep{wang2024driving} advances this by introducing a multiview world model features joint spatial-temporal view factorization. It is capable of generating high-quality, controllable, and consistent multiview videos in driving scenes. 
WoVoGen \citep{lu2024wovogen} addresses the challenge of generating multi-camera street-view videos by incorporating a world volume-aware diffusion model. This approach ensures that the generated videos maintain both intra-world consistency and inter-sensor coherence. 

The integration of LLMs with physics engines (world models) yields unprecedented scenario customization. For example, DriveDreamer-2 \citep{zhao2024drivedreamer} employs an LLM interface to convert user queries into agent trajectories, which are then used to generate high-definition maps adhering to traffic regulations.
The versatility of LLMs enable the world model to generate customized driving videos from user's textual prompt, including uncommon scenarios like vehicles abruptly cutting in. 
DriveMM \citep{huang2024drivemm} integrates LLMs into world models by developing a large multimodal model that synthesizes heterogeneous inputs to simulate dynamic driving environments and generate actionable outputs.
It demonstrates that combining LLMs with multimodal data and structured training pipelines can produce world simulators for context-aware driving.

\subsection{Data synthesizer}
High-quality and meaningful data are valuable for the applications of data-centric ITS. Since the backbone Transformer architecture enables deep data interaction through the self-attention mechanism, LLMs can address data scarcity through generative synthesis of system parameters and data engineering.

The first use case is to synthesize system parameters or parameterized traffic scenarios.
\cite{chang2024llmscenario} proposed a framework that leverages LLMs to generate parameters for safety-critical traffic scenarios, particularly rare corner cases, called LLMScenario. LLMScenario involves three stages: scenario prompt engineering, LLM-driven parameter generation, and evaluation feedback tuning. The LLM translates textual descriptions into actionable parameters for traffic simulations, addressing the challenge of generating diverse and realistic rare cases efficiently. 
SeGPT \citep{li2024chatgpt} leverages ChatGPT to parse user queries and synthesize parameterized scenarios, including vehicle trajectories and environmental conditions. By combining CoT prompting with domain-specific templates, SeGPT produces complex interactions that improve the robustness of prediction algorithms. This method addresses the data scarcity issue in autonomous vehicle testing by automating scenario creation without manual annotation.
Beyond scenario generation, LLMs optimize system dynamics modeling by generating new parameters. 
\cite{da2024prompt} leverage the knowledge of LLMs to understand and profile the system dynamics by a prompt-based grounded
action transform in traffic control systems. They exploit LLMs to infer how traffic dynamics change with weather conditions, traffic states, and road types. LLMs then synthesize new parameters for system dynamics. The policies’ action is taken and grounded based on generated dynamics, thus helping the agent learn a more realistic policy.

For the role in traffic data engineering, 
TransCompressor \citep{yang2024transcompressor} demonstrates LLMs' dual role in traffic data synthesis.
It employs GPT-4 for zero-shot compression and reconstruction of multimodal transportation sensor data. 
The LLM reconstructs data via minimalist prompts, eliminating the need for fine-tuning. Evaluated on bus, taxi, and MTR scenarios, it achieves high reconstruction accuracy, addressing storage and transmission inefficiencies.
LLMs can also be applied to synthesize features in latent spaces. For example, \cite{wang2024drive} integrates multi-modal foundation models to address the challenges of open-set generalization in autonomous driving. 
The model employs latent space simulation for data augmentation, where text-based prompts dynamically adjust the policy’s response to rare or unseen driving conditions. This approach enhancing robustness to out-of-distribution environments.

\subsection{Evaluator and interpreter}
After adopting LLMs as generators for functions or systems, the quality of the generation result needs to be evaluated before it is brought into service. This process typically requires human evaluations and explainable AI (XAI) techniques \citep{dwivedi2023explainable}.
Fortunately, LLMs can bring human-like reasoning to system evaluation and decision self-refinement, simplifying the traditional procedure.

LLMs can be prompted to generate human-readable language interpretations and evaluations of current results directly. The evaluations based on the decision trajectory or performance record can further improve the quality of the generator.
CRITICAL \citep{tian2024enhancing} is a framework designed to improve the training of autonomous vehicles by generating critical driving scenarios for RL agents. CRITICAL uses LLMs to interpret RL training episodes to evaluate failure patterns, suggesting modifications based on analysis of traffic dynamics and risk metrics. This closed-loop system refines AV behavior by continuously feeding back critical scenarios, improving training performance and safety resilience. 
In \citep{lin2024drplanner}, LLMs were leveraged to provide feedback for current driving policies based on performance on the leaderboard. LLMs provide refinement suggestions for both rule- and optimization-based policies by regenerating objective or heuristic functions.
Similarly, \cite{chen2023feedback} use ChatGPT to provide feedback on architectural choices for driving agents.

This paradigm also extends to other tasks such as traffic control.
\cite{pang2024illm} introduced iLLM-TSC, a hybrid framework integrating RL and LLMs to enhance traffic signal control. The proposed method employs a two-step process: (1) An RL agent first generates preliminary signal control decisions based on real-time traffic observations. (2) An LLM then evaluates these decisions for reasonableness, refining them using contextual knowledge and compensating for gaps in state information. The LLM acts as a corrective layer, dynamically adjusting actions through prompt engineering to align with real-world constraints and safety priorities.

\subsection{Summary and outlook}
As generators, LLMs transform transportation systems through three key capabilities: 1) translating abstract requirements into functional algorithms \citep{xia2024language,han2024autoreward}, 2) synchronizing photorealistic environments with controllable dynamics \citep{zhao2024drivedreamer, gao2023magicdrive}, and 3) providing interpretable system evaluations for self-refinement \citep{tian2024enhancing, pang2024illm}. Maintaining physical fidelity in the generated results is crucial for real-world deployment.
For example, while the generative world models are impressive in simulating traffic scenarios, they still trail real-world data distributions. However, the integration of LLMs with simulation engines and self-refinement mechanisms points to a future where ITS can self-generate their training ecosystems while maintaining alignment with physical and regulatory constraints.
{Despite these advances, important limitations constrain their deployment. First, generated functions and synthetic environments often diverge from real-world physics or omit critical semantics. For example, world models may misrepresent vehicle dynamics under extreme maneuvers, and synthetic scenarios can lack corner-case diversity beyond prompt scope. Second, high-quality generation, especially photorealistic video or complex code, can incur prohibitive inference costs and latency, limiting real-time or embedded use. Third, without rigorous validation, the generated components can encode unsafe behaviors or reflect biases from training data, posing risks in safety-critical contexts.
}

We list several potential directions for future explorations in the following:
\begin{enumerate}
    \item \textbf{Inspiring novel neural architecture designs}: Adopting appropriate neural network architectures often requires a lot of empirical experience. However, practitioners in traffic engineering often lack sufficient practice of deep learning projects, especially as current models become increasingly complex. The automatic architecture search and design utility of LLMs \citep{nasir2024llmatic} can facilitate this process. 
    \item \textbf{Large-scale simulation with LLM-based agents}: Large-scale LLM-based agent simulation harnesses the context-aware behavioral plasticity of LLMs to simulate complex behaviors in human-centric systems \citep{li2023metaagents,gao2024large}. 
    This offers great potential in large-scale traffic simulation, automated testing, and the development of traffic digital twins. For example, using LLMs for interaction modeling through latent social psychology simulation can enable naturalistic representation of crowd dynamics in multimodal hubs and self-organized traffic patterns during infrastructure failures. 
    \item \textbf{Identifying and alleviating inherent bias of LLMs}: The inherent bias of LLMs can be converted to the generated functions or data \citep{yu2023large}, which can be harmful for safety-critical applications such as autonomous driving. Human alignment techniques such as RLHF \citep{ziegler2019fine} can adjust LLMs to comply with human preferences, ethical guidelines, and safety standards.
    \item {\textbf{Physics-aware generation pipelines.}  Incorporate differentiable physics engines or domain constraints into the generation loop, e.g., embedding vehicle dynamics simulators into LLM prompting, to ensure physically plausible code and components.}

\end{enumerate}

\section{LLMs as Decision Facilitators}\label{sec:decision_facilatator}
\begin{figure}[!htbp]
  \centering
  \captionsetup{skip=1pt}
  \includegraphics[width=0.8\columnwidth]{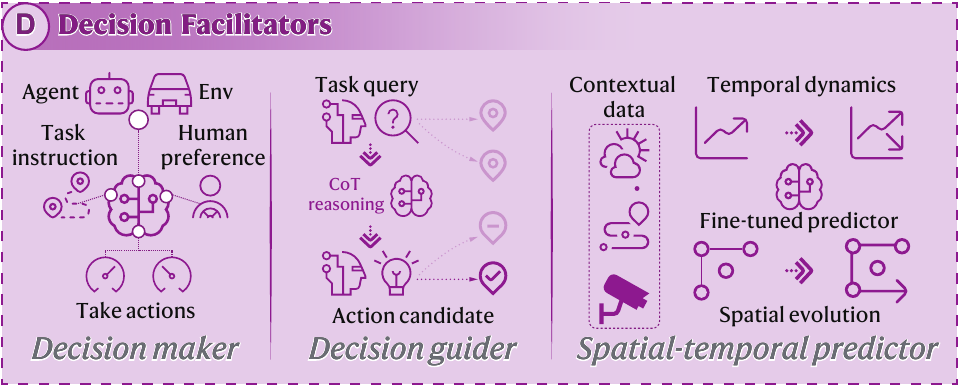}
  \caption{{LLMs as decision facilitators.}}
  \label{fig:role4}
\end{figure}

Beyond the language modeling abilities, LLMs can serve as generalized problem solvers by step-by-step reasoning, task planning, and tool manipulation.
Recent advances have positioned LLMs as powerful decision facilitators in transportation systems, capable of simulating human-like reasoning to forecast outcomes, optimize decisions, and adapt to unseen scenarios. 
As shown in Fig. \ref{fig:role4}, this section examines their predictive roles across decision-making, guidance, and spatial-temporal forecasting tasks, highlighting their ability to tackle complex tasks in transportation systems and generalize across domains.

\subsection{Decision maker}

Traffic signal control (TSC) represents a critical application in which LLMs exhibit human-like adaptability \citep{lai2023large,tang2024large,da2024prompt,movahedi2025crossroads,pang2024illm,wang2024llm,masri2025large} . 
\cite{lai2023large} pioneered this direction with LLMLight, which employs GPT-3.5/4 as intuitive decision makers for traffic light optimization. By prompting LLMs with real-time traffic conditions, the framework mimics human operators' contextual reasoning.
Additionally, the authors developed LightGPT, an cost-efficient backbone LLM pretrained on traffic patterns tailored for TSC tasks. 
The framework outperforms RL-based methods in reducing waiting times and generalizes across diverse traffic datasets without retraining.
Subsequent studies such as LA-Light \citep{wang2024llm} integrates LLMs with perception tools to process static and dynamic data and hybridizes LLM reasoning with RL outputs, demonstrating superior fault tolerance during sensor outages. 
For complex intersections that encounter unpredictable traffic patterns, \cite{movahedi2025crossroads} advances closed-loop adaptation through their Generally Capable Agent (GCA) framework, where an actor-critic architecture enables iterative refinement of phase plans based on simulated outcomes. In SUMO simulations, the GCA-based controller outperformed conventional methods, reducing halted vehicles by 48.03\% and increasing average speeds by 25.29\%. 
These frameworks collectively address the limitations of static rule-based systems and data-hungry RL approaches through explainable, transferable decision-making.


In travel behavior modeling, LLMs bypass the traditional data-driven parameter calibration parameter calibration by leveraging semantic reasoning. \cite{mo2023large} demonstrates that structured prompts enable GPT-family models to match supervised models like multinomial logit models and neural networks in mode choice accuracy.
However, the study notes occasional "hallucinations" in explanations where outputs lack logical consistency.
\cite{liu2024can} further evaluate the ability of LLMs to simulate human decision making in mode choice using a stated preference dataset. They first test zero-shot LLMs, finding significant behavioral misalignment due to discrepancies between LLM reasoning and real traveler preferences. To address this, they introduce persona-based few-shot learning, effectively bridging the gap between LLM reasoning patterns and empirical traveler preferences.
Beyond discrete choices, \cite{tang2024itinera} employs LLMs for personalized itinerary generation. 
LLMs generate human-readable itineraries by integrating optimized POI sequences and contextual descriptions. This hybrid method overcomes the limitations of spatial unawareness and static knowledge of standalone LLMs, ensuring both personalization and geographic feasibility.
The LLM-based decision making process can also be integrated with behavior theory. In \citep{chen2025perceptions}, the authors synthesize behavioral theory (Protective Action Decision Model), contextual cues, and a memory-based RL to enhance wildfire evacuation decision prediction.
LLMs are prompted to simulate human cognitive processes, structured into threat assessment and risk perception stages. The memory module refines decisions by storing and retrieving past errors and self-reflections.

Autonomous driving systems increasingly integrate LLMs as cognitive engines and central decision makers \citep{jin2023surrealdriver,cui2023drivellm,liu2023mtd,wang2023drivemlm,jiang2024koma,zhou2024safedrive,fang2024towards,chen2024genfollower,pang2024large,chen2024drivinggpt,zhou2025behaviorgpt}. 
The direct utilization of their reasoning capabilities is to interpret dynamic environments and generate explainable actions.
Specifically, \cite{cui2024drive} integrate LLMs with voice interfaces to enable natural language interaction in autonomous vehicles. They employ a two-stage architecture in which an LLM processes driver/passenger verbal commands and translates them into structured vehicle control signals. 
Similarly, \cite{chen2024driving} aligns vectorized object-level data with LLM representations using pretraining on "vector captioning" datasets, allowing the model to answer driving-related questions and generate contextualized control commands.
VLMs like DriveLM \citep{sima2024drivelm} extend this paradigm further by structuring scene understanding as Graph Visual Question Answering (GVQA), enabling multistep reasoning about object interactions through perception-prediction-planning QA pairs. Complementing these multimodal approaches, DriveGPT4 \citep{xu2024drivegpt4} processes temporal video sequences and textual queries in an end-to-end framework, directly predicting low-level control signals while providing human-interpretable action rationale.

Another line of work focuses on enhancing decision-making through memory-augmented reasoning and experience-based reflection. \cite{fu2024drive} addresses long-tail corner cases by deploying LLMs in a closed-loop simulator, where tools like trajectory planning and environmental memory enable continuous adaptation to unseen scenarios. Agent-Driver \citep{mao2023language} formalizes this concept with a cognitive architecture featuring a tool library, experiential memory of common sense and knowledge, and a reasoning engine capable of CoT planning and self-correction. DiLu \citep{wen2023dilu} further improves generalization by decoupling reasoning (applying common-sense knowledge) and reflection (learning from past decisions), outperforming traditional RL methods.
Building upon these frameworks, recent studies directly reformulate core autonomous driving tasks through the lens of language modeling. 
GPT-Driver \citep{mao2023gpt} reformulates motion planning as a language modeling problem, representing planner inputs and outputs as language tokens. 
They introduce a prompting-reasoning-finetuning strategy to stimulate the LLM's numerical reasoning potential, enabling it to describe precise trajectory coordinates and its internal decision-making process in natural language. 
DrivingGPT \citep{chen2024drivinggpt} proposes to unify both driving simulation and trajectory planning into a single sequence modeling problem. They introduce a multimodal driving language based on image and action tokens and train the model through standard next-token prediction.
Together, these approaches collectively shift autonomous driving from modular pipelines to flexible and language-grounded systems.

Operational traffic optimization also benefits from multi-agent coordination capabilities of LLMs. 
CoMAL \citep{yao2024comal} integrates multiple LLM agents to tackle mixed-autonomy traffic problems by optimizing traffic flow. It employs a collaboration module where autonomous vehicles communicate using LLMs to allocate roles and discuss strategies in real-time. 
It demonstrates superior performance in optimizing mixed-autonomy traffic compared to RL-based models.
Meanwhile, \cite{orfanoudakis2025optimizing} combines the Decision Transformer (DT) \citep{chen2021decision} with GNNs to optimize the charging schedules of electric vehicles. GPT-2-based DT is trained to predict actions by modeling sequences of states, actions, and rewards. This hybrid approach outperforms the heuristic baselines and RL methods in the EV2Gym simulator.
These works illustrate LLMs' roles as both collaborative planners and sequential action learners in infrastructure-scale optimization.

\subsection{Decision guider}
In addition to serving as the central decision maker, LLMs can also be employed to guide decision making by generating action candidates or language instructions.
Benefiting from extensive prior knowledge, LLMs can even provide guidance for unseen tasks, thus improving the sample efficiency of the control subsystem in ITS. LLMs increasingly guide safety-critical decisions through interpretable intermediate representations. 
AccidentGPT \citep{wang2023accidentgpt} establishes a multimodal safety advisor that converts multi-sensor data to anticipate accidents, issue long-range safety warnings and dialogue-based contextualized recommendations, bridging perception with human-understandable guidance.
For control systems, LLMs act as high-level planners whose predictions guide low-level controllers. 
\cite{sha2023languagempc} propose LanguageMPC, where LLMs reason about traffic scenarios to adjust the priorities of a Model Predictive Control (MPC) system. For example, reweighting cost functions for safety or efficiency and focusing observation matrices on critical vehicles. Similarly, \cite{long2024vlm} integrates VLMs with MPC in VLM-MPC. The VLM as a high-level planner processes camera inputs and driving histories to predict trajectory parameters, while MPC handles dynamic execution, addressing real-world delays. 
These approaches demonstrate how LLMs can inject semantic reasoning into traditional control paradigms without compromising operational safety.
LLMs further predict user intent to align machine actions with human goals. \cite{wang2023chatgpt} embeds ChatGPT as a vehicle “Co-Pilot” that translates natural language commands (e.g., “overtake the truck ahead”) into domain-specific actions. By encoding user instructions alongside contextual memory, the model generates controller selections or trajectory plans.

\subsection{Spatial-temporal predictor}
As Transformers naturally excel at handling sequential data, LLMs have shown outstanding performance in time series analysis \citep{jin2023time,gruver2023large} as well as in spatial-temporal data mining \citep{jin2023large,li2024urbangpt}. One of the most fundamental macroscopic quantities in transportation systems is the traffic flow, usually structured as spatial-temporal data. 
Traffic flow forecasting is one of the fundamental tasks in data-centric transportation systems. Traditional methods include using convolutional neural networks (CNNs), recurrent neural networks (RNNs), and {GNNs} to model graph-based traffic time series \citep{tedjopurnomo2020survey,yin2021deep,xue2025data}. MLP-based \citep{shao2022spatial,qin2023spatio,nie2025contextualizing} and Transformer-based \citep{xu2020spatial,yan2021learning,liu2023spatio} architectures have emerged as new alternatives.

LLMs are revolutionizing traffic flow prediction through novel spatiotemporal tokenization strategies \citep{de2023llm,zhang2024large}.
Early efforts draw inspiration from language models and train forecasters using traffic data such as TrafficBERT \citep{jin2021trafficbert} and Transportation Foundation Model (TFM) \citep{wang2023building}.
TrafficBERT is pre-trained on traffic flow datasets to capture time-series information through self-attention mechanisms, outperforming models trained on specific roads. TFM integrates traffic simulation principles into traffic prediction. Using graph structures and dynamic graph generation algorithms, it is able to model interactions within the transportation system. 
Both studies demonstrate the potential of language models in enhancing traffic forecasting by effectively capturing spatiotemporal semantics.

More recently, GPT-like architectures have been adopted as backbone forecasters for traffic flow.
ST-LLM \citep{liu2024spatial} introduces an embedding module to create a unified spatial-temporal representation and feeds the embeddings to LLMs to predict future traffic time series. 
To adapt the LLM to traffic prediction tasks, ST-LLM employs a partially frozen attention mechanism, where the frozen layers preserve foundational knowledge, and unfrozen attention layers focus on capturing the specific dependencies.
To bridge the gap between sequential text and traffic data, STG-LLM \citep{liu2024stgllm} introduces a spatial-temporal graph tokenizer that transforms traffic data into tokens. 
This transformation reduces the complexity of the graph-structured data, making it more accessible for LLMs.
For explainability, xTP-LLM \citep{guo2024explainable} transforms multimodal traffic information into natural language descriptions. 
CoT prompts are used to guide LLMs in identifying relevant factors from the given information.
Then LLaMA-2 is fine-tuned using language-based instructions to align with the specific requirements of traffic prediction. Empirical evaluations show that xTP-LLM not only achieves competitive accuracy, but also provides intuitive explanations for its predictions. 
Both of the above methods utilize the fine-tuning technique to adapt LLMs to the traffic domain.
To improve efficiency and reduce computational demands, TPLLM \citep{ren2024tpllm} further introduces a {LoRA} fine-tuning approach for GPT-2, allowing effective learning with fewer parameters. 
These architectures are also applied to traffic data imputation tasks \citep{chen2023gatgpt,zhang2024semantic,nie2024imputeformer,fang2025spatiotemporal}.

Moreover, LLMs have also been used as backbone predictors for trajectory prediction by decoding mobility patterns across scales \citep{wang2023would,xue2024prompt,liang2024exploring,haydari2024mobilitygpt,long2024universal,zhu2024unitraj,zhang2024large}.
This can be categorized as vehicle trajectory modeling from a micro-perspective and mobility prediction from a macro-perspective. 
For microscopic trajectory modeling, existing studies explore different trajectory encoding strategies for LLMs. Specifically, LMTraj \citep{bae2024can} transforms trajectory prediction into a QA problem. They convert pedestrian trajectory coordinates and scene images into textual prompts using numerical tokenizers, and integrate them into a QA template for LLMs. \cite{chib2024lg} presents LG-Traj, a method that uses an LLM-based architecture with a motion encoder to capture motion patterns and a social decoder to capture social interactions among pedestrians. 
In \citep{lan2024traj}, Traj-LLM is proposed to leverage pretrained LLMs without explicit prompt engineering. The approach begins with sparse context joint encoding to process agent and scene features into a form comprehensible by LLMs. 
In addition, LC-LLM \citep{peng2024lc} reformulates lane change prediction as a language modeling problem. It processes heterogeneous driving scenario information in natural language prompts for LLMs and employs {SFT} to tailor the LLM for lane change prediction tasks. 

For macroscopic mobility prediction, LLMs are applied to decode complex spatial-temporal dependencies, contextual cues, and behavioral trends embedded in diverse datasets such as GPS trajectories, transit schedules, social media activity, and traffic reports \citep{liu2024human}. Unlike traditional machine learning approaches, LLMs excel at synthesizing unstructured text with structured mobility data based on their fundamental knowledge, allowing them to model nuanced interactions between infrastructure, environment, and user behavior.
To help LLMs analyze human mobility data, LLM-Mob introduced by
\cite{wang2023would} presents concepts of historical stays and context stays to capture both long-term and short-term dependencies in human movement. Additionally, context-inclusive prompts are designed to improve the accuracy of LLMs in generating time-aware predictions. 
Similarly, prompt-based prediction is also explored in LLM-MPE \citep{liang2024exploring}. LLM-MPE pompts LLMs to process textual descriptions of public events and historical mobility data to predict human mobility during such events. 
It converts raw, unstructured event descriptions into a standardized format and segments historical mobility data into regular and event-related components in the prompt. 
Training foundation models using pure mobility data also demonstrates great potentials.
In \citep{haydari2024mobilitygpt}, MobilityGPT is introduced as a geospatially-aware generative model that formulates human mobility modeling as an autoregressive generation task using the GPT architecture. 
They fine-tune MobilityGPT using a Reinforcement Learning from Trajectory Feedback (RLTF) mechanism, minimizing the travel distance between training and synthetically generated trajectories.
UniMob \citep{long2024universal} extents this paradigm and endeavors to unify individual trajectory and crowd flow predictions. UniMob employs a multi-view mobility tokenizer to transform both trajectory and flow data into spatiotemporal tokens, facilitating unified sequential modeling through a diffusion Transformer architecture. 
Finally, LLMs are also integrated into multimodal demand prediction by fusing heterogeneous data sources and reformulating time series forecasting, such as electric vehicles charging demand \citep{qu2024chatev}, taxi and bike usage demand \citep{liu2024spatial}, and package delivery demand \citep{nie2025joint}.

\subsection{Summary and outlook}
The integration of LLMs as decision facilitators has introduced remarkable capabilities in transportation systems, spanning decision-making, action guidance, and spatial-temporal forecasting. As decision makers, LLMs demonstrate human-like adaptability in traffic signal control, autonomous driving, and route optimization, outperforming traditional methods while offering explainable reasoning. In guiding decisions, LLMs bridge high-level reasoning with low-level control systems through interpretable instructions, enhancing safety and efficiency in complex scenarios. For spatial-temporal forecasting, LLMs decode intricate mobility patterns by unifying multimodal data representations, achieving competitive performance in traffic flow and trajectory prediction through tokenization and fine-tuning strategies. 
As LLMs evolve from auxiliary tools to central predictors, their integration with ITS techniques such as digital twin platforms and IoT ecosystems will likely promote predictive intelligence in smart transportation applications.
{Yet, several critical limitations hinder these advances. First, LLMs incur high latency and resource demands, challenging their use in real-time control loops or embedded ITS hardware without significant model compression or architectural redesign. Second, LLMs struggle with exact numerical reasoning and fine-grained geometric constraints inherent to traffic dynamics, leading to suboptimal or unsafe control suggestions when precise calculations matter.
}

We list several potential directions for future explorations in the following:
\begin{itemize}
\item \textbf{Data representation and computation:} LLMs' text-centric training limits their effectiveness in processing numerical and geometric data inherent to transportation systems. Although tokenization methods such as spatial-temporal graph tokenizers \citep{liu2024stgllm} and vector captioning \citep{chen2024driving} show promise, fundamental gaps persist in representing continuous physical spaces and performing precise numerical computations. It remains an open question how to develop a customized data representation method that is suitable for multiscale traffic data \citep{nie2024spatiotemporal} and adaptable to LLMs.

\item \textbf{Safety and alignment:} Critical applications like autonomous driving require rigorous safety guarantees. Current frameworks address this through simulation sandboxes \citep{fu2024drive} and safety alignment techniques \citep{yang2024driving,xie2025vlms}, but real-world deployment demands formal verification methods and enhanced robustness against adversarial inputs \citep{liu2024adversarial}. Equipping LLMs with safety evaluation modules is a crucial step before practical implementation.

\item \textbf{Efficient domain-specific adaptation:} While fine-tuning approaches like LoRA \citep{hu2022lora} improve the efficiency of adapting LLMs, the scalability of LLMs for infrastructure-scale optimization and real-time decision-making remains computationally intensive. Hybrid architectures such as MoE that combine general LLMs with lightweight domain-specific models \citep{fedus2022switch} could balance adaptability with operational efficiency. This suggests pathways towards resource-efficient LLM predictors. 

\item \textbf{Standardized evaluation:} Currently, the research community lacks standardized benchmarks to assess the predictive capabilities of LLMs in transportation tasks. 
Emerging datasets such as DriveLM \citep{sima2024drivelm} and evaluation frameworks \citep{fan2024learning} represent initial steps towards unified evaluation protocols. Future efforts are needed to establish benchmarks for reproducible and open-source studies.
\end{itemize}



\section{Practical Guidance}\label{sec:practical_guidance}
In this section, we provide a review of publicly available resources that can facilitate the deployment of LLMs in transportation domains and help solve practical problems. As a practical guidance, we focus on related datasets, collection of literature, available software libraries, and hardware requirements.

\subsection{Language-enhanced datasets}
Adapting and applying LLMs in transportation requires customized datasets. A natural customization is language-enhanced datasets, i.e., the raw traffic data is coupled with language descriptions or language labels. Such datasets are necessary for grounding LLMs in transportation domains. To this end, we summarize several emerging language-enhanced ITS and autonomous driving datasets for LLM development, which are synthesized from the provided research papers.
Tab. \ref{tab:list_of_dataset} demonstrates the rapid evolution of LLM applications in transportation systems, particularly in bridging the gap between raw sensor data and human-understandable decision processes.
However, due to the rapid development of this field, we list only a few emerging datasets and benchmarks. These resources will be updated frequently on our online project page. \textbf{We welcome researchers to contribute their related work, datasets, and benchmarks to our collections via GitHub.}

\begin{table}[!htbp]
    \centering
    \caption{Summary of language-enhanced ITS and autonomous driving datasets for LLM development and evaluation benchmarks.}
    \label{tab:list_of_dataset}
    \begin{small}
    \begin{tabularx}{0.99\textwidth}{X|p{0.05\columnwidth}|p{0.07\columnwidth}|X|X}
    \toprule
    \textbf{Dataset} & \textbf{Year} & \textbf{Venue} & \textbf{Task} & \textbf{Use Case in LLM Development} \\
    \hline
    BDD-X \citep{kim2018textual} & 2018 & ECCV & Action interpretation and control signal prediction & Explainable end-to-end driving through visual question answering. \\

    \hline
    SUTD-TrafficQA \citep{xu2021sutd} & 2021 & CVPR & Video causal reasoning over traffic events & Evaluating the reasoning capability over 6 tasks. \\

    \hline
    TrafficSafety-2K \citeauthor{zheng2023trafficsafetygpt} & 2023 & arXiv & Annotated traffic incident and crash report analysis & GPT fine-tuning for safety situational awareness. \\

    \hline
    NuPrompt \citep{wu2023language} & 2023 & AAAI & Object-centric language prompt set for 3D driving scenes & Prompt-based driving task to predict the described object trajectory. \\
    
    \hline
    LaMPilot \citep{ma2024lampilot} & 2024 & CVPR & Code generation for autonomous driving decisions & CoT reasoning and instruction following for lane changes and speed adaptation. \\
    
    \hline
    CoVLA \citep{arai2024covla} & 2024 & arXiv & Vision-Language-Action alignment (80+ hrs driving videos) & Trajectory planning with natural language maneuver descriptions. \\
    
    \hline
    VLAAD \citep{park2024vlaad} & 2024 & WACV & Natural language description of driving scenarios & QA systems for driving situation understanding. \\

    \hline
    CrashLLM \citep{fan2024learning} & 2024 & arXiv & Crash outcome prediction (severity, injuries) & What-if causal analysis for traffic safety using 19k crash reports. \\
    
    \hline
    TransportBench \citep{syed2024benchmarking} & 2024 & arXiv & Answering undergraduate-level transportation engineering problem & Benchmarking LLMs on planning, design, management, and control questions. \\
    
    \hline
    Driving QA \citep{chen2024driving} & 2024 & ICRA & 160k driving QA pairs with control commands & Interpreting scenarios, answering questions, and decision-making. \\
    
    \hline
    MAPLM \citep{cao2024maplm} & 2024 & CVPR & Multimodal traffic scene dataset including context, image, point cloud, and HD map & Visual instruction-tuning LLMs and VLMs and vision QA tasks. \\
    
    \hline
    DrivingDojo \citep{wang2024drivingdojo} & 2024 & NeurIPS & Video clips
    with maneuvers, multi-agent interplay, and driving knowledge & Training and action instruction following benchmark for driving world models. \\
    
    \hline
    TransportationGames \citep{{zhang2024transportationgames}} & 2024 & arXiv & Benchmarks of LLMs in memorizing, understanding, and applying transportation knowledge on 10 tasks & Grounding (M)LLMs in transportation-related tasks. \\
    
    \hline
    NuScenes-QA \citep{qian2024nuscenes} & 2024 & AAAI & Benchmark for vision QA in autonomous driving, including 34K visual scenes and 460K QA pairs  & Developing 3D detection and VQA techniques for end-to-end autonomous driving systems.  \\
    
    \hline
    TUMTraffic-VideoQA \citep{zhou2025tumtraffic} & 2025 & aXiv & Temporal traffic video understanding & Benchmarking video reasoning for multiple-choice video question answering. \\
    
    \hline
    V2V-QA \citep{chiu2025v2v} & 2025 & arXiv & Cooperative perception via V2V communication & Fuse perception information from multiple CAVs and answer driving-related questions. \\

    \hline
    DriveBench \citep{xie2025vlms} & 2025 & arXiv & A comprehensive benchmark of VLMs for perception, prediction, planning, and explanation & Visual grounding and multi-modal understanding for autonomous driving. \\

    \bottomrule
    \end{tabularx}
    \end{small}
    \end{table}

\subsection{Available resources}
The rapid progress of LLMs has catalyzed extensive research and tool development, shaping a dynamic ecosystem of academic surveys and software libraries in the AI community. 
To help make it easier to access these open-source resources, we briefly summarize several critical surveys on LLMs and catalog representative open-source frameworks, providing researchers with accessible resources about foundational studies and practical tools for developing LLM applications in transportation.

Tab. \ref{tab:list_of_surveys} organizes influential surveys that systematically discuss the evolution, development, advancements, techniques, and challenges of LLMs. Tab. \ref{tab:list_of_library} outlines widely adopted libraries that involve application development, deployment, evaluation, and experimentation of LLMs.

\begin{table}[!htbp]
    \centering
    \caption{Representative surveys on LLMs and related techniques. Note that many of these surveys are still being updated.}
    \label{tab:list_of_surveys}
    \begin{small}
    \begin{tabularx}{0.95\textwidth}{X|p{0.05\columnwidth}|p{0.08\columnwidth}|X}
    \toprule
    \textbf{Paper Title} & \textbf{Year} & \textbf{Venue} & \textbf{Scope and Focus} \\ 
    \hline
    A survey of Large Language Models \citep{zhao2023survey} & 2023 & arXiv & Reviews the evolution of LLMs, pretraining, adaptation, post-training, evaluation, and benchmarks. \\ 
    \hline
    Large Language Models: A Survey \citep{minaee2024largelanguagemodelssurvey} & 2024 & arXiv & Reviews LLM families (GPT, LLaMA, PaLM), training techniques, datasets, and benchmark performance. \\ 
    \hline
    Retrieval-Augmented Generation for Large Language Models: A Survey \citep{gao2023retrieval} & 2023 & arXiv  & Introduces the progress of RAG paradigms, including the naive RAG, the advanced RAG, and the modular RAG. \\ 
    \hline
    A Survey on In-context Learning:  \citep{dong2022survey} & 2022 & arXiv & Summarizes training strategies, prompt designing strategies, and various ICL application scenarios, such as data engineering and knowledge updating. \\ 
    \hline
    Instruction Tuning for Large Language Models: A Survey \citep{zhang2023instruction} & 2023 & arXiv & Reviews methodology of SFT, SFT datasets, applications to different modalities, and influence factors. \\ 
    \hline
    Towards Reasoning in Large Language Models: A Survey \citep{huang2022towards} & 2022 & ACL & Examines techniques for improving and eliciting reasoning in LLMs, methods and benchmarks for evaluating reasoning abilities. \\ 
    \hline
    A Survey of LLM Surveys:  \url{https://github.com/NiuTrans/ABigSurveyOfLLMs} & 2024 & GitHub & Compiles 150+ surveys across subfields like alignment, safety, and {MLLMs}. \\ 
    \bottomrule
    \end{tabularx}
    \end{small}
    \end{table}

\begin{table}[!htbp]
    \centering
    \caption{Popular open-source libraries for LLM development.}
    \label{tab:list_of_library}
    \begin{small}
    \begin{tabularx}{0.95\textwidth}{p{0.15\columnwidth}|p{0.3\columnwidth}|p{0.2\columnwidth}|X}
    \toprule
    \textbf{Library Name} & \textbf{Basic Functions} & \textbf{Use Cases} & \textbf{URL} \\ 
    \hline
    Hugging Face Transformers & Pretrained models (NLP, vision) and fine-tuning pipelines & Model deployment, adapt tuning & \url{https://huggingface.co/docs/transformers} \\ 
    \hline
    DeepEval & Framework for evaluating LLM outputs using metrics like groundedness and bias & Educational applications, hallucination detection & \url{https://github.com/confident-ai/deepeval} \\ 
    \hline
    RAGAS & Quantifies RAG pipeline performance & Context relevance scoring, answer quality & \url{https://github.com/explodinggradients/ragas} \\ 
    \hline
    Sentence Transformers & Generates dense text embeddings for semantic similarity tasks & Survey item correlation analysis, retrieval & \url{https://www.sbert.net/} \\ 
    \hline
    LangChain & Chains LLM calls with external tools for multi-step workflows & RAG, agentic reasoning, data preprocessing & \url{https://www.langchain.com/} \\ 
    \hline
    DeepSpeed & A deep learning optimization library developed by Microsoft, which has been used to train LLMs & Distributed training, memory optimization, pipeline parallelism & \url{https://www.deepspeed.ai/} \\ 
    \hline
    FastMoE & A specialized training library for MoE models based on PyTorch & Transfer Transformer models to MoE models, data parallelism, model parallelism & \url{https://fastmoe.ai/} \\
    \hline
    Ollama & Local LLM serving with support for models like Llama and Mistral & Offline inference, privacy-sensitive apps & \url{https://ollama.ai} \\
    \hline
    OpenLLM & 	Optimizes LLM deployment as production APIs compatible with OpenAI standards & Scalable model serving, cloud/on-prem hosting & \url{https://github.com/bentoml/OpenLLM} \\
    \bottomrule
    \end{tabularx}
    \end{small}
    \end{table}

\subsection{Computational requirement}
Pretraining a foundational LLM in a vertical domain such as transportation can be infeasible due to extremely high resource consumption. However, recent advances in {PEFT} methods such as LoRA \citep{hu2022lora} and QLoRA \citep{dettmers2023qlora} have made customization of LLMs widely accessible, allowing mid-range hardware to handle models previously restricted to enterprise-grade infrastructure. 

To help the researcher have a basic understanding of the hardware requirements to adapt LLMs, 
we collect and organize data from peer-reviewed studies, community benchmarks, and industry experiments to provide a preliminary analysis of hardware requirements.
Tab. \ref{tab:model_specifications} is a summary of approximate hardware requirements and performance statistics for fine‐tuning LLaMA models across sizes. Note that these numbers vary widely depending on the exact training setup, precision (FP16, 8‑bit, 4‑bit quantization), and optimization strategies used. 
As can be seen, the full-parameter fine‑tuning method requires significantly more GPU memory and typically a distributed training setup, while PEFT methods like LoRA dramatically reduce the number of trainable parameters and memory usage so that even a single high-VRAM consumer GPU (or a small GPU cluster) can be used.

\begin{table}[!htbp]
    \centering
    \caption{Hardware requirements for fine-tuning and inference across LLaMA model sizes. BS = Batch Size. Estimated values marked "(est.)" derive from scaling laws. Inference rates measured at batch size 1 unless noted. The numbers below are rough estimates aggregated from various community benchmarks and articles. Actual requirements and performance may differ for specific configurations.}
    \label{tab:model_specifications}
    \begin{small}
    \begin{tabularx}{0.95\textwidth}{l|X|X|X|X|X|X}
    \toprule
        \textbf{Model Size} & \textbf{Full Tuning GPUs} & \textbf{LoRA Tuning GPUs} & \textbf{Full Tuning BS/GPU} & \textbf{LoRA BS/GPU} & \textbf{Tuning Time (Hours)} & \textbf{Inference Rate (Tokens/s)} \\
        \hline
        \textbf{7B}   & 2$\times$A100 80GB            & 1$\times$RTX 4090 24GB       & 1-2            & 4-8           & 3-5            & 27-30           \\
        \midrule
        \textbf{13B}  & 4$\times$A100 80GB (est.)     & 2$\times$A100 40GB           & 1              & 2-4           & 8-12           & 18-22           \\
        \midrule
        \textbf{70B}  & 8$\times$H200 80GB & 4$\times$H200 80GB           & 1              & 1-2           & 24-36          & 12-15           \\
        \midrule
        \textbf{405B} & 64$\times$H200 80GB (est.)    & 16$\times$H200 80GB (est.)    & 1 (est.)       & 1 (est.)      & 72-96 (est.)   & 5-8             \\
        \bottomrule
    \end{tabularx}
     \end{small}
\end{table}

Please note that these statistics provide a rough guide for planning hardware budget for fine‑tuning LLaMA models with different methods. For more detailed and up‑to‑date benchmarks, reviewing community resources and vendor documentation is recommended. For more concise measures derived from field experiments, see \citet{zhao2023survey}.


\section{Discussion}\label{sec:discussion}

\subsection{Future opportunities of LLM4TR}
While previous sections have discussed possible future directions for LLM-driven transportation research from a methodological perspective, this section provides a broader view of opportunities for future studies, particularly focusing on the deployment of LLMs in real-world transportation systems. In the following, we highlight five potential directions that can address current gaps and shape the evolution of our research field.


\begin{enumerate}
    \item \textbf{Bridging the industry-academia deployment gap in LLM-driven solutions}:
    Despite advances in LLM-driven smart traffic analysis tools (e.g., Open-TI \citep{da2024open} and GenAI-ITS \citep{xu2024genai}), a persistent gap exists between experimental prototypes and real-world deployment. For example, Open-TI shows progress by combining conversational interfaces (via GPT-3.5) with domain tools (SUMO, CityFlow) to automate multistep workflows, from parsing user queries (e.g., "Optimize bike lanes") to executing simulations. However, challenges remain in standardizing tool integration, ensuring computational scalability, and adapting LLMs to mainstream industry software. Future work should prioritize collaborative frameworks where academia co-designs LLM agents with transportation agencies, enhancing technical interoperability with existing infrastructure (e.g., traffic signal controllers, data management APIs). Additionally, "LLM-as-a-service" platforms could facilitate access to advanced tools, enabling entities with limited resources to get access to AI-driven ITS tools through natural language interfaces.

    \item \textbf{Situational awareness through hybrid knowledge grounding:}  
    LLMs risk generating plausible but rule-violating solutions without robust domain grounding in transportation scenarios. Approaches such as the RAG-enhanced multi-agent system in \cite{xu2024genai} represent a promising solution by fusing real-time IoT data, traffic theory, and policy documents. Future systems could adopt hierarchical grounding strategies to better adapt LLMs to traffic problems: (1) {spatiotemporal grounding} via live sensor streams and vehicle-to-everything (V2X) communications for context-aware responses; (2) {theoretical grounding} through explicitly encoding knowledge about transportation principles (e.g., traffic flow theory, traffic rules); and (3) {political grounding} using municipal regulations to ensure legal compliance. For instance, an LLM congestion pricing recommendation scheme should refer to emission models, equilibrium analysis, and local legislation. Achieving this requires hybrid architectures that closely couple LLMs with traffic models, enabling decisions that are both data-driven and knowledge-consistent.
 
    \item 
     \textbf{Eliciting latent problem-solving abilities for emergent challenges:} 
    After being pre-trained on large-scale corpora, LLMs are possessed with potential abilities as general-purpose task solvers. These abilities might not be explicitly exhibited when LLMs perform some specific traffic problems. Therefore, it is useful to design suitable task instructions or specific {ICL} strategies to elicit such underexplored abilities \citep{shin2020autoprompt,wei2022chain}.
    Strategic ability elicitation could unlock these capacities. For example, \textit{role-playing or persona adoption} strategy instructs the model to adopt the persona of a domain expert (e.g., an experienced traveler \citep{liu2024can}). This approach leverages the latent knowledge by framing responses in the voice and reasoning style of an expert, which can yield more domain-appropriate answers. The {problem decomposition} strategy breaks a complex domain-specific problem into a series of simpler sub-problems. Then the compositional tasks can be solved by meta agents.
    Finally, using {domain-contextualized instruction tuning} to guide LLMs to focus on relevant domain aspects and follow a structured approach to the problem.
    \item \textbf{Model compression for real-time decision-making:}  
    Deploying LLMs on resource-constrained edge devices (e.g., onboard vehicle computers, roadside units) necessitates lightweight yet capable models \citep{liu2024resource}. 
    Recent advances in model compression suggests a promising pathway. By dynamically {pruning non-critical parameters}, e.g., tailoring sparsification to preserve traffic-related knowledge such as route optimization while compressing unrelated linguistic knowledge, models can be significantly reduced in size without undermining task-specific performance \citep{ma2023llm}. 
    Furthermore, {hardware-aware quantization} methods \citep{lin2024awq}, such as mixed-precision and adaptive quantization schemes, are emerging to exploit the computational strengths of modern GPUs/TPUs. 
    {One such advancement is the use of 4-bit quantization techniques for efficient LLMs serving \citep{kim2023memory,zhao2024atom}. 
    In addition to quantization, optimizing the KV cache is also crucial for efficient LLM inference \citep{ge2023model,liu2024kivi,shi2024keep,liu2024minicache}. For example, MiniCache \citep{liu2024minicache} compresses the key-value (KV) cache by merging similar states across adjacent layers of LLMs, significantly reducing the memory footprint.}
    These methods can reduce memory footprint and inference latency in a way that is sensitive to the unique requirements of {real-time decision-making in} transportation applications, without reliance on cloud APIs. 
    \item 
    \textbf{Toward interpretable and trustworthy transportation AI:} Traditional learning-based decision-making tools such as RL can be a black-box system, which is difficult for transportation agencies to understand and evaluate. LLMs can elucidate the design logic and action-making trajectory of themselves through CoT reasoning \citep{wei2022chain}. This property can be used for the interpretation of the results.
    While the CoT reasoning offers some transparency benefits, transportation agencies require stricter interpretability. Future frameworks can explore:  
    {implementing formal verification} to verify safety-critical outputs;   
    {developing hybrid interpretability tools} by merging CoT with simulation-based "digital twins" to test proposals of LLM in virtual environments before deployment. 
    Crucially, building public traffic participants' trust in LLMs also requires calling on more people to participate in the development, use, and validation of LLM applications in transportation systems.
    \item \textbf{LLMs for social good in transportation systems:} AI for social good initiatives harness AI technologies to empower communities, drive equitable decision-making, and tackle complex social and environmental challenges \citep{tomavsev2020ai,cowls2021definition}.
    The potential of LLMs to address societal inequities and promote inclusive transportation ecosystems expands significantly. Future research should prioritize harnessing LLMs as equity amplifiers that identify and mitigate systemic biases in transportation infrastructure, policies, and services. By analyzing diverse data from socioeconomic demographics and mobility patterns to public feedback, LLMs can optimize transit routes for underserved communities and adjust ride-sharing subsidies dynamically based on real-time equity metrics. Moreover, LLM-driven participatory frameworks \citep{zhou2024large} can offer an opportunity for marginalized groups to co-design transportation solutions through natural language interactions. 
    Additionally, by personalizing eco-routing suggestions and gamifying carbon footprint reduction through conversational agents, LLMs could navigate travel behaviors toward environmentally conscious choices.

\end{enumerate}

These opportunities stress the need for interdisciplinary collaboration. LLMs for transportation must evolve beyond conversation interfaces into cyber-physical-social systems that harmonize sensing, learning, modeling, and managing traffic. Realizing this vision will depend on tackling shared technical challenges: curating high-quality multimodal traffic corpora, establishing standardized evaluation pipeline, and fostering open ecosystems where modular LLM tools can be safely composed by diverse stakeholders. In the following section, we discuss the potential challenges and concerns of integrating LLMs into transportation systems.

\subsection{Challenges and concerns}
Despite the impressive advances in LLMs and their potential to transform transportation systems, the integration and deployment of LLMs into the safety-critical, ethically sensitive and computationally constrained domains raises several challenges that need careful attention.
Below, we discuss these concerns related to model biases, domain limitations, computational requirements, data privacy, and ethical implications in depth. {Several potential or emerging methods to mitigate them are also mentioned.}

\begin{enumerate}
    \item \textbf{Bias and hallucinations inherent in LLMs:} LLMs are typically trained on massive, heterogeneous datasets collected from the internet. Consequently, they may inherit and even amplify biases present in the training data. In transportation contexts, such biases can lead to erroneous predictions or inappropriate recommendations. In addition, the presence of hallucinations, which can cause LLMs to generate factually incorrect information, poses significant risks. In safety-critical systems like autonomous vehicles or traffic control centers, even minor hallucinations can have severe real-world consequences. {Recent techniques such as RAG \citep{guu2020retrieval} can reduce hallucinations by grounding LLMs in factual and domain-specific data. In addition, fine-tuning on curated transportation datasets, use of bias-aware training objectives \citep{ferrara2023should}, and post-hoc validation layers can help mitigate embedded biases and improve factuality.}
    \item \textbf{Ensuring rigor and controllability in LLM-driven solutions:}
    While LLMs offer great potential, their integration into transportation systems requires rigor in performance guarantees. A core challenge lies in reconciling the probabilistic, opaque nature of LLMs with the deterministic nature of traffic rules and reliability required for real-time traffic management. {Researchers are exploring certifiable validation frameworks that quantify uncertainty bounds \citep{lin2023generating}, enforce context consistency in model outputs, and rigorously test LLM-based solutions against adversarial scenarios \citep{kumar2023certifying}. To maintain human supervision, explainability tools like counterfactual reasoning interfaces \citep{feder2021causalm} or rule-constrained decoding methods should be mandated, allowing operators to monitor how LLM-derived recommendations align with controllability and operational constraints.}
    \item \textbf{Domain-specific problem-solving and numeric computing limitations:} While LLMs excel at processing and generating natural language, their capabilities in precise numerical computing and domain-specific problem solving remain limited. Transportation problems often include precise and quantitative analysis that depend on accurate numerical computations or mathematical optimization, such as real-time traffic prediction, equilibrium analysis, and dynamic routing. Standard LLMs are not inherently designed for these tasks, which can lead to suboptimal performance when used in isolation. 
    {Hybrid models that combine LLMs with symbolic solvers \citep{pan2023logic}, constraint programming, or external simulation tools \citep{gao2024large} are emerging. Alternatively, tool-using LLMs \citep{nakano2021webgpt,schick2023toolformer} that call external APIs or plug-ins (e.g., for solving equations or optimization) show promise for extending their capabilities beyond language.}
    \item \textbf{Computational demand for adapting LLMs:} The impressive performance of LLMs comes at a high computational cost. 
    Even fine-tuning and adapting these models on user-defined datasets necessitates prohibitive computational resources. 
    Previous data-driven transportation modeling typically required only moderate or small computational overhead. Thus, computation in the era of LLMs becomes a technical bottleneck for transportation researchers and practitioners with limited hardware computing capacity.
    {{PEFT} methods such as LoRA, adapter modules, and prompt tuning dramatically reduce resource requirements. Lightweight open-source models or cloud-hosted inference APIs can also enable access to powerful LLMs without full deployment.} While API-based solutions offer an accessible approach to using pre-trained LLMs provided by third-part companies, the limited access and model agnosticism largely limits the ways in which capabilities of LLMs can be explored and applied.
    \item \textbf{Privacy and security concerns:} LLMs require large amounts of data, some of which may be sensitive or personal such as real-time location data, travel patterns, or driver behavior information. 
    This raises significant privacy concerns when LLMs memorize and leak personal information. In addition to privacy issues, there are also security risks related with deploying LLMs as decision makers in critical infrastructure. For instance, cyberattacks using adversarial prompts can manipulate LLM-based traffic controller. Such adversarial vulnerabilities of LLMs are adverse for user's security and system's functionality. 
    {Differential privacy, federated learning, and on-device inference can help protect user data when using LLMs \citep{yao2024survey}. In terms of security, recent work explores adversarial robustness training and input sanitization to defend against prompt injection attacks in safety-critical deployments \citep{he2023large,das2025security}. }

    \item \textbf{Ethical, legal, and equitable concerns:} Finally, the integration of LLMs in transportation systems raises several ethical, legal, and equitable issues. 
    The black-box nature of many LLMs complicates interpretability, making it difficult to trace decisions or explain errors when they occur. This opacity is problematic in scenarios where transparency is required for public safety and trust, such as autonomous vehicles. 
    Furthermore, there are concerns about equitable access. Biased outputs of LLMs may disproportionately affect underrepresented or vulnerable groups. 
    For example, route optimization models may prioritize routes through lower-income neighborhoods as "acceptable" congestion zones. 
    While the high cost of deploying and maintaining advanced LLMs may also widen the gap between well-funded urban centers and less affluent regions. Lastly, legal liability in the event of system failures remains largely undefined. 
    If an LLM-recommended merging strategy causes a collision of autonomous vehicles, is the responsibility with the model developer, traffic agency, or vehicle original equipment manufacturer? 
    {
    Addressing these challenges will require interdisciplinary collaboration among technologists, policymakers, and ethicist to create guidelines and practices for the responsible use of LLMs in transportation, such as transparent model auditing \citep{mokander2024auditing}, impact assessments, and equity-aware objective functions during development. 
    On the legal front, ongoing research explores frameworks for assigning liability and setting regulatory standards for LLM-driven decision systems \citep{hacker2023regulating}.}
\end{enumerate}

\section{Conclusion}\label{sec:conclusions}
The integration of LLMs into transportation systems marks a pivotal shift from traditional, data-centric approaches to a unified, language-driven paradigm. Through this survey, we have demonstrated that LLMs are not merely tools for natural language processing but foundational enablers of cyber-physical-social intelligence in transportation systems. By systematically categorizing their roles as information processors, knowledge encoders, component generators, and decision facilitators, the proposed LLM4TR framework redefines how transportation systems sense, learn, model, and manage complex urban dynamics. 
As information processors, they harmonize multimodal data streams; as knowledge encoders, they distill domain expertise into actionable insights; as component generators, they automate system design; and as decision facilitators, they orchestrate human-like reasoning for real-time control. This synergy fosters a self-improving cycle in which LLM enhances operational efficiency, interpretability, and adaptability in ITS.

In conclusion, this survey underscores that LLMs are not just incremental advancements but paradigm-shifting technologies for transportation. By unifying language, knowledge, and generative intelligence, they pave the way for transportation systems that are not only smarter and more efficient but also inherently human-centric. As we stand at the intersection of AI and urban mobility, the LLM4TR framework provides both a roadmap and a call to action, to reshape transportation as a collaborative, adaptive, and sustainable ecosystem powered by the language interface.
Despite early successes, challenges persist in ensuring safety, mitigating biases, and scaling deployments. Future research in this fresh field needs to prioritize hybrid architectures that embed LLMs within physics-aware simulations, foster industry-academia collaboration for real-world validation, and address ethical concerns in equitable access.

\section*{Acknowledgement}
This research was sponsored by the National Natural Science Foundation of China (52125208), the National Natural Science Foundation of China's Fundamental Research Program for Young Students (524B2164), grants from the Research Grants Council of the Hong Kong Special Administrative Region, China (Project No. PolyU/15206322 and PolyU/15227424), and a grant from Otto Poon Charitable Foundation Smart Cities Research Institute, The Hong Kong Polytechnic University (CD06).

\section*{Declaration of generative AI and AI-assisted technologies in the writing process}
During the preparation of this work, the authors used DeepSeek (an AI-assisted tool) to improve language clarity and readability. 
After using this tool, the authors reviewed and edited the content as needed and take full responsibility for the content of the published article.

\footnotesize
\bibliographystyle{elsarticle-harv}
\bibliography{main}

\begin{thebibliography}{352}
\expandafter\ifx\csname natexlab\endcsname\relax\def\natexlab#1{#1}\fi
\expandafter\ifx\csname url\endcsname\relax
  \def\url#1{\texttt{#1}}\fi
\expandafter\ifx\csname urlprefix\endcsname\relax\def\urlprefix{URL }\fi

\bibitem[{Abdelrahman et~al.(2024)Abdelrahman, Abdel-Aty, and Wang}]{abdelrahman2024video}
Abdelrahman, A.~S., Abdel-Aty, M., Wang, D., 2024. Video-to-text pedestrian monitoring (vtpm): Leveraging computer vision and large language models for privacy-preserve pedestrian activity monitoring at intersections. arXiv preprint arXiv:2408.11649.

\bibitem[{Abu~Tami et~al.(2024)Abu~Tami, Ashqar, Elhenawy, Glaser, and Rakotonirainy}]{abu2024using}
Abu~Tami, M., Ashqar, H.~I., Elhenawy, M., Glaser, S., Rakotonirainy, A., 2024. Using multimodal large language models (mllms) for automated detection of traffic safety-critical events. Vehicles 6~(3), 1571--1590.

\bibitem[{Achiam et~al.(2023)Achiam, Adler, Agarwal, Ahmad, Akkaya, Aleman, Almeida, Altenschmidt, Altman, Anadkat, et~al.}]{achiam2023gpt}
Achiam, J., Adler, S., Agarwal, S., Ahmad, L., Akkaya, I., Aleman, F.~L., Almeida, D., Altenschmidt, J., Altman, S., Anadkat, S., et~al., 2023. Gpt-4 technical report. arXiv preprint arXiv:2303.08774.

\bibitem[{Arai et~al.(2024)Arai, Miwa, Sasaki, Yamaguchi, Watanabe, Aoki, and Yamamoto}]{arai2024covla}
Arai, H., Miwa, K., Sasaki, K., Yamaguchi, Y., Watanabe, K., Aoki, S., Yamamoto, I., 2024. Covla: Comprehensive vision-language-action dataset for autonomous driving. arXiv preprint arXiv:2408.10845.

\bibitem[{Arteaga and Park(2025)}]{arteaga2025large}
Arteaga, C., Park, J., 2025. A large language model framework to uncover underreporting in traffic crashes. Journal of Safety Research 92, 1--13.

\bibitem[{Arthurs et~al.(2021)Arthurs, Gillam, Krause, Wang, Halder, and Mouzakitis}]{arthurs2021taxonomy}
Arthurs, P., Gillam, L., Krause, P., Wang, N., Halder, K., Mouzakitis, A., 2021. A taxonomy and survey of edge cloud computing for intelligent transportation systems and connected vehicles. IEEE Transactions on Intelligent Transportation Systems 23~(7), 6206--6221.

\bibitem[{Ba et~al.(2016)Ba, Kiros, and Hinton}]{ba2016layer}
Ba, J.~L., Kiros, J.~R., Hinton, G.~E., 2016. Layer normalization. arXiv preprint arXiv:1607.06450.

\bibitem[{Bach et~al.(2022)Bach, Sanh, Yong, Webson, Raffel, Nayak, Sharma, Kim, Bari, Fevry, et~al.}]{bach2022promptsource}
Bach, S.~H., Sanh, V., Yong, Z.-X., Webson, A., Raffel, C., Nayak, N.~V., Sharma, A., Kim, T., Bari, M.~S., Fevry, T., et~al., 2022. Promptsource: An integrated development environment and repository for natural language prompts. arXiv preprint arXiv:2202.01279.

\bibitem[{Bae et~al.(2024)Bae, Lee, and Jeon}]{bae2024can}
Bae, I., Lee, J., Jeon, H.-G., 2024. Can language beat numerical regression? language-based multimodal trajectory prediction. In: Proceedings of the IEEE/CVF Conference on Computer Vision and Pattern Recognition. pp. 753--766.

\bibitem[{Bai et~al.(2023)Bai, Bai, Chu, Cui, Dang, Deng, Fan, Ge, Han, Huang, et~al.}]{bai2023qwen}
Bai, J., Bai, S., Chu, Y., Cui, Z., Dang, K., Deng, X., Fan, Y., Ge, W., Han, Y., Huang, F., et~al., 2023. Qwen technical report. arXiv preprint arXiv:2309.16609.

\bibitem[{Bai et~al.(2022)Bai, Kadavath, Kundu, Askell, Kernion, Jones, Chen, Goldie, Mirhoseini, McKinnon, et~al.}]{bai2022constitutional}
Bai, Y., Kadavath, S., Kundu, S., Askell, A., Kernion, J., Jones, A., Chen, A., Goldie, A., Mirhoseini, A., McKinnon, C., et~al., 2022. Constitutional ai: Harmlessness from ai feedback. arXiv preprint arXiv:2212.08073.

\bibitem[{Besta et~al.(2024)Besta, Blach, Kubicek, Gerstenberger, Podstawski, Gianinazzi, Gajda, Lehmann, Niewiadomski, Nyczyk, et~al.}]{besta2024graph}
Besta, M., Blach, N., Kubicek, A., Gerstenberger, R., Podstawski, M., Gianinazzi, L., Gajda, J., Lehmann, T., Niewiadomski, H., Nyczyk, P., et~al., 2024. Graph of thoughts: Solving elaborate problems with large language models. In: Proceedings of the AAAI Conference on Artificial Intelligence. Vol.~38. pp. 17682--17690.

\bibitem[{Brown et~al.(2020)Brown, Mann, Ryder, Subbiah, Kaplan, Dhariwal, Neelakantan, Shyam, Sastry, Askell, et~al.}]{brown2020language}
Brown, T., Mann, B., Ryder, N., Subbiah, M., Kaplan, J.~D., Dhariwal, P., Neelakantan, A., Shyam, P., Sastry, G., Askell, A., et~al., 2020. Language models are few-shot learners. Advances in neural information processing systems 33, 1877--1901.

\bibitem[{Burghout et~al.(2005)Burghout, Koutsopoulos, and Andreasson}]{burghout2005hybrid}
Burghout, W., Koutsopoulos, H.~N., Andreasson, I., 2005. Hybrid mesoscopic--microscopic traffic simulation. Transportation Research Record 1934~(1), 218--225.

\bibitem[{Cao et~al.(2024{\natexlab{a}})Cao, Zhou, Ma, Ye, Cui, Tang, Cao, Liang, Wang, Rehg, et~al.}]{cao2024maplm}
Cao, X., Zhou, T., Ma, Y., Ye, W., Cui, C., Tang, K., Cao, Z., Liang, K., Wang, Z., Rehg, J.~M., et~al., 2024{\natexlab{a}}. Maplm: A real-world large-scale vision-language benchmark for map and traffic scene understanding. In: Proceedings of the IEEE/CVF Conference on Computer Vision and Pattern Recognition. pp. 21819--21830.

\bibitem[{Cao et~al.(2024{\natexlab{b}})Cao, Zhao, Cheng, Shu, Chen, Liu, Liang, Zhao, Yan, and Li}]{cao2024survey}
Cao, Y., Zhao, H., Cheng, Y., Shu, T., Chen, Y., Liu, G., Liang, G., Zhao, J., Yan, J., Li, Y., 2024{\natexlab{b}}. Survey on large language model-enhanced reinforcement learning: Concept, taxonomy, and methods. IEEE Transactions on Neural Networks and Learning Systems, 1--21.

\bibitem[{Chang et~al.(2024)Chang, Wang, Zhang, Ge, and Li}]{chang2024llmscenario}
Chang, C., Wang, S., Zhang, J., Ge, J., Li, L., 2024. Llmscenario: Large language model driven scenario generation. IEEE Transactions on Systems, Man, and Cybernetics: Systems 54~(11), 6581--6594.

\bibitem[{Chen and Cheng(2010)}]{chen2010review}
Chen, B., Cheng, H.~H., 2010. A review of the applications of agent technology in traffic and transportation systems. IEEE Transactions on intelligent transportation systems 11~(2), 485--497.

\bibitem[{Chen et~al.(2023{\natexlab{a}})Chen, Zhang, Langren{\'e}, and Zhu}]{chen2023unleashing}
Chen, B., Zhang, Z., Langren{\'e}, N., Zhu, S., 2023{\natexlab{a}}. Unleashing the potential of prompt engineering in large language models: a comprehensive review. arXiv preprint arXiv:2310.14735.

\bibitem[{Chen et~al.(2022)Chen, Huang, Lam, Pan, Hsu, Sumalee, and Zhong}]{chen2022data}
Chen, C., Huang, Y., Lam, W.~H., Pan, T., Hsu, S., Sumalee, A., Zhong, R., 2022. Data efficient reinforcement learning and adaptive optimal perimeter control of network traffic dynamics. Transportation Research Part C: Emerging Technologies 142, 103759.

\bibitem[{Chen et~al.(2016)Chen, Ma, Susilo, Liu, and Wang}]{chen2016promises}
Chen, C., Ma, J., Susilo, Y., Liu, Y., Wang, M., 2016. The promises of big data and small data for travel behavior (aka human mobility) analysis. Transportation research part C: emerging technologies 68, 285--299.

\bibitem[{Chen et~al.(2024{\natexlab{a}})Chen, Zhu, Yang, Wang, and Wang}]{chen2024data}
Chen, D., Zhu, M., Yang, H., Wang, X., Wang, Y., 2024{\natexlab{a}}. Data-driven traffic simulation: A comprehensive review. IEEE Transactions on Intelligent Vehicles 9~(4), 4730--4748.

\bibitem[{Chen et~al.(2023{\natexlab{b}})Chen, Yuan, Huang, Guo, Wang, and Chen}]{chen2023feedback}
Chen, H., Yuan, K., Huang, Y., Guo, L., Wang, Y., Chen, J., 2023{\natexlab{b}}. Feedback is all you need: from chatgpt to autonomous driving. Science China. Information Sciences 66~(6), 166201.

\bibitem[{Chen et~al.(2021)Chen, Lu, Rajeswaran, Lee, Grover, Laskin, Abbeel, Srinivas, and Mordatch}]{chen2021decision}
Chen, L., Lu, K., Rajeswaran, A., Lee, K., Grover, A., Laskin, M., Abbeel, P., Srinivas, A., Mordatch, I., 2021. Decision transformer: Reinforcement learning via sequence modeling. Advances in neural information processing systems 34, 15084--15097.

\bibitem[{Chen et~al.(2024{\natexlab{b}})Chen, Sinavski, H{\"u}nermann, Karnsund, Willmott, Birch, Maund, and Shotton}]{chen2024driving}
Chen, L., Sinavski, O., H{\"u}nermann, J., Karnsund, A., Willmott, A.~J., Birch, D., Maund, D., Shotton, J., 2024{\natexlab{b}}. Driving with llms: Fusing object-level vector modality for explainable autonomous driving. In: 2024 IEEE International Conference on Robotics and Automation (ICRA). IEEE, pp. 14093--14100.

\bibitem[{Chen et~al.(2025)Chen, Wang, Sun, Zhao, and Xu}]{chen2025perceptions}
Chen, R., Wang, C., Sun, Y., Zhao, X., Xu, S., 2025. From perceptions to decisions: Wildfire evacuation decision prediction with behavioral theory-informed llms. arXiv preprint arXiv:2502.17701.

\bibitem[{Chen et~al.(2024{\natexlab{c}})Chen, Peng, Tiu, Wu, Chen, Zhu, and Zheng}]{chen2024genfollower}
Chen, X., Peng, M., Tiu, P., Wu, Y., Chen, J., Zhu, M., Zheng, X., 2024{\natexlab{c}}. Genfollower: Enhancing car-following prediction with large language models. IEEE Transactions on Intelligent Vehicles, 1--11.

\bibitem[{Chen et~al.(2023{\natexlab{c}})Chen, Wang, and Xu}]{chen2023gatgpt}
Chen, Y., Wang, X., Xu, G., 2023{\natexlab{c}}. Gatgpt: A pre-trained large language model with graph attention network for spatiotemporal imputation. arXiv preprint arXiv:2311.14332.

\bibitem[{Chen et~al.(2024{\natexlab{d}})Chen, Wang, and Zhang}]{chen2024drivinggpt}
Chen, Y., Wang, Y., Zhang, Z., 2024{\natexlab{d}}. Drivinggpt: Unifying driving world modeling and planning with multi-modal autoregressive transformers. arXiv preprint arXiv:2412.18607.

\bibitem[{Chib and Singh(2024)}]{chib2024lg}
Chib, P.~S., Singh, P., 2024. Lg-traj: Llm guided pedestrian trajectory prediction. arXiv preprint arXiv:2403.08032.

\bibitem[{Child et~al.(2019)Child, Gray, Radford, and Sutskever}]{child2019generating}
Child, R., Gray, S., Radford, A., Sutskever, I., 2019. Generating long sequences with sparse transformers. arXiv preprint arXiv:1904.10509.

\bibitem[{Chiu et~al.(2025)Chiu, Hachiuma, Wang, Smith, Wang, and Chen}]{chiu2025v2v}
Chiu, H.-k., Hachiuma, R., Wang, C.-Y., Smith, S.~F., Wang, Y.-C.~F., Chen, M.-H., 2025. V2v-llm: Vehicle-to-vehicle cooperative autonomous driving with multi-modal large language models. arXiv preprint arXiv:2502.09980.

\bibitem[{Choi et~al.(2024)Choi, Jin, Ham, Kim, and Sun}]{choi2024gentle}
Choi, S., Jin, Z., Ham, S.~W., Kim, J., Sun, L., 2024. A gentle introduction and tutorial on deep generative models in transportation research. arXiv preprint arXiv:2410.07066.

\bibitem[{Chowdhery et~al.(2023)Chowdhery, Narang, Devlin, Bosma, Mishra, Roberts, Barham, Chung, Sutton, Gehrmann, et~al.}]{chowdhery2023palm}
Chowdhery, A., Narang, S., Devlin, J., Bosma, M., Mishra, G., Roberts, A., Barham, P., Chung, H.~W., Sutton, C., Gehrmann, S., et~al., 2023. Palm: Scaling language modeling with pathways. Journal of Machine Learning Research 24~(240), 1--113.

\bibitem[{Christiano et~al.(2017)Christiano, Leike, Brown, Martic, Legg, and Amodei}]{christiano2017deep}
Christiano, P.~F., Leike, J., Brown, T., Martic, M., Legg, S., Amodei, D., 2017. Deep reinforcement learning from human preferences. Advances in neural information processing systems 30, 4299--4307.

\bibitem[{Chu et~al.(2019)Chu, Wang, Codec{\`a}, and Li}]{chu2019multi}
Chu, T., Wang, J., Codec{\`a}, L., Li, Z., 2019. Multi-agent deep reinforcement learning for large-scale traffic signal control. IEEE transactions on intelligent transportation systems 21~(3), 1086--1095.

\bibitem[{Chung et~al.(2024)Chung, Hou, Longpre, Zoph, Tay, Fedus, Li, Wang, Dehghani, Brahma, et~al.}]{chung2024scaling}
Chung, H.~W., Hou, L., Longpre, S., Zoph, B., Tay, Y., Fedus, W., Li, Y., Wang, X., Dehghani, M., Brahma, S., et~al., 2024. Scaling instruction-finetuned language models. Journal of Machine Learning Research 25~(70), 1--53.

\bibitem[{Cowls et~al.(2021)Cowls, Tsamados, Taddeo, and Floridi}]{cowls2021definition}
Cowls, J., Tsamados, A., Taddeo, M., Floridi, L., 2021. A definition, benchmark and database of ai for social good initiatives. Nature Machine Intelligence 3~(2), 111--115.

\bibitem[{Cui et~al.(2024)Cui, Ma, Cao, Ye, and Wang}]{cui2024drive}
Cui, C., Ma, Y., Cao, X., Ye, W., Wang, Z., 2024. Drive as you speak: Enabling human-like interaction with large language models in autonomous vehicles. In: Proceedings of the IEEE/CVF Winter Conference on Applications of Computer Vision. pp. 902--909.

\bibitem[{Cui et~al.(2023)Cui, Huang, Zhong, Liu, Wang, Sun, Li, Wang, and Khajepour}]{cui2023drivellm}
Cui, Y., Huang, S., Zhong, J., Liu, Z., Wang, Y., Sun, C., Li, B., Wang, X., Khajepour, A., 2023. Drivellm: Charting the path toward full autonomous driving with large language models. IEEE Transactions on Intelligent Vehicles 9~(1), 1450--1464.

\bibitem[{Da et~al.(2024{\natexlab{a}})Da, Gao, Mei, and Wei}]{da2024prompt}
Da, L., Gao, M., Mei, H., Wei, H., 2024{\natexlab{a}}. Prompt to transfer: Sim-to-real transfer for traffic signal control with prompt learning. In: Proceedings of the AAAI Conference on Artificial Intelligence. Vol.~38. pp. 82--90.

\bibitem[{Da et~al.(2024{\natexlab{b}})Da, Liou, Chen, Zhou, Luo, Yang, and Wei}]{da2024open}
Da, L., Liou, K., Chen, T., Zhou, X., Luo, X., Yang, Y., Wei, H., 2024{\natexlab{b}}. Open-ti: Open traffic intelligence with augmented language model. International Journal of Machine Learning and Cybernetics 15~(10), 4761--4786.

\bibitem[{Daganzo(1997)}]{daganzo1997fundamentals}
Daganzo, C.~F., 1997. Fundamentals of transportation and traffic operations. Emerald Group Publishing Limited.

\bibitem[{Dao et~al.(2022)Dao, Fu, Ermon, Rudra, and R{\'e}}]{dao2022flashattention}
Dao, T., Fu, D., Ermon, S., Rudra, A., R{\'e}, C., 2022. Flashattention: Fast and memory-efficient exact attention with io-awareness. Advances in neural information processing systems 35, 16344--16359.

\bibitem[{Das et~al.(2025)Das, Amini, and Wu}]{das2025security}
Das, B.~C., Amini, M.~H., Wu, Y., 2025. Security and privacy challenges of large language models: A survey. ACM Computing Surveys 57~(6), 1--39.

\bibitem[{Das et~al.(2023)Das, Oliaee, Le, Pratt, and Wu}]{das2023classifying}
Das, S., Oliaee, A.~H., Le, M., Pratt, M.~P., Wu, J., 2023. Classifying pedestrian maneuver types using the advanced language model. Transportation research record 2677~(7), 599--611.

\bibitem[{De~Palma and Lindsey(2011)}]{de2011traffic}
De~Palma, A., Lindsey, R., 2011. Traffic congestion pricing methodologies and technologies. Transportation Research Part C: Emerging Technologies 19~(6), 1377--1399.

\bibitem[{de~Zarz{\`a} et~al.(2023)de~Zarz{\`a}, de~Curt{\`o}, Roig, and Calafate}]{de2023llm}
de~Zarz{\`a}, I., de~Curt{\`o}, J., Roig, G., Calafate, C.~T., 2023. Llm multimodal traffic accident forecasting. Sensors 23~(22), 9225.

\bibitem[{Deng et~al.(2023)Deng, Yao, Tu, Zheng, Zhang, and Zhang}]{deng2023target}
Deng, Y., Yao, J., Tu, Z., Zheng, X., Zhang, M., Zhang, T., 2023. Target: Automated scenario generation from traffic rules for testing autonomous vehicles. arXiv preprint arXiv:2305.06018.

\bibitem[{Dettmers et~al.(2023)Dettmers, Pagnoni, Holtzman, and Zettlemoyer}]{dettmers2023qlora}
Dettmers, T., Pagnoni, A., Holtzman, A., Zettlemoyer, L., 2023. Qlora: Efficient finetuning of quantized llms. Advances in neural information processing systems 36, 10088--10115.

\bibitem[{Devlin et~al.(2019)Devlin, Chang, Lee, and Toutanova}]{devlin2019bert}
Devlin, J., Chang, M.-W., Lee, K., Toutanova, K., 2019. Bert: Pre-training of deep bidirectional transformers for language understanding. In: Proceedings of the 2019 conference of the North American chapter of the association for computational linguistics: human language technologies, volume 1 (long and short papers). pp. 4171--4186.

\bibitem[{Di and Liu(2016)}]{di2016boundedly}
Di, X., Liu, H.~X., 2016. Boundedly rational route choice behavior: A review of models and methodologies. Transportation Research Part B: Methodological 85, 142--179.

\bibitem[{Dimitrakopoulos and Demestichas(2010)}]{dimitrakopoulos2010intelligent}
Dimitrakopoulos, G., Demestichas, P., 2010. Intelligent transportation systems. IEEE Vehicular Technology Magazine 5~(1), 77--84.

\bibitem[{Ding et~al.(2024)Ding, Zhang, Shang, Zhang, Zong, Feng, Yuan, Su, Li, Sukiennik, et~al.}]{ding2024understanding}
Ding, J., Zhang, Y., Shang, Y., Zhang, Y., Zong, Z., Feng, J., Yuan, Y., Su, H., Li, N., Sukiennik, N., et~al., 2024. Understanding world or predicting future? a comprehensive survey of world models. arXiv preprint arXiv:2411.14499.

\bibitem[{Ding et~al.(2021)Ding, Yang, Hong, Zheng, Zhou, Yin, Lin, Zou, Shao, Yang, et~al.}]{ding2021cogview}
Ding, M., Yang, Z., Hong, W., Zheng, W., Zhou, C., Yin, D., Lin, J., Zou, X., Shao, Z., Yang, H., et~al., 2021. Cogview: Mastering text-to-image generation via transformers. Advances in neural information processing systems 34, 19822--19835.

\bibitem[{Ding et~al.(2023)Ding, Qin, Yang, Wei, Yang, Su, Hu, Chen, Chan, Chen, et~al.}]{ding2023parameter}
Ding, N., Qin, Y., Yang, G., Wei, F., Yang, Z., Su, Y., Hu, S., Chen, Y., Chan, C.-M., Chen, W., et~al., 2023. Parameter-efficient fine-tuning of large-scale pre-trained language models. Nature Machine Intelligence 5~(3), 220--235.

\bibitem[{Dong et~al.(2022)Dong, Li, Dai, Zheng, Ma, Li, Xia, Xu, Wu, Liu, et~al.}]{dong2022survey}
Dong, Q., Li, L., Dai, D., Zheng, C., Ma, J., Li, R., Xia, H., Xu, J., Wu, Z., Liu, T., et~al., 2022. A survey on in-context learning. arXiv preprint arXiv:2301.00234.

\bibitem[{Du et~al.(2022)Du, Huang, Dai, Tong, Lepikhin, Xu, Krikun, Zhou, Yu, Firat, et~al.}]{du2022glam}
Du, N., Huang, Y., Dai, A.~M., Tong, S., Lepikhin, D., Xu, Y., Krikun, M., Zhou, Y., Yu, A.~W., Firat, O., et~al., 2022. Glam: Efficient scaling of language models with mixture-of-experts. In: International conference on machine learning. PMLR, pp. 5547--5569.

\bibitem[{Dubey et~al.(2024)Dubey, Jauhri, Pandey, Kadian, Al-Dahle, Letman, Mathur, Schelten, Yang, Fan, et~al.}]{dubey2024llama}
Dubey, A., Jauhri, A., Pandey, A., Kadian, A., Al-Dahle, A., Letman, A., Mathur, A., Schelten, A., Yang, A., Fan, A., et~al., 2024. The llama 3 herd of models. arXiv preprint arXiv:2407.21783.

\bibitem[{Dwivedi et~al.(2023)Dwivedi, Dave, Naik, Singhal, Omer, Patel, Qian, Wen, Shah, Morgan, et~al.}]{dwivedi2023explainable}
Dwivedi, R., Dave, D., Naik, H., Singhal, S., Omer, R., Patel, P., Qian, B., Wen, Z., Shah, T., Morgan, G., et~al., 2023. Explainable ai (xai): Core ideas, techniques, and solutions. ACM Computing Surveys 55~(9), 1--33.

\bibitem[{Ekin(2023)}]{ekin2023prompt}
Ekin, S., 2023. Prompt engineering for chatgpt: a quick guide to techniques, tips, and best practices. Authorea Preprints.

\bibitem[{El~Faouzi et~al.(2011)El~Faouzi, Leung, and Kurian}]{el2011data}
El~Faouzi, N.-E., Leung, H., Kurian, A., 2011. Data fusion in intelligent transportation systems: Progress and challenges--a survey. Information Fusion 12~(1), 4--10.

\bibitem[{Esteban et~al.(2025)Esteban, Jannik, Uhlemann, and Lienkamp}]{esteban2025scenario}
Esteban, R., Jannik, L., Uhlemann, N., Lienkamp, M., 2025. Scenario understanding of traffic scenes through large visual language models. arXiv preprint arXiv:2501.17131.

\bibitem[{Fan et~al.(2024)Fan, Wang, Zhao, Zhao, Ivanovic, Wang, Pavone, and Yang}]{fan2024learning}
Fan, Z., Wang, P., Zhao, Y., Zhao, Y., Ivanovic, B., Wang, Z., Pavone, M., Yang, H.~F., 2024. Learning traffic crashes as language: Datasets, benchmarks, and what-if causal analyses. arXiv preprint arXiv:2406.10789.

\bibitem[{Fang et~al.(2025)Fang, Xiang, Pan, Salim, and Chen}]{fang2025spatiotemporal}
Fang, L., Xiang, W., Pan, S., Salim, F.~D., Chen, Y.-P.~P., 2025. Spatiotemporal pre-trained large language model for forecasting with missing values. IEEE Internet of Things Journal, 1--1.

\bibitem[{Fang et~al.(2024)Fang, Liu, Ding, Cui, Lv, Hang, and Sun}]{fang2024towards}
Fang, S., Liu, J., Ding, M., Cui, Y., Lv, C., Hang, P., Sun, J., 2024. Towards interactive and learnable cooperative driving automation: a large language model-driven decision-making framework. arXiv preprint arXiv:2409.12812.

\bibitem[{Farazi et~al.(2021)Farazi, Zou, Ahamed, and Barua}]{farazi2021deep}
Farazi, N.~P., Zou, B., Ahamed, T., Barua, L., 2021. Deep reinforcement learning in transportation research: A review. Transportation research interdisciplinary perspectives 11, 100425.

\bibitem[{Feder et~al.(2021)Feder, Oved, Shalit, and Reichart}]{feder2021causalm}
Feder, A., Oved, N., Shalit, U., Reichart, R., 2021. Causalm: Causal model explanation through counterfactual language models. Computational Linguistics 47~(2), 333--386.

\bibitem[{Fedus et~al.(2022)Fedus, Zoph, and Shazeer}]{fedus2022switch}
Fedus, W., Zoph, B., Shazeer, N., 2022. Switch transformers: Scaling to trillion parameter models with simple and efficient sparsity. Journal of Machine Learning Research 23~(120), 1--39.

\bibitem[{Feng et~al.(2025)Feng, Wang, and Yang}]{feng2025survey}
Feng, T., Wang, W., Yang, Y., 2025. A survey of world models for autonomous driving. arXiv preprint arXiv:2501.11260.

\bibitem[{Ferrara(2023)}]{ferrara2023should}
Ferrara, E., 2023. Should chatgpt be biased? challenges and risks of bias in large language models. arXiv preprint arXiv:2304.03738.

\bibitem[{Fries et~al.(2012)Fries, Gahrooei, Chowdhury, and Conway}]{fries2012meeting}
Fries, R.~N., Gahrooei, M.~R., Chowdhury, M., Conway, A.~J., 2012. Meeting privacy challenges while advancing intelligent transportation systems. Transportation Research Part C: Emerging Technologies 25, 34--45.

\bibitem[{Fu et~al.(2024{\natexlab{a}})Fu, Li, Wen, Dou, Cai, Shi, and Qiao}]{fu2024drive}
Fu, D., Li, X., Wen, L., Dou, M., Cai, P., Shi, B., Qiao, Y., 2024{\natexlab{a}}. Drive like a human: Rethinking autonomous driving with large language models. In: Proceedings of the IEEE/CVF Winter Conference on Applications of Computer Vision. pp. 910--919.

\bibitem[{Fu(2001)}]{fu2001adaptive}
Fu, L., 2001. An adaptive routing algorithm for in-vehicle route guidance systems with real-time information. Transportation Research Part B: Methodological 35~(8), 749--765.

\bibitem[{Fu et~al.(2024{\natexlab{b}})Fu, Jain, Chen, Mo, and Di}]{fu2024drivegenvlm}
Fu, Y., Jain, A., Chen, X., Mo, Z., Di, X., 2024{\natexlab{b}}. Drivegenvlm: Real-world video generation for vision language model based autonomous driving. In: 2024 IEEE International Automated Vehicle Validation Conference (IAVVC). IEEE, pp. 1--6.

\bibitem[{Gan et~al.(2024)Gan, Chu, Li, Tang, and Li}]{gan2024large}
Gan, L., Chu, W., Li, G., Tang, X., Li, K., 2024. Large models for intelligent transportation systems and autonomous vehicles: A survey. Advanced Engineering Informatics 62, 102786.

\bibitem[{Ganin et~al.(2019)Ganin, Mersky, Jin, Kitsak, Keisler, and Linkov}]{ganin2019resilience}
Ganin, A.~A., Mersky, A.~C., Jin, A.~S., Kitsak, M., Keisler, J.~M., Linkov, I., 2019. Resilience in intelligent transportation systems (its). Transportation Research Part C: Emerging Technologies 100, 318--329.

\bibitem[{Gao et~al.(2024{\natexlab{a}})Gao, Lan, Li, Yuan, Ding, Zhou, Xu, and Li}]{gao2024large}
Gao, C., Lan, X., Li, N., Yuan, Y., Ding, J., Zhou, Z., Xu, F., Li, Y., 2024{\natexlab{a}}. Large language models empowered agent-based modeling and simulation: A survey and perspectives. Humanities and Social Sciences Communications 11~(1), 1--24.

\bibitem[{Gao et~al.(2023{\natexlab{a}})Gao, Madaan, Zhou, Alon, Liu, Yang, Callan, and Neubig}]{gao2023pal}
Gao, L., Madaan, A., Zhou, S., Alon, U., Liu, P., Yang, Y., Callan, J., Neubig, G., 2023{\natexlab{a}}. Pal: Program-aided language models. In: International Conference on Machine Learning. PMLR, pp. 10764--10799.

\bibitem[{Gao et~al.(2023{\natexlab{b}})Gao, Chen, Xie, Hong, Li, Yeung, and Xu}]{gao2023magicdrive}
Gao, R., Chen, K., Xie, E., Hong, L., Li, Z., Yeung, D.-Y., Xu, Q., 2023{\natexlab{b}}. Magicdrive: Street view generation with diverse 3d geometry control. arXiv preprint arXiv:2310.02601.

\bibitem[{Gao et~al.(2024{\natexlab{b}})Gao, Yang, Chen, Chitta, Qiu, Geiger, Zhang, and Li}]{gao2024vista}
Gao, S., Yang, J., Chen, L., Chitta, K., Qiu, Y., Geiger, A., Zhang, J., Li, H., 2024{\natexlab{b}}. Vista: A generalizable driving world model with high fidelity and versatile controllability. arXiv preprint arXiv:2405.17398.

\bibitem[{Gao et~al.(2023{\natexlab{c}})Gao, Xiong, Gao, Jia, Pan, Bi, Dai, Sun, Wang, and Wang}]{gao2023retrieval}
Gao, Y., Xiong, Y., Gao, X., Jia, K., Pan, J., Bi, Y., Dai, Y., Sun, J., Wang, H., Wang, H., 2023{\natexlab{c}}. Retrieval-augmented generation for large language models: A survey. arXiv preprint arXiv:2312.10997 2.

\bibitem[{Ge et~al.(2023)Ge, Zhang, Liu, Zhang, Han, and Gao}]{ge2023model}
Ge, S., Zhang, Y., Liu, L., Zhang, M., Han, J., Gao, J., 2023. Model tells you what to discard: Adaptive kv cache compression for llms. arXiv preprint arXiv:2310.01801.

\bibitem[{Gentili and Mirchandani(2012)}]{gentili2012locating}
Gentili, M., Mirchandani, P.~B., 2012. Locating sensors on traffic networks: Models, challenges and research opportunities. Transportation research part C: emerging technologies 24, 227--255.

\bibitem[{Gipps(1981)}]{gipps1981behavioural}
Gipps, P.~G., 1981. A behavioural car-following model for computer simulation. Transportation research part B: methodological 15~(2), 105--111.

\bibitem[{Giray(2023)}]{giray2023prompt}
Giray, L., 2023. Prompt engineering with chatgpt: a guide for academic writers. Annals of biomedical engineering 51~(12), 2629--2633.

\bibitem[{GLM et~al.(2024)GLM, Zeng, Xu, Wang, Zhang, Yin, Zhang, Rojas, Feng, Zhao, et~al.}]{glm2024chatglm}
GLM, T., Zeng, A., Xu, B., Wang, B., Zhang, C., Yin, D., Zhang, D., Rojas, D., Feng, G., Zhao, H., et~al., 2024. Chatglm: A family of large language models from glm-130b to glm-4 all tools. arXiv preprint arXiv:2406.12793.

\bibitem[{Golob(2003)}]{golob2003structural}
Golob, T.~F., 2003. Structural equation modeling for travel behavior research. Transportation Research Part B: Methodological 37~(1), 1--25.

\bibitem[{Grigorev et~al.(2024)Grigorev, Saleh, and Ou}]{grigorev2024incidentresponsegpt}
Grigorev, A., Saleh, A.-S. M.~K., Ou, Y., 2024. Incidentresponsegpt: Generating traffic incident response plans with generative artificial intelligence. arXiv preprint arXiv:2404.18550.

\bibitem[{Gruver et~al.(2023)Gruver, Finzi, Qiu, and Wilson}]{gruver2023large}
Gruver, N., Finzi, M., Qiu, S., Wilson, A.~G., 2023. Large language models are zero-shot time series forecasters. Advances in Neural Information Processing Systems 36, 19622--19635.

\bibitem[{Gu and Dao(2023)}]{gu2023mamba}
Gu, A., Dao, T., 2023. Mamba: Linear-time sequence modeling with selective state spaces. arXiv preprint arXiv:2312.00752.

\bibitem[{Gu et~al.(2021)Gu, Goel, and R{\'e}}]{gu2021efficiently}
Gu, A., Goel, K., R{\'e}, C., 2021. Efficiently modeling long sequences with structured state spaces. arXiv preprint arXiv:2111.00396.

\bibitem[{Guan et~al.(2024)Guan, Liao, Li, Hu, Yuan, Li, Zhang, and Xu}]{guan2024world}
Guan, Y., Liao, H., Li, Z., Hu, J., Yuan, R., Li, Y., Zhang, G., Xu, C., 2024. World models for autonomous driving: An initial survey. IEEE Transactions on Intelligent Vehicles, 1--17.

\bibitem[{Guerrero-Ib{\'a}{\~n}ez et~al.(2018)Guerrero-Ib{\'a}{\~n}ez, Zeadally, and Contreras-Castillo}]{guerrero2018sensor}
Guerrero-Ib{\'a}{\~n}ez, J., Zeadally, S., Contreras-Castillo, J., 2018. Sensor technologies for intelligent transportation systems. Sensors 18~(4), 1212.

\bibitem[{Guerrero-Ibanez et~al.(2015)Guerrero-Ibanez, Zeadally, and Contreras-Castillo}]{guerrero2015integration}
Guerrero-Ibanez, J.~A., Zeadally, S., Contreras-Castillo, J., 2015. Integration challenges of intelligent transportation systems with connected vehicle, cloud computing, and internet of things technologies. IEEE Wireless Communications 22~(6), 122--128.

\bibitem[{Guo et~al.(2025)Guo, Yang, Zhang, Song, Zhang, Xu, Zhu, Ma, Wang, Bi, et~al.}]{guo2025deepseek}
Guo, D., Yang, D., Zhang, H., Song, J., Zhang, R., Xu, R., Zhu, Q., Ma, S., Wang, P., Bi, X., et~al., 2025. Deepseek-r1: Incentivizing reasoning capability in llms via reinforcement learning. arXiv preprint arXiv:2501.12948.

\bibitem[{Guo et~al.(2019)Guo, Li, and Ban}]{guo2019urban}
Guo, Q., Li, L., Ban, X.~J., 2019. Urban traffic signal control with connected and automated vehicles: A survey. Transportation research part C: emerging technologies 101, 313--334.

\bibitem[{Guo et~al.(2024)Guo, Zhang, Jiang, Peng, Zhu, and Yang}]{guo2024explainable}
Guo, X., Zhang, Q., Jiang, J., Peng, M., Zhu, M., Yang, H.~F., 2024. Towards explainable traffic flow prediction with large language models. Communications in Transportation Research 4, 100150.

\bibitem[{Gurnee and Tegmark(2023)}]{gurnee2023language}
Gurnee, W., Tegmark, M., 2023. Language models represent space and time. arXiv preprint arXiv:2310.02207.

\bibitem[{Guu et~al.(2020)Guu, Lee, Tung, Pasupat, and Chang}]{guu2020retrieval}
Guu, K., Lee, K., Tung, Z., Pasupat, P., Chang, M., 2020. Retrieval augmented language model pre-training. In: International conference on machine learning. PMLR, pp. 3929--3938.

\bibitem[{Hacker et~al.(2023)Hacker, Engel, and Mauer}]{hacker2023regulating}
Hacker, P., Engel, A., Mauer, M., 2023. Regulating chatgpt and other large generative ai models. In: Proceedings of the 2023 ACM conference on fairness, accountability, and transparency. pp. 1112--1123.

\bibitem[{Han et~al.(2024)Han, Yang, Chen, Cai, Chu, and Zhu}]{han2024autoreward}
Han, X., Yang, Q., Chen, X., Cai, Z., Chu, X., Zhu, M., 2024. Autoreward: Closed-loop reward design with large language models for autonomous driving. IEEE Transactions on Intelligent Vehicles, 1--13.

\bibitem[{Han et~al.(2023)Han, Wang, and Leclercq}]{han2023leveraging}
Han, Y., Wang, M., Leclercq, L., 2023. Leveraging reinforcement learning for dynamic traffic control: A survey and challenges for field implementation. Communications in Transportation Research 3, 100104.

\bibitem[{Haydari et~al.(2024)Haydari, Chen, Lai, Zhang, and Chuah}]{haydari2024mobilitygpt}
Haydari, A., Chen, D., Lai, Z., Zhang, M., Chuah, C.-N., 2024. Mobilitygpt: Enhanced human mobility modeling with a gpt model. arXiv preprint arXiv:2402.03264.

\bibitem[{Haydari and Y{\i}lmaz(2020)}]{haydari2020deep}
Haydari, A., Y{\i}lmaz, Y., 2020. Deep reinforcement learning for intelligent transportation systems: A survey. IEEE Transactions on Intelligent Transportation Systems 23~(1), 11--32.

\bibitem[{He et~al.(2024)He, Nie, and Ma}]{he2024geolocation}
He, J., Nie, T., Ma, W., 2024. Geolocation representation from large language models are generic enhancers for spatio-temporal learning. arXiv preprint arXiv:2408.12116.

\bibitem[{He and Vechev(2023)}]{he2023large}
He, J., Vechev, M., 2023. Large language models for code: Security hardening and adversarial testing. In: Proceedings of the 2023 ACM SIGSAC Conference on Computer and Communications Security. pp. 1865--1879.

\bibitem[{Hoffmann et~al.(2022)Hoffmann, Borgeaud, Mensch, Buchatskaya, Cai, Rutherford, Casas, Hendricks, Welbl, Clark, et~al.}]{hoffmann2022training}
Hoffmann, J., Borgeaud, S., Mensch, A., Buchatskaya, E., Cai, T., Rutherford, E., Casas, D. d.~L., Hendricks, L.~A., Welbl, J., Clark, A., et~al., 2022. Training compute-optimal large language models. arXiv preprint arXiv:2203.15556.

\bibitem[{Houlsby et~al.(2019)Houlsby, Giurgiu, Jastrzebski, Morrone, De~Laroussilhe, Gesmundo, Attariyan, and Gelly}]{houlsby2019parameter}
Houlsby, N., Giurgiu, A., Jastrzebski, S., Morrone, B., De~Laroussilhe, Q., Gesmundo, A., Attariyan, M., Gelly, S., 2019. Parameter-efficient transfer learning for nlp. In: International conference on machine learning. PMLR, pp. 2790--2799.

\bibitem[{Hu et~al.(2023)Hu, Russell, Yeo, Murez, Fedoseev, Kendall, Shotton, and Corrado}]{hu2023gaia}
Hu, A., Russell, L., Yeo, H., Murez, Z., Fedoseev, G., Kendall, A., Shotton, J., Corrado, G., 2023. Gaia-1: A generative world model for autonomous driving. arXiv preprint arXiv:2309.17080.

\bibitem[{Hu et~al.(2022)Hu, Shen, Wallis, Allen-Zhu, Li, Wang, Wang, Chen, et~al.}]{hu2022lora}
Hu, E.~J., Shen, Y., Wallis, P., Allen-Zhu, Z., Li, Y., Wang, S., Wang, L., Chen, W., et~al., 2022. Lora: Low-rank adaptation of large language models. ICLR 1~(2), 3.

\bibitem[{Huang and Chang(2022)}]{huang2022towards}
Huang, J., Chang, K. C.-C., 2022. Towards reasoning in large language models: A survey. arXiv preprint arXiv:2212.10403.

\bibitem[{Huang et~al.(2024{\natexlab{a}})Huang, Feng, Yan, Xiao, Jie, Zhong, Liang, and Ma}]{huang2024drivemm}
Huang, Z., Feng, C., Yan, F., Xiao, B., Jie, Z., Zhong, Y., Liang, X., Ma, L., 2024{\natexlab{a}}. Drivemm: All-in-one large multimodal model for autonomous driving. arXiv preprint arXiv:2412.07689.

\bibitem[{Huang et~al.(2024{\natexlab{b}})Huang, Sheng, Qu, You, and Chen}]{huang2024vlm}
Huang, Z., Sheng, Z., Qu, Y., You, J., Chen, S., 2024{\natexlab{b}}. Vlm-rl: A unified vision language models and reinforcement learning framework for safe autonomous driving. arXiv preprint arXiv:2412.15544.

\bibitem[{Janson(1991)}]{janson1991dynamic}
Janson, B.~N., 1991. Dynamic traffic assignment for urban road networks. Transportation Research Part B: Methodological 25~(2-3), 143--161.

\bibitem[{Jiang et~al.(2023)Jiang, Sablayrolles, Mensch, Bamford, Chaplot, de~las Casas, Bressand, Lengyel, Lample, Saulnier, Lavaud, Lachaux, Stock, Scao, Lavril, Wang, Lacroix, and Sayed}]{jiang2023mistral7b}
Jiang, A.~Q., Sablayrolles, A., Mensch, A., Bamford, C., Chaplot, D.~S., de~las Casas, D., Bressand, F., Lengyel, G., Lample, G., Saulnier, L., Lavaud, L.~R., Lachaux, M.-A., Stock, P., Scao, T.~L., Lavril, T., Wang, T., Lacroix, T., Sayed, W.~E., 2023. Mistral 7b.
\newline\urlprefix\url{https://arxiv.org/abs/2310.06825}

\bibitem[{Jiang et~al.(2024)Jiang, Cai, Cui, Li, Ren, Yu, Yang, Fu, Wen, and Cai}]{jiang2024koma}
Jiang, K., Cai, X., Cui, Z., Li, A., Ren, Y., Yu, H., Yang, H., Fu, D., Wen, L., Cai, P., 2024. Koma: Knowledge-driven multi-agent framework for autonomous driving with large language models. IEEE Transactions on Intelligent Vehicles, 1--15.

\bibitem[{Jiang et~al.(2020)Jiang, Xu, Araki, and Neubig}]{jiang2020can}
Jiang, Z., Xu, F.~F., Araki, J., Neubig, G., 2020. How can we know what language models know? Transactions of the Association for Computational Linguistics 8, 423--438.

\bibitem[{Jin et~al.(2021)Jin, Wi, Lee, Kang, Kim, and Kim}]{jin2021trafficbert}
Jin, K., Wi, J., Lee, E., Kang, S., Kim, S., Kim, Y., 2021. Trafficbert: Pre-trained model with large-scale data for long-range traffic flow forecasting. Expert Systems with Applications 186, 115738.

\bibitem[{Jin et~al.(2023{\natexlab{a}})Jin, Wang, Ma, Chu, Zhang, Shi, Chen, Liang, Li, Pan, et~al.}]{jin2023time}
Jin, M., Wang, S., Ma, L., Chu, Z., Zhang, J.~Y., Shi, X., Chen, P.-Y., Liang, Y., Li, Y.-F., Pan, S., et~al., 2023{\natexlab{a}}. Time-llm: Time series forecasting by reprogramming large language models. arXiv preprint arXiv:2310.01728.

\bibitem[{Jin et~al.(2023{\natexlab{b}})Jin, Wen, Liang, Zhang, Xue, Wang, Zhang, Wang, Chen, Li, et~al.}]{jin2023large}
Jin, M., Wen, Q., Liang, Y., Zhang, C., Xue, S., Wang, X., Zhang, J., Wang, Y., Chen, H., Li, X., et~al., 2023{\natexlab{b}}. Large models for time series and spatio-temporal data: A survey and outlook. arXiv preprint arXiv:2310.10196.

\bibitem[{Jin et~al.(2023{\natexlab{c}})Jin, Shen, Peng, Liu, Qin, Li, Xie, Gao, Zhou, and Gong}]{jin2023surrealdriver}
Jin, Y., Shen, X., Peng, H., Liu, X., Qin, J., Li, J., Xie, J., Gao, P., Zhou, G., Gong, J., 2023{\natexlab{c}}. Surrealdriver: Designing generative driver agent simulation framework in urban contexts based on large language model. arXiv preprint arXiv:2309.13193.

\bibitem[{Kaddour et~al.(2023)Kaddour, Harris, Mozes, Bradley, Raileanu, and McHardy}]{kaddour2023challenges}
Kaddour, J., Harris, J., Mozes, M., Bradley, H., Raileanu, R., McHardy, R., 2023. Challenges and applications of large language models. arXiv preprint arXiv:2307.10169.

\bibitem[{Kanarachos et~al.(2018)Kanarachos, Christopoulos, and Chroneos}]{kanarachos2018smartphones}
Kanarachos, S., Christopoulos, S.-R.~G., Chroneos, A., 2018. Smartphones as an integrated platform for monitoring driver behaviour: The role of sensor fusion and connectivity. Transportation research part C: emerging technologies 95, 867--882.

\bibitem[{Kaplan et~al.(2020)Kaplan, McCandlish, Henighan, Brown, Chess, Child, Gray, Radford, Wu, and Amodei}]{kaplan2020scaling}
Kaplan, J., McCandlish, S., Henighan, T., Brown, T.~B., Chess, B., Child, R., Gray, S., Radford, A., Wu, J., Amodei, D., 2020. Scaling laws for neural language models. arXiv preprint arXiv:2001.08361.

\bibitem[{Keysan et~al.(2023)Keysan, Look, Kosman, G{\"u}rsun, Wagner, Yao, and Rakitsch}]{keysan2023can}
Keysan, A., Look, A., Kosman, E., G{\"u}rsun, G., Wagner, J., Yao, Y., Rakitsch, B., 2023. Can you text what is happening? integrating pre-trained language encoders into trajectory prediction models for autonomous driving. arXiv preprint arXiv:2309.05282.

\bibitem[{Kim et~al.(2023)Kim, Lee, Kim, Park, Yoo, Kwon, and Lee}]{kim2023memory}
Kim, J., Lee, J.~H., Kim, S., Park, J., Yoo, K.~M., Kwon, S.~J., Lee, D., 2023. Memory-efficient fine-tuning of compressed large language models via sub-4-bit integer quantization. Advances in Neural Information Processing Systems 36, 36187--36207.

\bibitem[{Kim et~al.(2018)Kim, Rohrbach, Darrell, Canny, and Akata}]{kim2018textual}
Kim, J., Rohrbach, A., Darrell, T., Canny, J., Akata, Z., 2018. Textual explanations for self-driving vehicles. In: Proceedings of the European conference on computer vision (ECCV). pp. 563--578.

\bibitem[{Kojima et~al.(2022)Kojima, Gu, Reid, Matsuo, and Iwasawa}]{kojima2022large}
Kojima, T., Gu, S.~S., Reid, M., Matsuo, Y., Iwasawa, Y., 2022. Large language models are zero-shot reasoners. Advances in neural information processing systems 35, 22199--22213.

\bibitem[{Kuang et~al.(2024)Kuang, Liu, Wang, Wu, and Wei}]{kuang2024harnessing}
Kuang, S., Liu, Y., Wang, X., Wu, X., Wei, Y., 2024. Harnessing multimodal large language models for traffic knowledge graph generation and decision-making. Communications in Transportation Research 4, 100146.

\bibitem[{Kumar et~al.(2023)Kumar, Agarwal, Srinivas, Li, Feizi, and Lakkaraju}]{kumar2023certifying}
Kumar, A., Agarwal, C., Srinivas, S., Li, A.~J., Feizi, S., Lakkaraju, H., 2023. Certifying llm safety against adversarial prompting. arXiv preprint arXiv:2309.02705.

\bibitem[{Kumar and Raubal(2021)}]{kumar2021applications}
Kumar, N., Raubal, M., 2021. Applications of deep learning in congestion detection, prediction and alleviation: A survey. Transportation Research Part C: Emerging Technologies 133, 103432.

\bibitem[{Kwon et~al.(2023)Kwon, Xie, Bullard, and Sadigh}]{kwon2023reward}
Kwon, M., Xie, S.~M., Bullard, K., Sadigh, D., 2023. Reward design with language models. arXiv preprint arXiv:2303.00001.

\bibitem[{Lai et~al.(2023)Lai, Xu, Zhang, Liu, and Xiong}]{lai2023large}
Lai, S., Xu, Z., Zhang, W., Liu, H., Xiong, H., 2023. Large language models as traffic signal control agents: Capacity and opportunity. arXiv preprint arXiv:2312.16044.

\bibitem[{Lan et~al.(2024)Lan, Liu, Fan, Lv, Ren, and Cui}]{lan2024traj}
Lan, Z., Liu, L., Fan, B., Lv, Y., Ren, Y., Cui, Z., 2024. Traj-llm: A new exploration for empowering trajectory prediction with pre-trained large language models. IEEE Transactions on Intelligent Vehicles, 1--14.

\bibitem[{Lester et~al.(2021)Lester, Al-Rfou, and Constant}]{lester2021power}
Lester, B., Al-Rfou, R., Constant, N., 2021. The power of scale for parameter-efficient prompt tuning. arXiv preprint arXiv:2104.08691.

\bibitem[{Lewis et~al.(2019)Lewis, Liu, Goyal, Ghazvininejad, Mohamed, Levy, Stoyanov, and Zettlemoyer}]{lewis2019bart}
Lewis, M., Liu, Y., Goyal, N., Ghazvininejad, M., Mohamed, A., Levy, O., Stoyanov, V., Zettlemoyer, L., 2019. Bart: Denoising sequence-to-sequence pre-training for natural language generation, translation, and comprehension. arXiv preprint arXiv:1910.13461.

\bibitem[{Lewis et~al.(2020)Lewis, Perez, Piktus, Petroni, Karpukhin, Goyal, K{\"u}ttler, Lewis, Yih, Rockt{\"a}schel, et~al.}]{lewis2020retrieval}
Lewis, P., Perez, E., Piktus, A., Petroni, F., Karpukhin, V., Goyal, N., K{\"u}ttler, H., Lewis, M., Yih, W.-t., Rockt{\"a}schel, T., et~al., 2020. Retrieval-augmented generation for knowledge-intensive nlp tasks. Advances in neural information processing systems 33, 9459--9474.

\bibitem[{Li et~al.(2021)Li, Cao, and Li}]{li2021reinforcement}
Li, M., Cao, Z., Li, Z., 2021. A reinforcement learning-based vehicle platoon control strategy for reducing energy consumption in traffic oscillations. IEEE Transactions on Neural Networks and Learning Systems 32~(12), 5309--5322.

\bibitem[{Li et~al.(2023{\natexlab{a}})Li, Peng, Feng, Liu, Duan, Mo, and Zhou}]{li2023scenarionet}
Li, Q., Peng, Z.~M., Feng, L., Liu, Z., Duan, C., Mo, W., Zhou, B., 2023{\natexlab{a}}. Scenarionet: Open-source platform for large-scale traffic scenario simulation and modeling. Advances in neural information processing systems 36, 3894--3920.

\bibitem[{Li et~al.(2024{\natexlab{a}})Li, Azfar, and Ke}]{li2024chatsumo}
Li, S., Azfar, T., Ke, R., 2024{\natexlab{a}}. Chatsumo: Large language model for automating traffic scenario generation in simulation of urban mobility. IEEE Transactions on Intelligent Vehicles, 1--12.

\bibitem[{Li et~al.(2024{\natexlab{b}})Li, Liu, Shen, Huang, and Wang}]{li2024chatgpt}
Li, X., Liu, E., Shen, T., Huang, J., Wang, F.-Y., 2024{\natexlab{b}}. Chatgpt-based scenario engineer: A new framework on scenario generation for trajectory prediction. IEEE Transactions on Intelligent Vehicles 9~(3), 4422--4431.

\bibitem[{Li and Liang(2021)}]{li2021prefix}
Li, X.~L., Liang, P., 2021. Prefix-tuning: Optimizing continuous prompts for generation. arXiv preprint arXiv:2101.00190.

\bibitem[{Li et~al.(2023{\natexlab{b}})Li, Zhang, and Sun}]{li2023metaagents}
Li, Y., Zhang, Y., Sun, L., 2023{\natexlab{b}}. Metaagents: Simulating interactions of human behaviors for llm-based task-oriented coordination via collaborative generative agents. arXiv preprint arXiv:2310.06500.

\bibitem[{Li et~al.(2024{\natexlab{c}})Li, Xia, Tang, Xu, Shi, Xia, Yin, and Huang}]{li2024urbangpt}
Li, Z., Xia, L., Tang, J., Xu, Y., Shi, L., Xia, L., Yin, D., Huang, C., 2024{\natexlab{c}}. Urbangpt: Spatio-temporal large language models. In: Proceedings of the 30th ACM SIGKDD Conference on Knowledge Discovery and Data Mining. pp. 5351--5362.

\bibitem[{Liang et~al.(2024{\natexlab{a}})Liang, Liu, Wang, and Zhao}]{liang2024exploring}
Liang, Y., Liu, Y., Wang, X., Zhao, Z., 2024{\natexlab{a}}. Exploring large language models for human mobility prediction under public events. Computers, Environment and Urban Systems 112, 102153.

\bibitem[{Liang et~al.(2024{\natexlab{b}})Liang, Xu, Hong, Shang, Wang, Fu, and Liu}]{liang2024survey}
Liang, Z., Xu, Y., Hong, Y., Shang, P., Wang, Q., Fu, Q., Liu, K., 2024{\natexlab{b}}. A survey of multimodel large language models. In: Proceedings of the 3rd International Conference on Computer, Artificial Intelligence and Control Engineering. pp. 405--409.

\bibitem[{Liao et~al.(2024)Liao, Shen, Li, Wang, Li, Bie, and Xu}]{liao2024gpt}
Liao, H., Shen, H., Li, Z., Wang, C., Li, G., Bie, Y., Xu, C., 2024. Gpt-4 enhanced multimodal grounding for autonomous driving: Leveraging cross-modal attention with large language models. Communications in Transportation Research 4, 100116.

\bibitem[{Lin et~al.(2023{\natexlab{a}})Lin, Liu, Li, and Qu}]{lin2023generative}
Lin, H., Liu, Y., Li, S., Qu, X., 2023{\natexlab{a}}. How generative adversarial networks promote the development of intelligent transportation systems: A survey. IEEE/CAA journal of automatica sinica 10~(9), 1781--1796.

\bibitem[{Lin et~al.(2024{\natexlab{a}})Lin, Tang, Tang, Yang, Chen, Wang, Xiao, Dang, Gan, and Han}]{lin2024awq}
Lin, J., Tang, J., Tang, H., Yang, S., Chen, W.-M., Wang, W.-C., Xiao, G., Dang, X., Gan, C., Han, S., 2024{\natexlab{a}}. Awq: Activation-aware weight quantization for on-device llm compression and acceleration. Proceedings of Machine Learning and Systems 6, 87--100.

\bibitem[{Lin et~al.(2024{\natexlab{b}})Lin, Li, Ding, Tomizuka, Zhan, and Althoff}]{lin2024drplanner}
Lin, Y., Li, C., Ding, M., Tomizuka, M., Zhan, W., Althoff, M., 2024{\natexlab{b}}. Drplanner: Diagnosis and repair of motion planners for automated vehicles using large language models. IEEE Robotics and Automation Letters 9~(10), 8218--8225.

\bibitem[{Lin et~al.(2023{\natexlab{b}})Lin, Trivedi, and Sun}]{lin2023generating}
Lin, Z., Trivedi, S., Sun, J., 2023{\natexlab{b}}. Generating with confidence: Uncertainty quantification for black-box large language models. arXiv preprint arXiv:2305.19187.

\bibitem[{Liu et~al.(2024{\natexlab{a}})Liu, Feng, Xue, Wang, Wu, Lu, Zhao, Deng, Zhang, Ruan, et~al.}]{liu2024deepseek}
Liu, A., Feng, B., Xue, B., Wang, B., Wu, B., Lu, C., Zhao, C., Deng, C., Zhang, C., Ruan, C., et~al., 2024{\natexlab{a}}. Deepseek-v3 technical report. arXiv preprint arXiv:2412.19437.

\bibitem[{Liu et~al.(2024{\natexlab{b}})Liu, Liu, Pan, He, Haffari, and Zhuang}]{liu2024minicache}
Liu, A., Liu, J., Pan, Z., He, Y., Haffari, R., Zhuang, B., 2024{\natexlab{b}}. Minicache: Kv cache compression in depth dimension for large language models. Advances in Neural Information Processing Systems 37, 139997--140031.

\bibitem[{Liu et~al.(2024{\natexlab{c}})Liu, Yang, Xu, Li, Long, Li, and Zhao}]{liu2024spatial}
Liu, C., Yang, S., Xu, Q., Li, Z., Long, C., Li, Z., Zhao, R., 2024{\natexlab{c}}. Spatial-temporal large language model for traffic prediction. arXiv preprint arXiv:2401.10134.

\bibitem[{Liu and Zhao(2024)}]{liu2024resource}
Liu, C., Zhao, J., 2024. Resource allocation in large language model integrated 6g vehicular networks. In: 2024 IEEE 99th Vehicular Technology Conference (VTC2024-Spring). IEEE, pp. 1--6.

\bibitem[{Liu et~al.(2024{\natexlab{d}})Liu, Jiang, Miranda-Moreno, Choi, and Sun}]{liu2024adversarial}
Liu, F., Jiang, S., Miranda-Moreno, L., Choi, S., Sun, L., 2024{\natexlab{d}}. Adversarial vulnerabilities in large language models for time series forecasting. arXiv preprint arXiv:2412.08099.

\bibitem[{Liu et~al.(2023{\natexlab{a}})Liu, Dong, Jiang, Deng, Deng, Chen, and Song}]{liu2023spatio}
Liu, H., Dong, Z., Jiang, R., Deng, J., Deng, J., Chen, Q., Song, X., 2023{\natexlab{a}}. Spatio-temporal adaptive embedding makes vanilla transformer sota for traffic forecasting. In: Proceedings of the 32nd ACM international conference on information and knowledge management. pp. 4125--4129.

\bibitem[{Liu et~al.(2022)Liu, Tam, Muqeeth, Mohta, Huang, Bansal, and Raffel}]{liu2022few}
Liu, H., Tam, D., Muqeeth, M., Mohta, J., Huang, T., Bansal, M., Raffel, C.~A., 2022. Few-shot parameter-efficient fine-tuning is better and cheaper than in-context learning. Advances in Neural Information Processing Systems 35, 1950--1965.

\bibitem[{Liu et~al.(2023{\natexlab{b}})Liu, Hang, Qi, Wang, and Sun}]{liu2023mtd}
Liu, J., Hang, P., Qi, X., Wang, J., Sun, J., 2023{\natexlab{b}}. Mtd-gpt: A multi-task decision-making gpt model for autonomous driving at unsignalized intersections. In: 2023 IEEE 26th International Conference on Intelligent Transportation Systems (ITSC). IEEE, pp. 5154--5161.

\bibitem[{Liu et~al.(2024{\natexlab{e}})Liu, Yu, Wang, Ma, and Shen}]{liu2024stgllm}
Liu, L., Yu, S., Wang, R., Ma, Z., Shen, Y., 2024{\natexlab{e}}. How can large language models understand spatial-temporal data? arXiv preprint arXiv:2401.14192.

\bibitem[{Liu et~al.(2024{\natexlab{f}})Liu, Li, and Yin}]{liu2024can}
Liu, T., Li, M., Yin, Y., 2024{\natexlab{f}}. Can large language models capture human travel behavior? evidence and insights on mode choice. Evidence and Insights on Mode Choice (August 26, 2024).

\bibitem[{Liu et~al.(2021)Liu, Ji, Fu, Tam, Du, Yang, and Tang}]{liu2021p}
Liu, X., Ji, K., Fu, Y., Tam, W.~L., Du, Z., Yang, Z., Tang, J., 2021. P-tuning v2: Prompt tuning can be comparable to fine-tuning universally across scales and tasks. arXiv preprint arXiv:2110.07602.

\bibitem[{Liu et~al.(2024{\natexlab{g}})Liu, Qian, Teo, and Ma}]{liu2024estimating}
Liu, X., Qian, S., Teo, H.-H., Ma, W., 2024{\natexlab{g}}. Estimating and mitigating the congestion effect of curbside pick-ups and drop-offs: A causal inference approach. Transportation Science 58~(2), 355--376.

\bibitem[{Liu et~al.(2024{\natexlab{h}})Liu, Liao, Ma, He, Stanford, and Ma}]{liu2024human}
Liu, Y., Liao, X., Ma, H., He, B.~Y., Stanford, C., Ma, J., 2024{\natexlab{h}}. Human mobility modeling with limited information via large language models. arXiv preprint arXiv:2409.17495.

\bibitem[{Liu et~al.(2024{\natexlab{i}})Liu, Yuan, Jin, Zhong, Xu, Braverman, Chen, and Hu}]{liu2024kivi}
Liu, Z., Yuan, J., Jin, H., Zhong, S., Xu, Z., Braverman, V., Chen, B., Hu, X., 2024{\natexlab{i}}. Kivi: A tuning-free asymmetric 2bit quantization for kv cache. arXiv preprint arXiv:2402.02750.

\bibitem[{Lohner et~al.(2024)Lohner, Compagno, Francis, and Oltramari}]{lohner2024enhancing}
Lohner, A., Compagno, F., Francis, J., Oltramari, A., 2024. Enhancing vision-language models with scene graphs for traffic accident understanding. In: 2024 IEEE International Automated Vehicle Validation Conference (IAVVC). IEEE, pp. 1--7.

\bibitem[{Long et~al.(2024{\natexlab{a}})Long, Shi, Liu, and Li}]{long2024vlm}
Long, K., Shi, H., Liu, J., Li, X., 2024{\natexlab{a}}. Vlm-mpc: Vision language foundation model (vlm)-guided model predictive controller (mpc) for autonomous driving. arXiv preprint arXiv:2408.04821.

\bibitem[{Long et~al.(2024{\natexlab{b}})Long, Yuan, and Li}]{long2024universal}
Long, Q., Yuan, Y., Li, Y., 2024{\natexlab{b}}. A universal model for human mobility prediction. arXiv preprint arXiv:2412.15294.

\bibitem[{Lu et~al.(2024{\natexlab{a}})Lu, Huang, Yang, Zhang, and Zhang}]{lu2024wovogen}
Lu, J., Huang, Z., Yang, Z., Zhang, J., Zhang, L., 2024{\natexlab{a}}. Wovogen: World volume-aware diffusion for controllable multi-camera driving scene generation. In: European Conference on Computer Vision. Springer, pp. 329--345.

\bibitem[{Lu et~al.(2024{\natexlab{b}})Lu, Wang, Jiang, Zhao, Ma, and Feng}]{lu2024multimodal}
Lu, Q., Wang, X., Jiang, Y., Zhao, G., Ma, M., Feng, S., 2024{\natexlab{b}}. Multimodal large language model driven scenario testing for autonomous vehicles. arXiv preprint arXiv:2409.06450.

\bibitem[{Lv(2023)}]{lv2023large}
Lv, Y., 2023. Large transportation models on the horizon: Challenges and issues [editor’s column]. IEEE Intelligent Transportation Systems Magazine 15~(6), 3--5.

\bibitem[{Ma et~al.(2020)Ma, Pi, and Qian}]{ma2020estimating}
Ma, W., Pi, X., Qian, S., 2020. Estimating multi-class dynamic origin-destination demand through a forward-backward algorithm on computational graphs. Transportation Research Part C: Emerging Technologies 119, 102747.

\bibitem[{Ma and Qian(2018)}]{ma2018estimating}
Ma, W., Qian, Z.~S., 2018. Estimating multi-year 24/7 origin-destination demand using high-granular multi-source traffic data. Transportation Research Part C: Emerging Technologies 96, 96--121.

\bibitem[{Ma et~al.(2023{\natexlab{a}})Ma, Fang, and Wang}]{ma2023llm}
Ma, X., Fang, G., Wang, X., 2023{\natexlab{a}}. Llm-pruner: On the structural pruning of large language models. Advances in neural information processing systems 36, 21702--21720.

\bibitem[{Ma et~al.(2024)Ma, Cui, Cao, Ye, Liu, Lu, Abdelraouf, Gupta, Han, Bera, et~al.}]{ma2024lampilot}
Ma, Y., Cui, C., Cao, X., Ye, W., Liu, P., Lu, J., Abdelraouf, A., Gupta, R., Han, K., Bera, A., et~al., 2024. Lampilot: An open benchmark dataset for autonomous driving with language model programs. In: Proceedings of the IEEE/CVF Conference on Computer Vision and Pattern Recognition. pp. 15141--15151.

\bibitem[{Ma et~al.(2023{\natexlab{b}})Ma, Liang, Wang, Huang, Bastani, Jayaraman, Zhu, Fan, and Anandkumar}]{ma2023eureka}
Ma, Y.~J., Liang, W., Wang, G., Huang, D.-A., Bastani, O., Jayaraman, D., Zhu, Y., Fan, L., Anandkumar, A., 2023{\natexlab{b}}. Eureka: Human-level reward design via coding large language models. arXiv preprint arXiv:2310.12931.

\bibitem[{Mahmud et~al.(2025)Mahmud, Hajmohamed, Almentheri, Alqaydi, Aldhaheri, Khalil, and Saeed}]{mahmud2025integrating}
Mahmud, D., Hajmohamed, H., Almentheri, S., Alqaydi, S., Aldhaheri, L., Khalil, R.~A., Saeed, N., 2025. Integrating llms with its: Recent advances, potentials, challenges, and future directions. IEEE Transactions on Intelligent Transportation Systems, 1--36.

\bibitem[{Mao et~al.(2023{\natexlab{a}})Mao, Qian, Ye, Zhao, and Wang}]{mao2023gpt}
Mao, J., Qian, Y., Ye, J., Zhao, H., Wang, Y., 2023{\natexlab{a}}. Gpt-driver: Learning to drive with gpt. arXiv preprint arXiv:2310.01415.

\bibitem[{Mao et~al.(2023{\natexlab{b}})Mao, Ye, Qian, Pavone, and Wang}]{mao2023language}
Mao, J., Ye, J., Qian, Y., Pavone, M., Wang, Y., 2023{\natexlab{b}}. A language agent for autonomous driving. arXiv preprint arXiv:2311.10813.

\bibitem[{Masri et~al.(2025)Masri, Ashqar, and Elhenawy}]{masri2025large}
Masri, S., Ashqar, H.~I., Elhenawy, M., 2025. Large language models (llms) as traffic control systems at urban intersections: A new paradigm. Vehicles 7~(1), 11.

\bibitem[{Mei et~al.(2025)Mei, Nie, Sun, and Tian}]{mei2025llm}
Mei, Y., Nie, T., Sun, J., Tian, Y., 2025. Llm-attacker: Enhancing closed-loop adversarial scenario generation for autonomous driving with large language models. arXiv preprint arXiv:2501.15850.

\bibitem[{Micikevicius et~al.(2017)Micikevicius, Narang, Alben, Diamos, Elsen, Garcia, Ginsburg, Houston, Kuchaiev, Venkatesh, et~al.}]{micikevicius2017mixed}
Micikevicius, P., Narang, S., Alben, J., Diamos, G., Elsen, E., Garcia, D., Ginsburg, B., Houston, M., Kuchaiev, O., Venkatesh, G., et~al., 2017. Mixed precision training. arXiv preprint arXiv:1710.03740.

\bibitem[{Min et~al.(2023)Min, Ross, Sulem, Veyseh, Nguyen, Sainz, Agirre, Heintz, and Roth}]{min2023recent}
Min, B., Ross, H., Sulem, E., Veyseh, A. P.~B., Nguyen, T.~H., Sainz, O., Agirre, E., Heintz, I., Roth, D., 2023. Recent advances in natural language processing via large pre-trained language models: A survey. ACM Computing Surveys 56~(2), 1--40.

\bibitem[{Minaee et~al.(2024)Minaee, Mikolov, Nikzad, Chenaghlu, Socher, Amatriain, and Gao}]{minaee2024largelanguagemodelssurvey}
Minaee, S., Mikolov, T., Nikzad, N., Chenaghlu, M., Socher, R., Amatriain, X., Gao, J., 2024. Large language models: A survey. arXiv preprint arXiv:2402.06196.

\bibitem[{Mo et~al.(2023)Mo, Xu, Zhuang, Ma, Guo, and Zhao}]{mo2023large}
Mo, B., Xu, H., Zhuang, D., Ma, R., Guo, X., Zhao, J., 2023. Large language models for travel behavior prediction. arXiv preprint arXiv:2312.00819.

\bibitem[{M{\"o}kander et~al.(2024)M{\"o}kander, Schuett, Kirk, and Floridi}]{mokander2024auditing}
M{\"o}kander, J., Schuett, J., Kirk, H.~R., Floridi, L., 2024. Auditing large language models: a three-layered approach. AI and Ethics 4~(4), 1085--1115.

\bibitem[{Mourad et~al.(2019)Mourad, Puchinger, and Chu}]{mourad2019survey}
Mourad, A., Puchinger, J., Chu, C., 2019. A survey of models and algorithms for optimizing shared mobility. Transportation Research Part B: Methodological 123, 323--346.

\bibitem[{Movahedi and Choi(2025)}]{movahedi2025crossroads}
Movahedi, M., Choi, J., 2025. The crossroads of llm and traffic control: A study on large language models in adaptive traffic signal control. IEEE Transactions on Intelligent Transportation Systems 26~(2), 1701--1716.

\bibitem[{Mumtarin et~al.(2023)Mumtarin, Chowdhury, and Wood}]{mumtarin2023large}
Mumtarin, M., Chowdhury, M.~S., Wood, J., 2023. Large language models in analyzing crash narratives--a comparative study of chatgpt, bard and gpt-4. arXiv preprint arXiv:2308.13563.

\bibitem[{Nakano et~al.(2021)Nakano, Hilton, Balaji, Wu, Ouyang, Kim, Hesse, Jain, Kosaraju, Saunders, et~al.}]{nakano2021webgpt}
Nakano, R., Hilton, J., Balaji, S., Wu, J., Ouyang, L., Kim, C., Hesse, C., Jain, S., Kosaraju, V., Saunders, W., et~al., 2021. Webgpt: Browser-assisted question-answering with human feedback. arXiv preprint arXiv:2112.09332.

\bibitem[{Nasir et~al.(2024)Nasir, Earle, Togelius, James, and Cleghorn}]{nasir2024llmatic}
Nasir, M.~U., Earle, S., Togelius, J., James, S., Cleghorn, C., 2024. Llmatic: neural architecture search via large language models and quality diversity optimization. In: proceedings of the Genetic and Evolutionary Computation Conference. pp. 1110--1118.

\bibitem[{Naveed et~al.(2023)Naveed, Khan, Qiu, Saqib, Anwar, Usman, Akhtar, Barnes, and Mian}]{naveed2023comprehensive}
Naveed, H., Khan, A.~U., Qiu, S., Saqib, M., Anwar, S., Usman, M., Akhtar, N., Barnes, N., Mian, A., 2023. A comprehensive overview of large language models. arXiv preprint arXiv:2307.06435.

\bibitem[{Nie et~al.(2025{\natexlab{a}})Nie, He, Mei, Qin, Li, Sun, and Ma}]{nie2025joint}
Nie, T., He, J., Mei, Y., Qin, G., Li, G., Sun, J., Ma, W., 2025{\natexlab{a}}. Joint estimation and prediction of city-wide delivery demand: A large language model empowered graph-based learning approach. Transportation Research Part E: Logistics and Transportation Review 197, 104075.

\bibitem[{Nie et~al.(2024{\natexlab{a}})Nie, Qin, Ma, Mei, and Sun}]{nie2024imputeformer}
Nie, T., Qin, G., Ma, W., Mei, Y., Sun, J., 2024{\natexlab{a}}. Imputeformer: Low rankness-induced transformers for generalizable spatiotemporal imputation. In: Proceedings of the 30th ACM SIGKDD Conference on Knowledge Discovery and Data Mining. pp. 2260--2271.

\bibitem[{Nie et~al.(2024{\natexlab{b}})Nie, Qin, Ma, and Sun}]{nie2024spatiotemporal}
Nie, T., Qin, G., Ma, W., Sun, J., 2024{\natexlab{b}}. Spatiotemporal implicit neural representation as a generalized traffic data learner. Transportation Research Part C: Emerging Technologies 169, 104890.

\bibitem[{Nie et~al.(2025{\natexlab{b}})Nie, Qin, Sun, Ma, Mei, and Sun}]{nie2025contextualizing}
Nie, T., Qin, G., Sun, L., Ma, W., Mei, Y., Sun, J., 2025{\natexlab{b}}. Contextualizing mlp-mixers spatiotemporally for urban traffic data forecast at scale. IEEE Transactions on Intelligent Transportation Systems 26~(1), 1241--1256.

\bibitem[{Omama et~al.(2023)Omama, Inani, Paul, Yellapragada, Jatavallabhula, Chinchali, and Krishna}]{omama2023alt}
Omama, M., Inani, P., Paul, P., Yellapragada, S.~C., Jatavallabhula, K.~M., Chinchali, S., Krishna, M., 2023. Alt-pilot: Autonomous navigation with language augmented topometric maps. arXiv preprint arXiv:2310.02324.

\bibitem[{Orfanoudakis et~al.(2025)Orfanoudakis, Palensky, and Vergara}]{orfanoudakis2025optimizing}
Orfanoudakis, S., Palensky, P., Vergara, P.~P., 2025. Optimizing electric vehicles charging using large language models and graph neural networks. arXiv preprint arXiv:2502.03067.

\bibitem[{Osorio and Bierlaire(2013)}]{osorio2013simulation}
Osorio, C., Bierlaire, M., 2013. A simulation-based optimization framework for urban transportation problems. Operations Research 61~(6), 1333--1345.

\bibitem[{Osorio and Chong(2015)}]{osorio2015computationally}
Osorio, C., Chong, L., 2015. A computationally efficient simulation-based optimization algorithm for large-scale urban transportation problems. Transportation Science 49~(3), 623--636.

\bibitem[{Ouyang et~al.(2022)Ouyang, Wu, Jiang, Almeida, Wainwright, Mishkin, Zhang, Agarwal, Slama, Ray, et~al.}]{ouyang2022training}
Ouyang, L., Wu, J., Jiang, X., Almeida, D., Wainwright, C., Mishkin, P., Zhang, C., Agarwal, S., Slama, K., Ray, A., et~al., 2022. Training language models to follow instructions with human feedback. Advances in neural information processing systems 35, 27730--27744.

\bibitem[{Pan et~al.(2023)Pan, Albalak, Wang, and Wang}]{pan2023logic}
Pan, L., Albalak, A., Wang, X., Wang, W.~Y., 2023. Logic-lm: Empowering large language models with symbolic solvers for faithful logical reasoning. arXiv preprint arXiv:2305.12295.

\bibitem[{Pang et~al.(2024{\natexlab{a}})Pang, Wang, Pun, Chen, and Xiong}]{pang2024illm}
Pang, A., Wang, M., Pun, M.-O., Chen, C.~S., Xiong, X., 2024{\natexlab{a}}. illm-tsc: Integration reinforcement learning and large language model for traffic signal control policy improvement. arXiv preprint arXiv:2407.06025.

\bibitem[{Pang et~al.(2024{\natexlab{b}})Pang, Wang, and Li}]{pang2024large}
Pang, H., Wang, Z., Li, G., 2024{\natexlab{b}}. Large language model guided deep reinforcement learning for decision making in autonomous driving. arXiv preprint arXiv:2412.18511.

\bibitem[{Papageorgiou(1998)}]{papageorgiou1998some}
Papageorgiou, M., 1998. Some remarks on macroscopic traffic flow modelling. Transportation Research Part A: Policy and Practice 32~(5), 323--329.

\bibitem[{Papageorgiou et~al.(2003)Papageorgiou, Diakaki, Dinopoulou, Kotsialos, and Wang}]{papageorgiou2003review}
Papageorgiou, M., Diakaki, C., Dinopoulou, V., Kotsialos, A., Wang, Y., 2003. Review of road traffic control strategies. Proceedings of the IEEE 91~(12), 2043--2067.

\bibitem[{Park et~al.(2024)Park, Lee, Kang, Choi, Park, Cho, Lee, and Kim}]{park2024vlaad}
Park, S., Lee, M., Kang, J., Choi, H., Park, Y., Cho, J., Lee, A., Kim, D., 2024. Vlaad: Vision and language assistant for autonomous driving. In: Proceedings of the IEEE/CVF Winter Conference on Applications of Computer Vision. pp. 980--987.

\bibitem[{Peng et~al.(2023)Peng, Alcaide, Anthony, Albalak, Arcadinho, Biderman, Cao, Cheng, Chung, Grella, et~al.}]{peng2023rwkv}
Peng, B., Alcaide, E., Anthony, Q., Albalak, A., Arcadinho, S., Biderman, S., Cao, H., Cheng, X., Chung, M., Grella, M., et~al., 2023. Rwkv: Reinventing rnns for the transformer era. arXiv preprint arXiv:2305.13048.

\bibitem[{Peng et~al.(2024)Peng, Guo, Chen, Zhu, Chen, Wang, Wang, et~al.}]{peng2024lc}
Peng, M., Guo, X., Chen, X., Zhu, M., Chen, K., Wang, X., Wang, Y., et~al., 2024. Lc-llm: Explainable lane-change intention and trajectory predictions with large language models. arXiv preprint arXiv:2403.18344.

\bibitem[{Peng et~al.(2025)Peng, Wang, Han, Zheng, and Ma}]{peng2025learningflow}
Peng, Z., Wang, Y., Han, X., Zheng, L., Ma, J., 2025. Learningflow: Automated policy learning workflow for urban driving with large language models. arXiv preprint arXiv:2501.05057.

\bibitem[{Pfeiffer et~al.(2020)Pfeiffer, R{\"u}ckl{\'e}, Poth, Kamath, Vuli{\'c}, Ruder, Cho, and Gurevych}]{pfeiffer2020adapterhub}
Pfeiffer, J., R{\"u}ckl{\'e}, A., Poth, C., Kamath, A., Vuli{\'c}, I., Ruder, S., Cho, K., Gurevych, I., 2020. Adapterhub: A framework for adapting transformers. arXiv preprint arXiv:2007.07779.

\bibitem[{Pi et~al.(2019)Pi, Ma, and Qian}]{pi2019general}
Pi, X., Ma, W., Qian, Z.~S., 2019. A general formulation for multi-modal dynamic traffic assignment considering multi-class vehicles, public transit and parking. Transportation Research Part C: Emerging Technologies 104, 369--389.

\bibitem[{Press et~al.(2021)Press, Smith, and Lewis}]{press2021train}
Press, O., Smith, N.~A., Lewis, M., 2021. Train short, test long: Attention with linear biases enables input length extrapolation. arXiv preprint arXiv:2108.12409.

\bibitem[{Qasemi et~al.(2023)Qasemi, Francis, and Oltramari}]{qasemi2023traffic}
Qasemi, E., Francis, J.~M., Oltramari, A., 2023. Traffic-domain video question answering with automatic captioning. arXiv preprint arXiv:2307.09636.

\bibitem[{Qian et~al.(2024)Qian, Chen, Zhuo, Jiao, and Jiang}]{qian2024nuscenes}
Qian, T., Chen, J., Zhuo, L., Jiao, Y., Jiang, Y.-G., 2024. Nuscenes-qa: A multi-modal visual question answering benchmark for autonomous driving scenario. In: Proceedings of the AAAI Conference on Artificial Intelligence. Vol.~38. pp. 4542--4550.

\bibitem[{Qin et~al.(2023)Qin, Luo, Zhao, Fang, Tao, and Wang}]{qin2023spatio}
Qin, Y., Luo, H., Zhao, F., Fang, Y., Tao, X., Wang, C., 2023. Spatio-temporal hierarchical mlp network for traffic forecasting. Information Sciences 632, 543--554.

\bibitem[{Qu et~al.(2024)Qu, Li, You, Zhu, Yan, Santi, Ratti, and Yuen}]{qu2024chatev}
Qu, H., Li, H., You, L., Zhu, R., Yan, J., Santi, P., Ratti, C., Yuen, C., 2024. Chatev: Predicting electric vehicle charging demand as natural language processing. Transportation Research Part D: Transport and Environment 136, 104470.

\bibitem[{Qu et~al.(2023)Qu, Lin, and Liu}]{qu2023envisioning}
Qu, X., Lin, H., Liu, Y., 2023. Envisioning the future of transportation: Inspiration of chatgpt and large models. Communications in Transportation Research 3, 100103.

\bibitem[{Raadsen et~al.(2020)Raadsen, Bliemer, and Bell}]{raadsen2020aggregation}
Raadsen, M.~P., Bliemer, M.~C., Bell, M.~G., 2020. Aggregation, disaggregation and decomposition methods in traffic assignment: historical perspectives and new trends. Transportation research part B: methodological 139, 199--223.

\bibitem[{Radford et~al.(2021)Radford, Kim, Hallacy, Ramesh, Goh, Agarwal, Sastry, Askell, Mishkin, Clark, et~al.}]{radford2021learning}
Radford, A., Kim, J.~W., Hallacy, C., Ramesh, A., Goh, G., Agarwal, S., Sastry, G., Askell, A., Mishkin, P., Clark, J., et~al., 2021. Learning transferable visual models from natural language supervision. In: International conference on machine learning. PmLR, pp. 8748--8763.

\bibitem[{Radford et~al.(2018)Radford, Narasimhan, Salimans, Sutskever, et~al.}]{radford2018improving}
Radford, A., Narasimhan, K., Salimans, T., Sutskever, I., et~al., 2018. Improving language understanding by generative pre-training.

\bibitem[{Radford et~al.(2019)Radford, Wu, Child, Luan, Amodei, Sutskever, et~al.}]{radford2019language}
Radford, A., Wu, J., Child, R., Luan, D., Amodei, D., Sutskever, I., et~al., 2019. Language models are unsupervised multitask learners. OpenAI blog 1~(8), 9.

\bibitem[{Rafailov et~al.(2023)Rafailov, Sharma, Mitchell, Manning, Ermon, and Finn}]{rafailov2023direct}
Rafailov, R., Sharma, A., Mitchell, E., Manning, C.~D., Ermon, S., Finn, C., 2023. Direct preference optimization: Your language model is secretly a reward model. Advances in Neural Information Processing Systems 36, 53728--53741.

\bibitem[{Raffel et~al.(2020)Raffel, Shazeer, Roberts, Lee, Narang, Matena, Zhou, Li, and Liu}]{raffel2020exploring}
Raffel, C., Shazeer, N., Roberts, A., Lee, K., Narang, S., Matena, M., Zhou, Y., Li, W., Liu, P.~J., 2020. Exploring the limits of transfer learning with a unified text-to-text transformer. Journal of machine learning research 21~(140), 1--67.

\bibitem[{Rahmani et~al.(2023)Rahmani, Baghbani, Bouguila, and Patterson}]{rahmani2023graph}
Rahmani, S., Baghbani, A., Bouguila, N., Patterson, Z., 2023. Graph neural networks for intelligent transportation systems: A survey. IEEE Transactions on Intelligent Transportation Systems 24~(8), 8846--8885.

\bibitem[{Ren et~al.(2024)Ren, Chen, Liu, Wang, Yu, and Cui}]{ren2024tpllm}
Ren, Y., Chen, Y., Liu, S., Wang, B., Yu, H., Cui, Z., 2024. Tpllm: A traffic prediction framework based on pretrained large language models. arXiv preprint arXiv:2403.02221.

\bibitem[{Renze(2024)}]{renze2024effect}
Renze, M., 2024. The effect of sampling temperature on problem solving in large language models. In: Findings of the Association for Computational Linguistics: EMNLP 2024. pp. 7346--7356.

\bibitem[{Ruan et~al.(2024)Ruan, Tsui, Li, and Shuai}]{ruan2024traffic}
Ruan, B.-K., Tsui, H.-T., Li, Y.-H., Shuai, H.-H., 2024. Traffic scene generation from natural language description for autonomous vehicles with large language model. arXiv preprint arXiv:2409.09575.

\bibitem[{Sanh et~al.(2021)Sanh, Webson, Raffel, Bach, Sutawika, Alyafeai, Chaffin, Stiegler, Scao, Raja, et~al.}]{sanh2021multitask}
Sanh, V., Webson, A., Raffel, C., Bach, S.~H., Sutawika, L., Alyafeai, Z., Chaffin, A., Stiegler, A., Scao, T.~L., Raja, A., et~al., 2021. Multitask prompted training enables zero-shot task generalization. arXiv preprint arXiv:2110.08207.

\bibitem[{Schick et~al.(2023)Schick, Dwivedi-Yu, Dess{\`\i}, Raileanu, Lomeli, Hambro, Zettlemoyer, Cancedda, and Scialom}]{schick2023toolformer}
Schick, T., Dwivedi-Yu, J., Dess{\`\i}, R., Raileanu, R., Lomeli, M., Hambro, E., Zettlemoyer, L., Cancedda, N., Scialom, T., 2023. Toolformer: Language models can teach themselves to use tools. Advances in Neural Information Processing Systems 36, 68539--68551.

\bibitem[{Schulman et~al.(2017)Schulman, Wolski, Dhariwal, Radford, and Klimov}]{schulman2017proximal}
Schulman, J., Wolski, F., Dhariwal, P., Radford, A., Klimov, O., 2017. Proximal policy optimization algorithms. arXiv preprint arXiv:1707.06347.

\bibitem[{Sha et~al.(2023)Sha, Mu, Jiang, Chen, Xu, Luo, Li, Tomizuka, Zhan, and Ding}]{sha2023languagempc}
Sha, H., Mu, Y., Jiang, Y., Chen, L., Xu, C., Luo, P., Li, S.~E., Tomizuka, M., Zhan, W., Ding, M., 2023. Languagempc: Large language models as decision makers for autonomous driving. arXiv preprint arXiv:2310.03026.

\bibitem[{Shanahan(2024)}]{shanahan2024talking}
Shanahan, M., 2024. Talking about large language models. Communications of the ACM 67~(2), 68--79.

\bibitem[{Shao et~al.(2024)Shao, Wang, Zhu, Xu, Song, Bi, Zhang, Zhang, Li, Wu, et~al.}]{shao2024deepseekmath}
Shao, Z., Wang, P., Zhu, Q., Xu, R., Song, J., Bi, X., Zhang, H., Zhang, M., Li, Y., Wu, Y., et~al., 2024. Deepseekmath: Pushing the limits of mathematical reasoning in open language models. arXiv preprint arXiv:2402.03300.

\bibitem[{Shao et~al.(2022)Shao, Zhang, Wang, Wei, and Xu}]{shao2022spatial}
Shao, Z., Zhang, Z., Wang, F., Wei, W., Xu, Y., 2022. Spatial-temporal identity: A simple yet effective baseline for multivariate time series forecasting. In: Proceedings of the 31st ACM International Conference on Information \& Knowledge Management. pp. 4454--4458.

\bibitem[{Shaygan et~al.(2022)Shaygan, Meese, Li, Zhao, and Nejad}]{shaygan2022traffic}
Shaygan, M., Meese, C., Li, W., Zhao, X.~G., Nejad, M., 2022. Traffic prediction using artificial intelligence: Review of recent advances and emerging opportunities. Transportation research part C: emerging technologies 145, 103921.

\bibitem[{Shi et~al.(2024)Shi, Zhang, Yao, Li, and Zhao}]{shi2024keep}
Shi, L., Zhang, H., Yao, Y., Li, Z., Zhao, H., 2024. Keep the cost down: A review on methods to optimize llm's kv-cache consumption. arXiv preprint arXiv:2407.18003.

\bibitem[{Shi et~al.(2021)Shi, Mo, and Di}]{shi2021physics}
Shi, R., Mo, Z., Di, X., 2021. Physics-informed deep learning for traffic state estimation: A hybrid paradigm informed by second-order traffic models. In: Proceedings of the AAAI Conference on Artificial Intelligence. Vol.~35. pp. 540--547.

\bibitem[{Shin et~al.(2020)Shin, Razeghi, Logan~IV, Wallace, and Singh}]{shin2020autoprompt}
Shin, T., Razeghi, Y., Logan~IV, R.~L., Wallace, E., Singh, S., 2020. Autoprompt: Eliciting knowledge from language models with automatically generated prompts. arXiv preprint arXiv:2010.15980.

\bibitem[{Shinn et~al.(2023)Shinn, Labash, and Gopinath}]{shinn2023reflexion}
Shinn, N., Labash, B., Gopinath, A., 2023. Reflexion: an autonomous agent with dynamic memory and self-reflection. arXiv preprint arXiv:2303.11366 2~(5), 9.

\bibitem[{Shoeybi et~al.(2019)Shoeybi, Patwary, Puri, LeGresley, Casper, and Catanzaro}]{shoeybi2019megatron}
Shoeybi, M., Patwary, M., Puri, R., LeGresley, P., Casper, J., Catanzaro, B., 2019. Megatron-lm: Training multi-billion parameter language models using model parallelism. arXiv preprint arXiv:1909.08053.

\bibitem[{Sima et~al.(2024)Sima, Renz, Chitta, Chen, Zhang, Xie, Bei{\ss}wenger, Luo, Geiger, and Li}]{sima2024drivelm}
Sima, C., Renz, K., Chitta, K., Chen, L., Zhang, H., Xie, C., Bei{\ss}wenger, J., Luo, P., Geiger, A., Li, H., 2024. Drivelm: Driving with graph visual question answering. In: European Conference on Computer Vision. Springer, pp. 256--274.

\bibitem[{Song et~al.(2023)Song, Wu, Washington, Sadler, Chao, and Su}]{song2023llm}
Song, C.~H., Wu, J., Washington, C., Sadler, B.~M., Chao, W.-L., Su, Y., 2023. Llm-planner: Few-shot grounded planning for embodied agents with large language models. In: Proceedings of the IEEE/CVF international conference on computer vision. pp. 2998--3009.

\bibitem[{Su et~al.(2024)Su, Ahmed, Lu, Pan, Bo, and Liu}]{su2024roformer}
Su, J., Ahmed, M., Lu, Y., Pan, S., Bo, W., Liu, Y., 2024. Roformer: Enhanced transformer with rotary position embedding. Neurocomputing 568, 127063.

\bibitem[{Syed et~al.(2024)Syed, Light, Guo, Zhang, Qin, Ouyang, and Hu}]{syed2024benchmarking}
Syed, U., Light, E., Guo, X., Zhang, H., Qin, L., Ouyang, Y., Hu, B., 2024. Benchmarking the capabilities of large language models in transportation system engineering: Accuracy, consistency, and reasoning behaviors. arXiv preprint arXiv:2408.08302.

\bibitem[{Tan et~al.(2023)Tan, Ivanovic, Weng, Pavone, and Kraehenbuehl}]{tan2023language}
Tan, S., Ivanovic, B., Weng, X., Pavone, M., Kraehenbuehl, P., 2023. Language conditioned traffic generation. arXiv preprint arXiv:2307.07947.

\bibitem[{Tang et~al.(2024{\natexlab{a}})Tang, Dai, and Lv}]{tang2024large}
Tang, Y., Dai, X., Lv, Y., 2024{\natexlab{a}}. Large language model-assisted arterial traffic signal control. IEEE Journal of Radio Frequency Identification 8, 322--326.

\bibitem[{Tang et~al.(2022)Tang, Qu, Chow, Lam, Wong, and Ma}]{tang2022domain}
Tang, Y., Qu, A., Chow, A.~H., Lam, W.~H., Wong, S.~C., Ma, W., 2022. Domain adversarial spatial-temporal network: A transferable framework for short-term traffic forecasting across cities. In: Proceedings of the 31st ACM international conference on information \& knowledge management. pp. 1905--1915.

\bibitem[{Tang et~al.(2024{\natexlab{b}})Tang, Wang, Qu, Yan, Wu, Zhuang, Kai, Hou, Guo, Zhao, et~al.}]{tang2024itinera}
Tang, Y., Wang, Z., Qu, A., Yan, Y., Wu, Z., Zhuang, D., Kai, J., Hou, K., Guo, X., Zhao, J., et~al., 2024{\natexlab{b}}. Itinera: Integrating spatial optimization with large language models for open-domain urban itinerary planning. In: Proceedings of the 2024 Conference on Empirical Methods in Natural Language Processing: Industry Track. pp. 1413--1432.

\bibitem[{Tay et~al.(2022)Tay, Dehghani, Tran, Garcia, Wei, Wang, Chung, Shakeri, Bahri, Schuster, et~al.}]{tay2022ul2}
Tay, Y., Dehghani, M., Tran, V.~Q., Garcia, X., Wei, J., Wang, X., Chung, H.~W., Shakeri, S., Bahri, D., Schuster, T., et~al., 2022. Ul2: Unifying language learning paradigms. arXiv preprint arXiv:2205.05131.

\bibitem[{Team et~al.(2024)Team, Riviere, Pathak, Sessa, Hardin, Bhupatiraju, Hussenot, Mesnard, Shahriari, Ram{\'e}, et~al.}]{team2024gemma}
Team, G., Riviere, M., Pathak, S., Sessa, P.~G., Hardin, C., Bhupatiraju, S., Hussenot, L., Mesnard, T., Shahriari, B., Ram{\'e}, A., et~al., 2024. Gemma 2: Improving open language models at a practical size. arXiv preprint arXiv:2408.00118.

\bibitem[{Tedjopurnomo et~al.(2020)Tedjopurnomo, Bao, Zheng, Choudhury, and Qin}]{tedjopurnomo2020survey}
Tedjopurnomo, D.~A., Bao, Z., Zheng, B., Choudhury, F.~M., Qin, A.~K., 2020. A survey on modern deep neural network for traffic prediction: Trends, methods and challenges. IEEE Transactions on Knowledge and Data Engineering 34~(4), 1544--1561.

\bibitem[{Tian et~al.(2024)Tian, Reddy, Feng, Quddus, Demiris, and Angeloudis}]{tian2024enhancing}
Tian, H., Reddy, K., Feng, Y., Quddus, M., Demiris, Y., Angeloudis, P., 2024. Enhancing autonomous vehicle training with language model integration and critical scenario generation. arXiv preprint arXiv:2404.08570.

\bibitem[{Toma{\v{s}}ev et~al.(2020)Toma{\v{s}}ev, Cornebise, Hutter, Mohamed, Picciariello, Connelly, Belgrave, Ezer, Haert, Mugisha, et~al.}]{tomavsev2020ai}
Toma{\v{s}}ev, N., Cornebise, J., Hutter, F., Mohamed, S., Picciariello, A., Connelly, B., Belgrave, D.~C., Ezer, D., Haert, F. C. v.~d., Mugisha, F., et~al., 2020. Ai for social good: unlocking the opportunity for positive impact. Nature Communications 11~(1), 2468.

\bibitem[{Tong and Solmaz(2024)}]{tong2024connectgpt}
Tong, K., Solmaz, S., 2024. Connectgpt: Connect large language models with connected and automated vehicles. In: 2024 IEEE Intelligent Vehicles Symposium (IV). IEEE, pp. 581--588.

\bibitem[{Touvron et~al.(2023{\natexlab{a}})Touvron, Lavril, Izacard, Martinet, Lachaux, Lacroix, Rozi{\`e}re, Goyal, Hambro, Azhar, et~al.}]{touvron2023llama}
Touvron, H., Lavril, T., Izacard, G., Martinet, X., Lachaux, M.-A., Lacroix, T., Rozi{\`e}re, B., Goyal, N., Hambro, E., Azhar, F., et~al., 2023{\natexlab{a}}. Llama: Open and efficient foundation language models. arXiv preprint arXiv:2302.13971.

\bibitem[{Touvron et~al.(2023{\natexlab{b}})Touvron, Martin, Stone, Albert, Almahairi, Babaei, Bashlykov, Batra, Bhargava, Bhosale, et~al.}]{touvron2023llama2}
Touvron, H., Martin, L., Stone, K., Albert, P., Almahairi, A., Babaei, Y., Bashlykov, N., Batra, S., Bhargava, P., Bhosale, S., et~al., 2023{\natexlab{b}}. Llama 2: Open foundation and fine-tuned chat models. arXiv preprint arXiv:2307.09288.

\bibitem[{Vahidi and Sciarretta(2018)}]{vahidi2018energy}
Vahidi, A., Sciarretta, A., 2018. Energy saving potentials of connected and automated vehicles. Transportation Research Part C: Emerging Technologies 95, 822--843.

\bibitem[{Van~Brummelen et~al.(2018)Van~Brummelen, O’brien, Gruyer, and Najjaran}]{van2018autonomous}
Van~Brummelen, J., O’brien, M., Gruyer, D., Najjaran, H., 2018. Autonomous vehicle perception: The technology of today and tomorrow. Transportation research part C: emerging technologies 89, 384--406.

\bibitem[{Vaswani et~al.(2017)Vaswani, Shazeer, Parmar, Uszkoreit, Jones, Gomez, Kaiser, and Polosukhin}]{vaswani2017attention}
Vaswani, A., Shazeer, N., Parmar, N., Uszkoreit, J., Jones, L., Gomez, A.~N., Kaiser, {\L}., Polosukhin, I., 2017. Attention is all you need. Advances in neural information processing systems 30, 5998--6008.

\bibitem[{Veres and Moussa(2019)}]{veres2019deep}
Veres, M., Moussa, M., 2019. Deep learning for intelligent transportation systems: A survey of emerging trends. IEEE Transactions on Intelligent transportation systems 21~(8), 3152--3168.

\bibitem[{Villarreal et~al.(2023)Villarreal, Poudel, and Li}]{villarreal2023can}
Villarreal, M., Poudel, B., Li, W., 2023. Can chatgpt enable its? the case of mixed traffic control via reinforcement learning. In: 2023 IEEE 26th International Conference on Intelligent Transportation Systems (ITSC). IEEE, pp. 3749--3755.

\bibitem[{Wandelt et~al.(2024)Wandelt, Zheng, Wang, Liu, and Sun}]{wandelt2024large}
Wandelt, S., Zheng, C., Wang, S., Liu, Y., Sun, X., 2024. Large language models for intelligent transportation: A review of the state of the art and challenges. Applied Sciences 14~(17), 7455.

\bibitem[{Wang et~al.(2024{\natexlab{a}})Wang, Cai, Karim, Liu, and Wang}]{wang2024traffic}
Wang, B., Cai, Z., Karim, M.~M., Liu, C., Wang, Y., 2024{\natexlab{a}}. Traffic performance gpt (tp-gpt): Real-time data informed intelligent chatbot for transportation surveillance and management. arXiv preprint arXiv:2405.03076.

\bibitem[{Wang et~al.(2024{\natexlab{b}})Wang, Ma, Dong, Huang, Zhang, and Wei}]{wang2024deepnet}
Wang, H., Ma, S., Dong, L., Huang, S., Zhang, D., Wei, F., 2024{\natexlab{b}}. Deepnet: Scaling transformers to 1,000 layers. IEEE Transactions on Pattern Analysis and Machine Intelligence 46~(10), 6761--6774.

\bibitem[{Wang et~al.(2024{\natexlab{c}})Wang, Ma, Feng, Zhang, Yang, Zhang, Chen, Tang, Chen, Lin, et~al.}]{wang2024survey}
Wang, L., Ma, C., Feng, X., Zhang, Z., Yang, H., Zhang, J., Chen, Z., Tang, J., Chen, X., Lin, Y., et~al., 2024{\natexlab{c}}. A survey on large language model based autonomous agents. Frontiers of Computer Science 18~(6), 186345.

\bibitem[{Wang et~al.(2023{\natexlab{a}})Wang, Ren, Jiang, Cai, Fu, Wang, Cui, Yu, Wang, Zhou, et~al.}]{wang2023accidentgpt}
Wang, L., Ren, Y., Jiang, H., Cai, P., Fu, D., Wang, T., Cui, Z., Yu, H., Wang, X., Zhou, H., et~al., 2023{\natexlab{a}}. Accidentgpt: Accident analysis and prevention from v2x environmental perception with multi-modal large model. arXiv preprint arXiv:2312.13156.

\bibitem[{Wang et~al.(2024{\natexlab{d}})Wang, Pang, Kan, Pun, Chen, and Huang}]{wang2024llm}
Wang, M., Pang, A., Kan, Y., Pun, M.-O., Chen, C.~S., Huang, B., 2024{\natexlab{d}}. Llm-assisted light: Leveraging large language model capabilities for human-mimetic traffic signal control in complex urban environments. arXiv preprint arXiv:2403.08337.

\bibitem[{Wang et~al.(2024{\natexlab{e}})Wang, Wei, Hu, and Han}]{wang2024transgpt}
Wang, P., Wei, X., Hu, F., Han, W., 2024{\natexlab{e}}. Transgpt: Multi-modal generative pre-trained transformer for transportation. arXiv preprint arXiv:2402.07233.

\bibitem[{Wang et~al.(2023{\natexlab{b}})Wang, Zhu, Li, Wang, Li, and He}]{wang2023chatgpt}
Wang, S., Zhu, Y., Li, Z., Wang, Y., Li, L., He, Z., 2023{\natexlab{b}}. Chatgpt as your vehicle co-pilot: An initial attempt. IEEE Transactions on Intelligent Vehicles 8~(12), 4706--4721.

\bibitem[{Wang et~al.(2024{\natexlab{f}})Wang, Maalouf, Xiao, Ban, Amini, Rosman, Karaman, and Rus}]{wang2024drive}
Wang, T.-H., Maalouf, A., Xiao, W., Ban, Y., Amini, A., Rosman, G., Karaman, S., Rus, D., 2024{\natexlab{f}}. Drive anywhere: Generalizable end-to-end autonomous driving with multi-modal foundation models. In: 2024 IEEE International Conference on Robotics and Automation (ICRA). IEEE, pp. 6687--6694.

\bibitem[{Wang et~al.(2023{\natexlab{c}})Wang, Xie, Hu, Zou, Fan, Tong, Wen, Wu, Deng, Li, et~al.}]{wang2023drivemlm}
Wang, W., Xie, J., Hu, C., Zou, H., Fan, J., Tong, W., Wen, Y., Wu, S., Deng, H., Li, Z., et~al., 2023{\natexlab{c}}. Drivemlm: Aligning multi-modal large language models with behavioral planning states for autonomous driving. arXiv preprint arXiv:2312.09245.

\bibitem[{Wang et~al.(2023{\natexlab{d}})Wang, Fang, Zeng, and Cheng}]{wang2023would}
Wang, X., Fang, M., Zeng, Z., Cheng, T., 2023{\natexlab{d}}. Where would i go next? large language models as human mobility predictors. arXiv preprint arXiv:2308.15197.

\bibitem[{Wang et~al.(2023{\natexlab{e}})Wang, Wang, Chen, Wang, and Lin}]{wang2023building}
Wang, X., Wang, D., Chen, L., Wang, F.-Y., Lin, Y., 2023{\natexlab{e}}. Building transportation foundation model via generative graph transformer. In: 2023 IEEE 26th International Conference on Intelligent Transportation Systems (ITSC). IEEE, pp. 6042--6047.

\bibitem[{Wang et~al.(2022{\natexlab{a}})Wang, Wei, Schuurmans, Le, Chi, Narang, Chowdhery, and Zhou}]{wang2022self}
Wang, X., Wei, J., Schuurmans, D., Le, Q., Chi, E., Narang, S., Chowdhery, A., Zhou, D., 2022{\natexlab{a}}. Self-consistency improves chain of thought reasoning in language models. arXiv preprint arXiv:2203.11171.

\bibitem[{Wang et~al.(2024{\natexlab{g}})Wang, Zhu, Huang, Chen, Zhu, and Lu}]{wang2024drivedreamer}
Wang, X., Zhu, Z., Huang, G., Chen, X., Zhu, J., Lu, J., 2024{\natexlab{g}}. Drivedreamer: Towards real-world-drive world models for autonomous driving. In: European Conference on Computer Vision. Springer, pp. 55--72.

\bibitem[{Wang et~al.(2024{\natexlab{h}})Wang, Cheng, He, Wang, Dai, Chen, Xia, and Zhang}]{wang2024drivingdojo}
Wang, Y., Cheng, K., He, J., Wang, Q., Dai, H., Chen, Y., Xia, F., Zhang, Z., 2024{\natexlab{h}}. Drivingdojo dataset: Advancing interactive and knowledge-enriched driving world model. arXiv preprint arXiv:2410.10738.

\bibitem[{Wang et~al.(2024{\natexlab{i}})Wang, He, Fan, Li, Chen, and Zhang}]{wang2024driving}
Wang, Y., He, J., Fan, L., Li, H., Chen, Y., Zhang, Z., 2024{\natexlab{i}}. Driving into the future: Multiview visual forecasting and planning with world model for autonomous driving. In: Proceedings of the IEEE/CVF Conference on Computer Vision and Pattern Recognition. pp. 14749--14759.

\bibitem[{Wang et~al.(2022{\natexlab{b}})Wang, Mishra, Alipoormolabashi, Kordi, Mirzaei, Arunkumar, Ashok, Dhanasekaran, Naik, Stap, et~al.}]{wang2022super}
Wang, Y., Mishra, S., Alipoormolabashi, P., Kordi, Y., Mirzaei, A., Arunkumar, A., Ashok, A., Dhanasekaran, A.~S., Naik, A., Stap, D., et~al., 2022{\natexlab{b}}. Super-naturalinstructions: Generalization via declarative instructions on 1600+ nlp tasks. arXiv preprint arXiv:2204.07705.

\bibitem[{Wang et~al.(2018)Wang, Szeto, Han, and Friesz}]{wang2018dynamic}
Wang, Y., Szeto, W.~Y., Han, K., Friesz, T.~L., 2018. Dynamic traffic assignment: A review of the methodological advances for environmentally sustainable road transportation applications. Transportation Research Part B: Methodological 111, 370--394.

\bibitem[{Wang et~al.(2019)Wang, Zhang, Liu, Dai, and Lee}]{wang2019enhancing}
Wang, Y., Zhang, D., Liu, Y., Dai, B., Lee, L.~H., 2019. Enhancing transportation systems via deep learning: A survey. Transportation research part C: emerging technologies 99, 144--163.

\bibitem[{Wang et~al.(2023{\natexlab{f}})Wang, Zhong, Li, Mi, Zeng, Huang, Shang, Jiang, and Liu}]{wang2023aligning}
Wang, Y., Zhong, W., Li, L., Mi, F., Zeng, X., Huang, W., Shang, L., Jiang, X., Liu, Q., 2023{\natexlab{f}}. Aligning large language models with human: A survey. arXiv preprint arXiv:2307.12966.

\bibitem[{Washington et~al.(2020)Washington, Karlaftis, Mannering, and Anastasopoulos}]{washington2020statistical}
Washington, S., Karlaftis, M.~G., Mannering, F., Anastasopoulos, P., 2020. Statistical and econometric methods for transportation data analysis. Chapman and Hall/CRC.

\bibitem[{Wei et~al.(2021)Wei, Bosma, Zhao, Guu, Yu, Lester, Du, Dai, and Le}]{wei2021finetuned}
Wei, J., Bosma, M., Zhao, V.~Y., Guu, K., Yu, A.~W., Lester, B., Du, N., Dai, A.~M., Le, Q.~V., 2021. Finetuned language models are zero-shot learners. arXiv preprint arXiv:2109.01652.

\bibitem[{Wei et~al.(2022{\natexlab{a}})Wei, Tay, Bommasani, Raffel, Zoph, Borgeaud, Yogatama, Bosma, Zhou, Metzler, et~al.}]{wei2022emergent}
Wei, J., Tay, Y., Bommasani, R., Raffel, C., Zoph, B., Borgeaud, S., Yogatama, D., Bosma, M., Zhou, D., Metzler, D., et~al., 2022{\natexlab{a}}. Emergent abilities of large language models. arXiv preprint arXiv:2206.07682.

\bibitem[{Wei et~al.(2022{\natexlab{b}})Wei, Wang, Schuurmans, Bosma, Xia, Chi, Le, Zhou, et~al.}]{wei2022chain}
Wei, J., Wang, X., Schuurmans, D., Bosma, M., Xia, F., Chi, E., Le, Q.~V., Zhou, D., et~al., 2022{\natexlab{b}}. Chain-of-thought prompting elicits reasoning in large language models. Advances in neural information processing systems 35, 24824--24837.

\bibitem[{Wen et~al.(2023{\natexlab{a}})Wen, Fu, Li, Cai, Ma, Cai, Dou, Shi, He, and Qiao}]{wen2023dilu}
Wen, L., Fu, D., Li, X., Cai, X., Ma, T., Cai, P., Dou, M., Shi, B., He, L., Qiao, Y., 2023{\natexlab{a}}. Dilu: A knowledge-driven approach to autonomous driving with large language models. arXiv preprint arXiv:2309.16292.

\bibitem[{Wen et~al.(2023{\natexlab{b}})Wen, Yang, Fu, Wang, Cai, Li, Ma, Li, Xu, Shang, et~al.}]{wen2023road}
Wen, L., Yang, X., Fu, D., Wang, X., Cai, P., Li, X., Ma, T., Li, Y., Xu, L., Shang, D., et~al., 2023{\natexlab{b}}. On the road with gpt-4v (ision): Early explorations of visual-language model on autonomous driving. arXiv preprint arXiv:2311.05332.

\bibitem[{White et~al.(2023)White, Fu, Hays, Sandborn, Olea, Gilbert, Elnashar, Spencer-Smith, and Schmidt}]{white2023prompt}
White, J., Fu, Q., Hays, S., Sandborn, M., Olea, C., Gilbert, H., Elnashar, A., Spencer-Smith, J., Schmidt, D.~C., 2023. A prompt pattern catalog to enhance prompt engineering with chatgpt. arXiv preprint arXiv:2302.11382.

\bibitem[{Wu et~al.(2023)Wu, Han, Wang, Liu, Zhang, and Shen}]{wu2023language}
Wu, D., Han, W., Wang, T., Liu, Y., Zhang, X., Shen, J., 2023. Language prompt for autonomous driving. arXiv preprint arXiv:2309.04379.

\bibitem[{Wu et~al.(2018)Wu, Guo, Xian, and Zhou}]{wu2018hierarchical}
Wu, X., Guo, J., Xian, K., Zhou, X., 2018. Hierarchical travel demand estimation using multiple data sources: A forward and backward propagation algorithmic framework on a layered computational graph. Transportation Research Part C: Emerging Technologies 96, 321--346.

\bibitem[{Xia et~al.(2024)Xia, Xu, Xu, Xie, Wang, and Chen}]{xia2024language}
Xia, J., Xu, C., Xu, Q., Xie, C., Wang, Y., Chen, S., 2024. Language-driven interactive traffic trajectory generation. arXiv preprint arXiv:2405.15388.

\bibitem[{Xie et~al.(2025)Xie, Kong, Dong, Sima, Zhang, Chen, Liu, and Pan}]{xie2025vlms}
Xie, S., Kong, L., Dong, Y., Sima, C., Zhang, W., Chen, Q.~A., Liu, Z., Pan, L., 2025. Are vlms ready for autonomous driving? an empirical study from the reliability, data, and metric perspectives. arXiv preprint arXiv:2501.04003.

\bibitem[{Xiong et~al.(2020)Xiong, Yang, He, Zheng, Zheng, Xing, Zhang, Lan, Wang, and Liu}]{xiong2020layer}
Xiong, R., Yang, Y., He, D., Zheng, K., Zheng, S., Xing, C., Zhang, H., Lan, Y., Wang, L., Liu, T., 2020. On layer normalization in the transformer architecture. In: International conference on machine learning. PMLR, pp. 10524--10533.

\bibitem[{Xu et~al.(2024{\natexlab{a}})Xu, Yuan, Zhou, Xu, Li, Ban, and Ye}]{xu2024genai}
Xu, H., Yuan, J., Zhou, A., Xu, G., Li, W., Ban, X., Ye, X., 2024{\natexlab{a}}. Genai-powered multi-agent paradigm for smart urban mobility: Opportunities and challenges for integrating large language models (llms) and retrieval-augmented generation (rag) with intelligent transportation systems. arXiv preprint arXiv:2409.00494.

\bibitem[{Xu et~al.(2021)Xu, Huang, and Liu}]{xu2021sutd}
Xu, L., Huang, H., Liu, J., 2021. Sutd-trafficqa: A question answering benchmark and an efficient network for video reasoning over traffic events. In: Proceedings of the IEEE/CVF conference on computer vision and pattern recognition. pp. 9878--9888.

\bibitem[{Xu et~al.(2020)Xu, Dai, Liu, Gao, Lin, Qi, and Xiong}]{xu2020spatial}
Xu, M., Dai, W., Liu, C., Gao, X., Lin, W., Qi, G.-J., Xiong, H., 2020. Spatial-temporal transformer networks for traffic flow forecasting. arXiv preprint arXiv:2001.02908.

\bibitem[{Xu et~al.(2024{\natexlab{b}})Xu, Zhang, Xie, Zhao, Guo, Wong, Li, and Zhao}]{xu2024drivegpt4}
Xu, Z., Zhang, Y., Xie, E., Zhao, Z., Guo, Y., Wong, K.-Y.~K., Li, Z., Zhao, H., 2024{\natexlab{b}}. Drivegpt4: Interpretable end-to-end autonomous driving via large language model. IEEE Robotics and Automation Letters 9~(10), 8186--8193.

\bibitem[{Xue et~al.(2024)Xue, Tang, Payani, and Salim}]{xue2024prompt}
Xue, H., Tang, T., Payani, A., Salim, F.~D., 2024. Prompt mining for language-based human mobility forecasting. arXiv preprint arXiv:2403.03544.

\bibitem[{Xue et~al.(2022)Xue, Voutharoja, and Salim}]{xue2022leveraging}
Xue, H., Voutharoja, B.~P., Salim, F.~D., 2022. Leveraging language foundation models for human mobility forecasting. In: Proceedings of the 30th International Conference on Advances in Geographic Information Systems. pp. 1--9.

\bibitem[{Xue et~al.(2025)Xue, Tan, Ma, and Ukkusuri}]{xue2025data}
Xue, J., Tan, R., Ma, J., Ukkusuri, S.~V., 2025. Data mining in transportation networks with graph neural networks: A review and outlook. arXiv preprint arXiv:2501.16656.

\bibitem[{Yan and Li(2023)}]{yan2023survey}
Yan, H., Li, Y., 2023. A survey of generative ai for intelligent transportation systems. arXiv preprint arXiv:2312.08248.

\bibitem[{Yan et~al.(2021)Yan, Ma, and Pu}]{yan2021learning}
Yan, H., Ma, X., Pu, Z., 2021. Learning dynamic and hierarchical traffic spatiotemporal features with transformer. IEEE Transactions on Intelligent Transportation Systems 23~(11), 22386--22399.

\bibitem[{Yang et~al.(2024{\natexlab{a}})Yang, Wu, and Xu}]{yang2024transcompressor}
Yang, H., Wu, R., Xu, W., 2024{\natexlab{a}}. Transcompressor: Llm-powered multimodal data compression for smart transportation. In: Proceedings of the 30th Annual International Conference on Mobile Computing and Networking. pp. 2335--2340.

\bibitem[{Yang et~al.(2024{\natexlab{b}})Yang, Gao, Qiu, Chen, Li, Dai, Chitta, Wu, Zeng, Luo, et~al.}]{yang2024generalized}
Yang, J., Gao, S., Qiu, Y., Chen, L., Li, T., Dai, B., Chitta, K., Wu, P., Zeng, J., Luo, P., et~al., 2024{\natexlab{b}}. Generalized predictive model for autonomous driving. In: Proceedings of the IEEE/CVF Conference on Computer Vision and Pattern Recognition. pp. 14662--14672.

\bibitem[{Yang et~al.(2024{\natexlab{c}})Yang, Zhang, Fernandez-Laaksonen, Ding, and Gong}]{yang2024driving}
Yang, R., Zhang, X., Fernandez-Laaksonen, A., Ding, X., Gong, J., 2024{\natexlab{c}}. Driving style alignment for llm-powered driver agent. arXiv preprint arXiv:2403.11368.

\bibitem[{Yao et~al.(2024{\natexlab{a}})Yao, Da, Nandam, Turnau, Liu, Pang, and Wei}]{yao2024comal}
Yao, H., Da, L., Nandam, V., Turnau, J., Liu, Z., Pang, L., Wei, H., 2024{\natexlab{a}}. Comal: Collaborative multi-agent large language models for mixed-autonomy traffic. arXiv preprint arXiv:2410.14368.

\bibitem[{Yao et~al.(2023)Yao, Yu, Zhao, Shafran, Griffiths, Cao, and Narasimhan}]{yao2023tree}
Yao, S., Yu, D., Zhao, J., Shafran, I., Griffiths, T., Cao, Y., Narasimhan, K., 2023. Tree of thoughts: Deliberate problem solving with large language models. Advances in neural information processing systems 36, 11809--11822.

\bibitem[{Yao et~al.(2024{\natexlab{b}})Yao, Duan, Xu, Cai, Sun, and Zhang}]{yao2024survey}
Yao, Y., Duan, J., Xu, K., Cai, Y., Sun, Z., Zhang, Y., 2024{\natexlab{b}}. A survey on large language model (llm) security and privacy: The good, the bad, and the ugly. High-Confidence Computing, 100211.

\bibitem[{Ye et~al.(2019)Ye, Wu, Ruan, Li, Chen, Gao, and Chen}]{ye2019survey}
Ye, B.-L., Wu, W., Ruan, K., Li, L., Chen, T., Gao, H., Chen, Y., 2019. A survey of model predictive control methods for traffic signal control. IEEE/CAA Journal of Automatica Sinica 6~(3), 623--640.

\bibitem[{Yin et~al.(2021)Yin, Wu, Wei, Shen, Qi, and Yin}]{yin2021deep}
Yin, X., Wu, G., Wei, J., Shen, Y., Qi, H., Yin, B., 2021. Deep learning on traffic prediction: Methods, analysis, and future directions. IEEE Transactions on Intelligent Transportation Systems 23~(6), 4927--4943.

\bibitem[{You et~al.(2024)You, Shi, Jiang, Huang, Gan, Wu, Cheng, Li, and Ran}]{you2024v2x}
You, J., Shi, H., Jiang, Z., Huang, Z., Gan, R., Wu, K., Cheng, X., Li, X., Ran, B., 2024. V2x-vlm: End-to-end v2x cooperative autonomous driving through large vision-language models. arXiv preprint arXiv:2408.09251.

\bibitem[{Yu et~al.(2024)Yu, Wang, and Ma}]{yu2024large}
Yu, J., Wang, Y., Ma, W., 2024. Large language model-enhanced reinforcement learning for generic bus holding control strategies. arXiv preprint arXiv:2410.10212.

\bibitem[{Yu et~al.(2023)Yu, Zhuang, Zhang, Meng, Ratner, Krishna, Shen, and Zhang}]{yu2023large}
Yu, Y., Zhuang, Y., Zhang, J., Meng, Y., Ratner, A.~J., Krishna, R., Shen, J., Zhang, C., 2023. Large language model as attributed training data generator: A tale of diversity and bias. Advances in Neural Information Processing Systems 36, 55734--55784.

\bibitem[{Yuan et~al.(2024)Yuan, Sun, Omeiza, Zhao, Newman, Kunze, and Gadd}]{yuan2024rag}
Yuan, J., Sun, S., Omeiza, D., Zhao, B., Newman, P., Kunze, L., Gadd, M., 2024. Rag-driver: Generalisable driving explanations with retrieval-augmented in-context learning in multi-modal large language model. arXiv preprint arXiv:2402.10828.

\bibitem[{Zeng et~al.(2022)Zeng, Liu, Du, Wang, Lai, Ding, Yang, Xu, Zheng, Xia, et~al.}]{zeng2022glm}
Zeng, A., Liu, X., Du, Z., Wang, Z., Lai, H., Ding, M., Yang, Z., Xu, Y., Zheng, W., Xia, X., et~al., 2022. Glm-130b: An open bilingual pre-trained model. arXiv preprint arXiv:2210.02414.

\bibitem[{Zhang and Sennrich(2019)}]{zhang2019root}
Zhang, B., Sennrich, R., 2019. Root mean square layer normalization. Advances in Neural Information Processing Systems 32, 12381--12392.

\bibitem[{Zhang et~al.(2024{\natexlab{a}})Zhang, Zheng, Yue, and Wang}]{zhang2024advancing}
Zhang, D., Zheng, H., Yue, W., Wang, X., 2024{\natexlab{a}}. Advancing its applications with llms: A survey on traffic management, transportation safety, and autonomous driving. In: International Joint Conference on Rough Sets. Springer, pp. 295--309.

\bibitem[{Zhang et~al.(2011)Zhang, Wang, Wang, Lin, Xu, and Chen}]{zhang2011data}
Zhang, J., Wang, F.-Y., Wang, K., Lin, W.-H., Xu, X., Chen, C., 2011. Data-driven intelligent transportation systems: A survey. IEEE Transactions on Intelligent Transportation Systems 12~(4), 1624--1639.

\bibitem[{Zhang et~al.(2024{\natexlab{b}})Zhang, Xu, and Li}]{zhang2024chatscene}
Zhang, J., Xu, C., Li, B., 2024{\natexlab{b}}. Chatscene: Knowledge-enabled safety-critical scenario generation for autonomous vehicles. In: Proceedings of the IEEE/CVF Conference on Computer Vision and Pattern Recognition. pp. 15459--15469.

\bibitem[{Zhang et~al.(2024{\natexlab{c}})Zhang, Zhou, Wu, Xie, and He}]{zhang2024semantic}
Zhang, K., Zhou, F., Wu, L., Xie, N., He, Z., 2024{\natexlab{c}}. Semantic understanding and prompt engineering for large-scale traffic data imputation. Information Fusion 102, 102038.

\bibitem[{Zhang et~al.(2025{\natexlab{a}})Zhang, Wang, Zhang, Bian, Feng, and Ozbay}]{zhang2025language}
Zhang, R., Wang, B., Zhang, J., Bian, Z., Feng, C., Ozbay, K., 2025{\natexlab{a}}. When language and vision meet road safety: leveraging multimodal large language models for video-based traffic accident analysis. arXiv preprint arXiv:2501.10604.

\bibitem[{Zhang et~al.(2023)Zhang, Dong, Li, Zhang, Sun, Wang, Li, Hu, Zhang, Wu, et~al.}]{zhang2023instruction}
Zhang, S., Dong, L., Li, X., Zhang, S., Sun, X., Wang, S., Li, J., Hu, R., Zhang, T., Wu, F., et~al., 2023. Instruction tuning for large language models: A survey. arXiv preprint arXiv:2308.10792.

\bibitem[{Zhang et~al.(2024{\natexlab{d}})Zhang, Fu, Liang, Zhang, Yu, Cai, and Yao}]{zhang2024trafficgpt}
Zhang, S., Fu, D., Liang, W., Zhang, Z., Yu, B., Cai, P., Yao, B., 2024{\natexlab{d}}. Trafficgpt: Viewing, processing and interacting with traffic foundation models. Transport Policy 150, 95--105.

\bibitem[{Zhang et~al.(2024{\natexlab{e}})Zhang, Shi, Lou, Qi, Chen, Xu, and Han}]{zhang2024transportationgames}
Zhang, X., Shi, X., Lou, X., Qi, R., Chen, Y., Xu, J., Han, W., 2024{\natexlab{e}}. Transportationgames: Benchmarking transportation knowledge of (multimodal) large language models. arXiv preprint arXiv:2401.04471.

\bibitem[{Zhang et~al.(2024{\natexlab{f}})Zhang, Kong, Zhou, Liu, Fu, and Shen}]{zhang2024comprehensive}
Zhang, Y., Kong, X., Zhou, W., Liu, J., Fu, Y., Shen, G., 2024{\natexlab{f}}. A comprehensive survey on traffic missing data imputation. IEEE Transactions on Intelligent Transportation Systems.

\bibitem[{Zhang et~al.(2025{\natexlab{b}})Zhang, Tan, Gan, Liu, and Atasoy}]{zhang2025operational}
Zhang, Y., Tan, X., Gan, M., Liu, X., Atasoy, B., 2025{\natexlab{b}}. Operational synchromodal transport planning methodologies: Review and roadmap. Transportation Research Part E: Logistics and Transportation Review 194, 103915.

\bibitem[{Zhang et~al.(2024{\natexlab{g}})Zhang, Sun, Wang, Nie, Ma, Sun, and Li}]{zhang2024large}
Zhang, Z., Sun, Y., Wang, Z., Nie, Y., Ma, X., Sun, P., Li, R., 2024{\natexlab{g}}. Large language models for mobility in transportation systems: A survey on forecasting tasks. arXiv preprint arXiv:2405.02357.

\bibitem[{Zhao et~al.(2024{\natexlab{a}})Zhao, Wang, Zhu, Chen, Huang, Bao, and Wang}]{zhao2024drivedreamer}
Zhao, G., Wang, X., Zhu, Z., Chen, X., Huang, G., Bao, X., Wang, X., 2024{\natexlab{a}}. Drivedreamer-2: Llm-enhanced world models for diverse driving video generation. arXiv preprint arXiv:2403.06845.

\bibitem[{Zhao et~al.(2023)Zhao, Zhou, Li, Tang, Wang, Hou, Min, Zhang, Zhang, Dong, et~al.}]{zhao2023survey}
Zhao, W.~X., Zhou, K., Li, J., Tang, T., Wang, X., Hou, Y., Min, Y., Zhang, B., Zhang, J., Dong, Z., et~al., 2023. A survey of large language models. arXiv preprint arXiv:2303.18223 1~(2).

\bibitem[{Zhao et~al.(2024{\natexlab{b}})Zhao, Lin, Zhu, Ye, Chen, Zheng, Ceze, Krishnamurthy, Chen, and Kasikci}]{zhao2024atom}
Zhao, Y., Lin, C.-Y., Zhu, K., Ye, Z., Chen, L., Zheng, S., Ceze, L., Krishnamurthy, A., Chen, T., Kasikci, B., 2024{\natexlab{b}}. Atom: Low-bit quantization for efficient and accurate llm serving. Proceedings of Machine Learning and Systems 6, 196--209.

\bibitem[{Zheng et~al.(2023{\natexlab{a}})Zheng, Abdel-Aty, Wang, Wang, and Ding}]{zheng2023trafficsafetygpt}
Zheng, O., Abdel-Aty, M., Wang, D., Wang, C., Ding, S., 2023{\natexlab{a}}. Trafficsafetygpt: Tuning a pre-trained large language model to a domain-specific expert in transportation safety. arXiv preprint arXiv:2307.15311.

\bibitem[{Zheng et~al.(2023{\natexlab{b}})Zheng, Abdel-Aty, Wang, Wang, and Ding}]{zheng2023chatgpt}
Zheng, O., Abdel-Aty, M., Wang, D., Wang, Z., Ding, S., 2023{\natexlab{b}}. Chatgpt is on the horizon: could a large language model be suitable for intelligent traffic safety research and applications? arXiv preprint arXiv:2303.05382.

\bibitem[{Zheng et~al.(2024{\natexlab{a}})Zheng, Xia, Huang, Zuo, Zhou, and Lu}]{zheng2024doe}
Zheng, W., Xia, Z., Huang, Y., Zuo, S., Zhou, J., Lu, J., 2024{\natexlab{a}}. Doe-1: Closed-loop autonomous driving with large world model. arXiv preprint arXiv:2412.09627.

\bibitem[{Zheng et~al.(2024{\natexlab{b}})Zheng, Wu, Yan, Tang, Zhao, Zhong, Chen, and Gong}]{zheng2024large}
Zheng, X., Wu, L., Yan, Z., Tang, Y., Zhao, H., Zhong, C., Chen, B., Gong, J., 2024{\natexlab{b}}. Large language models powered context-aware motion prediction. arXiv preprint arXiv:2403.11057.

\bibitem[{Zheng et~al.(2025)Zheng, Wang, Fu, and Ma}]{zheng2025estimating}
Zheng, Z., Wang, Z., Fu, H., Ma, W., 2025. Estimating erratic measurement errors in network-wide traffic flow via virtual balance sensors. Transportation Science.

\bibitem[{Zhong et~al.(2024)Zhong, Guo, Gao, Ye, and Wang}]{zhong2024memorybank}
Zhong, W., Guo, L., Gao, Q., Ye, H., Wang, Y., 2024. Memorybank: Enhancing large language models with long-term memory. In: Proceedings of the AAAI Conference on Artificial Intelligence. Vol.~38. pp. 19724--19731.

\bibitem[{Zhong et~al.(2023)Zhong, Rempe, Chen, Ivanovic, Cao, Xu, Pavone, and Ray}]{zhong2023language}
Zhong, Z., Rempe, D., Chen, Y., Ivanovic, B., Cao, Y., Xu, D., Pavone, M., Ray, B., 2023. Language-guided traffic simulation via scene-level diffusion. In: Conference on Robot Learning. PMLR, pp. 144--177.

\bibitem[{Zhou and Knoll(2024)}]{zhou2024gpt}
Zhou, X., Knoll, A.~C., 2024. Gpt-4v as traffic assistant: An in-depth look at vision language model on complex traffic events. arXiv preprint arXiv:2402.02205.

\bibitem[{Zhou et~al.(2025{\natexlab{a}})Zhou, Larintzakis, Guo, Zimmer, Liu, Cao, Zhang, Lakshminarasimhan, Strand, and Knoll}]{zhou2025tumtraffic}
Zhou, X., Larintzakis, K., Guo, H., Zimmer, W., Liu, M., Cao, H., Zhang, J., Lakshminarasimhan, V., Strand, L., Knoll, A.~C., 2025{\natexlab{a}}. Tumtraffic-videoqa: A benchmark for unified spatio-temporal video understanding in traffic scenes. arXiv preprint arXiv:2502.02449.

\bibitem[{Zhou and List(2010)}]{zhou2010information}
Zhou, X., List, G.~F., 2010. An information-theoretic sensor location model for traffic origin-destination demand estimation applications. Transportation science 44~(2), 254--273.

\bibitem[{Zhou et~al.(2023)Zhou, Liu, Zagar, Yurtsever, and Knoll}]{zhou2023vision}
Zhou, X., Liu, M., Zagar, B.~L., Yurtsever, E., Knoll, A.~C., 2023. Vision language models in autonomous driving and intelligent transportation systems. arXiv preprint arXiv:2310.14414.

\bibitem[{Zhou et~al.(2015)Zhou, Mo, Xiao, Chen, and Yin}]{zhou2015privacy}
Zhou, Y., Mo, Z., Xiao, Q., Chen, S., Yin, Y., 2015. Privacy-preserving transportation traffic measurement in intelligent cyber-physical road systems. IEEE Transactions on Vehicular Technology 65~(5), 3749--3759.

\bibitem[{Zhou et~al.(2024{\natexlab{a}})Zhou, Gu, Qu, Liu, Liu, and Yu}]{zhou2024urban}
Zhou, Z., Gu, Z., Qu, X., Liu, P., Liu, Z., Yu, W., 2024{\natexlab{a}}. Urban mobility foundation model: A literature review and hierarchical perspective. Transportation Research Part E: Logistics and Transportation Review 192, 103795.

\bibitem[{Zhou et~al.(2025{\natexlab{b}})Zhou, Haibo, Chen, Wang, Guan, Wu, Li, Huang, and Xue}]{zhou2025behaviorgpt}
Zhou, Z., Haibo, H., Chen, X., Wang, J., Guan, N., Wu, K., Li, Y.-H., Huang, Y.-K., Xue, C.~J., 2025{\natexlab{b}}. Behaviorgpt: Smart agent simulation for autonomous driving with next-patch prediction. Advances in Neural Information Processing Systems 37, 79597--79617.

\bibitem[{Zhou et~al.(2024{\natexlab{b}})Zhou, Huang, Li, Zhao, and Mu}]{zhou2024safedrive}
Zhou, Z., Huang, H., Li, B., Zhao, S., Mu, Y., 2024{\natexlab{b}}. Safedrive: Knowledge-and data-driven risk-sensitive decision-making for autonomous vehicles with large language models. arXiv preprint arXiv:2412.13238.

\bibitem[{Zhou et~al.(2024{\natexlab{c}})Zhou, Lin, Jin, and Li}]{zhou2024large}
Zhou, Z., Lin, Y., Jin, D., Li, Y., 2024{\natexlab{c}}. Large language model for participatory urban planning. arXiv preprint arXiv:2402.17161.

\bibitem[{Zhu et~al.(2018)Zhu, Yu, Wang, Ning, and Tang}]{zhu2018big}
Zhu, L., Yu, F.~R., Wang, Y., Ning, B., Tang, T., 2018. Big data analytics in intelligent transportation systems: A survey. IEEE Transactions on Intelligent Transportation Systems 20~(1), 383--398.

\bibitem[{Zhu et~al.(2024)Zhu, Yu, Zhao, Wei, and Liang}]{zhu2024unitraj}
Zhu, Y., Yu, J.~J., Zhao, X., Wei, X., Liang, Y., 2024. Unitraj: Universal human trajectory modeling from billion-scale worldwide traces. arXiv preprint arXiv:2411.03859.

\bibitem[{Ziegler et~al.(2019)Ziegler, Stiennon, Wu, Brown, Radford, Amodei, Christiano, and Irving}]{ziegler2019fine}
Ziegler, D.~M., Stiennon, N., Wu, J., Brown, T.~B., Radford, A., Amodei, D., Christiano, P., Irving, G., 2019. Fine-tuning language models from human preferences. arXiv preprint arXiv:1909.08593.

\bibitem[{Zou et~al.(2025)Zou, Yan, Hao, Hu, Wen, Liu, Zhang, Li, Li, Zheng, et~al.}]{zou2025deep}
Zou, X., Yan, Y., Hao, X., Hu, Y., Wen, H., Liu, E., Zhang, J., Li, Y., Li, T., Zheng, Y., et~al., 2025. Deep learning for cross-domain data fusion in urban computing: Taxonomy, advances, and outlook. Information Fusion 113, 102606.

\end{thebibliography}

\end{document}